\documentclass[11pt]{article}
\usepackage{amsmath}
\usepackage{amsthm}
\usepackage{amssymb}
\usepackage{graphicx}
\usepackage{adjustbox}
\usepackage{mathtools}
\usepackage{bbm}
\usepackage[margin=2.5cm]{geometry}
\usepackage{setspace}
\usepackage{epstopdf}
\usepackage{subfig}
\usepackage{empheq}
\usepackage{algorithm}
\usepackage{algorithmicx}
\usepackage{algpseudocode}
\usepackage{multirow}
\usepackage{url}
\usepackage[title]{appendix}
\usepackage{enumitem}
\usepackage[svgnames]{xcolor}
\usepackage{diagbox}
\usepackage{tabu}
\usepackage{nicematrix}
\usepackage{mathrsfs}
\usepackage{amsbsy}
\usepackage{derivative}
\usepackage{caption}
\usepackage{hyperref}

\makeatletter

\theoremstyle{plain}
\newtheorem{thm}{\protect\theoremname}
  \theoremstyle{definition}

  \theoremstyle{plain}
  \newtheorem{lem}[thm]{\protect\lemmaname}

  \newtheorem{assump}{Assumption}

  \providecommand{\propname}{Proposition}
  \providecommand{\condname}{Condition}

\makeatother

  \providecommand{\definitionname}{Definition}
  \providecommand{\lemmaname}{Lemma}
  \providecommand{\corollaryname}{Corollary}
\providecommand{\theoremname}{Theorem}

\usepackage{endnotes}
\usepackage{color}
\usepackage{xparse}

\title{Blind Noise Equalization for Large Data Matrices \\under General Variance Profiles}

\author{Boris Landa${^{1,5,*}}$~~~~Francesco Strino${^{2}}$~~~~Fabio Parisi${^{2}}$\\
Ruiqi Li${^{3,4}}$~~~~Yuval Kluger${^{3,4,5}}$\\[0.5em]
\small{${^1}$Department of Electrical and Computer Engineering, Yale University}\\
\small{${^2}$PCMGF Limited, United Kingdom}\\
\small{${^3}$Computational Biology and Biomedical Informatics Program, Yale University}\\
\small{${^4}$Department of Pathology, Yale School of Medicine}\\
\small{${^5}$Program in Applied Mathematics, Yale University}\\
\small{${^*}$Corresponding author. Email: boris.landa@yale.edu}
}

\begin{document}

\maketitle

\begin{abstract}
Low-rank approximation is a basic tool for inference from large and noisy data matrices. However, when noise variances differ across entries—a setting known as heteroskedasticity—the noise can distort the observed spectrum, producing spurious components and obscuring weak signals. To address these challenges, we develop BRINE (Blind Resolvent-based Iterative Noise Equalizer), a data-driven procedure that iteratively rescales rows and columns to equalize the average noise variance across both dimensions. Under suitable conditions, this procedure restores the Marchenko–Pastur law for the noise spectrum and can amplify signal components without prior knowledge of the signal or noise structure. A key idea underlying our approach is to equalize the diagonal entries of the data resolvent as a proxy for equalizing the unknown noise variances. At the population level, we identify a potential function that characterizes this equalization and motivates an iterative minimization procedure. We prove convergence of the iterations to the true equalizer for arbitrary positive variance profiles, with a geometric rate under mild conditions. The updates can be estimated directly from noisy data, and we establish probabilistic error bounds for the scaling factors. Simulations and experiments across diverse biological data modalities and preprocessing pipelines illustrate the effectiveness of our approach.
\end{abstract}

\section{Introduction}
\label{sec:introduction}

Low-rank approximation is a central tool for extracting structure from large data matrices. A common pipeline is to apply the singular value decomposition (SVD) to an observed data matrix, possibly after preprocessing, or equivalently to eigendecompose the associated covariance matrix.
The leading singular vectors or eigenvectors are then used for visualization, denoising, clustering, dimensionality reduction, and downstream inference. A basic question is how to distinguish the low-rank signal from the noise, which is often formulated via the signal-plus-noise model
\begin{equation}
Y = X + E,
\label{eq: signal-plus-noise}
\end{equation}
where $Y \in \mathbb{R}^{m\times n}$ is the observed data, $X$ is an unknown signal matrix with rank $r \ll \min\{m,n\}$, and $E$ is a random noise matrix. The signal $X$ is either deterministic or random but independent of $E$. We assume for simplicity that $m\leq n$; otherwise, we can equivalently transpose the data. Common tasks include detecting the presence of a nonzero signal $X$, estimating the number of low-rank components to retain, and recovering or enhancing the signal from the noisy data $Y$.

A large body of work addresses these tasks through the spectrum of $Y$; see, e.g.,~\cite{cattell1966scree,wold1978cross,hoff2007model,owen2009bi,gavish2014optimal,chatterjee2015matrix,fan2014principal,johnstone2018pca,fan2020estimating,hong2020selecting,choi2017selecting,kritchman2008determining,johnstone2017roy} and references therein. In the classical homoskedastic setting, the entries of $E$ are independent with mean zero and identical variance $\sigma^2/n$. Then, under mild regularity conditions, as $m,n\to\infty$ with $m/n\to\gamma\in(0,1]$, the empirical eigenvalue distribution of $EE^T$ converges to the Marchenko--Pastur (MP) law with scale $\sigma^2$~\cite{marvcenko1967distribution,bai2010spectral}, while the largest eigenvalue of $EE^T$ converges almost surely to the upper edge $\beta_+=\sigma^2(1+\sqrt{\gamma})^2$~\cite{geman1980limit,yin1988limit}. These results motivate a simple spectral approach: estimate the signal rank by counting the eigenvalues of $YY^T$ that exceed $\beta_+$, or equivalently the singular values of $Y$ that exceed $\sigma(1+\sqrt{\gamma})$. A more refined justification comes from the theory of spiked random matrices~\cite{baik2005phase,baik2006eigenvalues,benaych2011eigenvalues,nadler2008finite,paul2007asymptotics}, which characterizes when empirical outliers separate from the noise bulk and how their singular vectors align with the corresponding signal directions. This line of work has led to improved denoising techniques based on precise singular-value thresholding and shrinkage~\cite{gavish2017optimal,gavish2014optimal,nadakuditi2014optshrink,leeb2022optimal,ding2020high}.

In many real datasets, however, the noise is heteroskedastic: different entries, rows, or columns may have substantially different noise levels. In this work, we assume that the entries of the noise  matrix $E$ are independent with mean zero, but have an unknown variance profile
\begin{equation}
S_{ij} = \mathbb{E} \left[ E_{ij}^2 \right] > 0 .
\label{eq:intro_variance_profile}
\end{equation}
Noise heteroskedasticity arises naturally in many applications involving count and nonnegative data, where the variance depends on the mean, such as in Poisson, binomial, negative binomial, or gamma-type models. Examples include network traffic data~\cite{shen2005analysis}, photon-limited imaging~\cite{salmon2014poisson}, document topic models~\cite{wallach2006topic}, single-cell RNA sequencing~\cite{hafemeister2019normalization}, spatial transcriptomics~\cite{burgess2019spatial}, and high-throughput chromosome conformation capture~\cite{johanson2018genome}. Heteroskedasticity also arises from nonlinear transformations of the data, clipping or saturation effects~\cite{foi2009clipped}, and varying experimental conditions~\cite{cochran1977statistically,tamuz2005correcting}.

In heteroskedastic settings, the spectrum of the noise can differ substantially from the homoskedastic case. When the variance profile is strongly nonuniform, the eigenvalues of $EE^T$ can be spread over a larger interval, making weak signal components harder to distinguish from noise. Additionally, if a small number of entries, rows, or columns have much larger variances than the rest, the noise itself can produce large singular values that may be mistaken for signal components.

Numerous methods for low-rank signal detection and recovery account for heteroskedasticity by estimating the noise spectral distribution or its upper edge and then adapting the corresponding thresholding or shrinkage rule; see, e.g.,~\cite{nadakuditi2014optshrink,gavish2022matrix,donoho2023screenot,ke2021estimation,hong2020selecting,su2022adaptive,ding2022tracy}. Such approaches, however, leave the observed spectrum unchanged and do not address the fundamental difficulty: heteroskedastic noise can distort the data spectrum in a prohibitive way, reducing the spectral separation between signal and noise and even producing outlying noise components that may be mistaken for signal.

\begin{figure}[H]
    \centering
    \includegraphics[width=\linewidth]{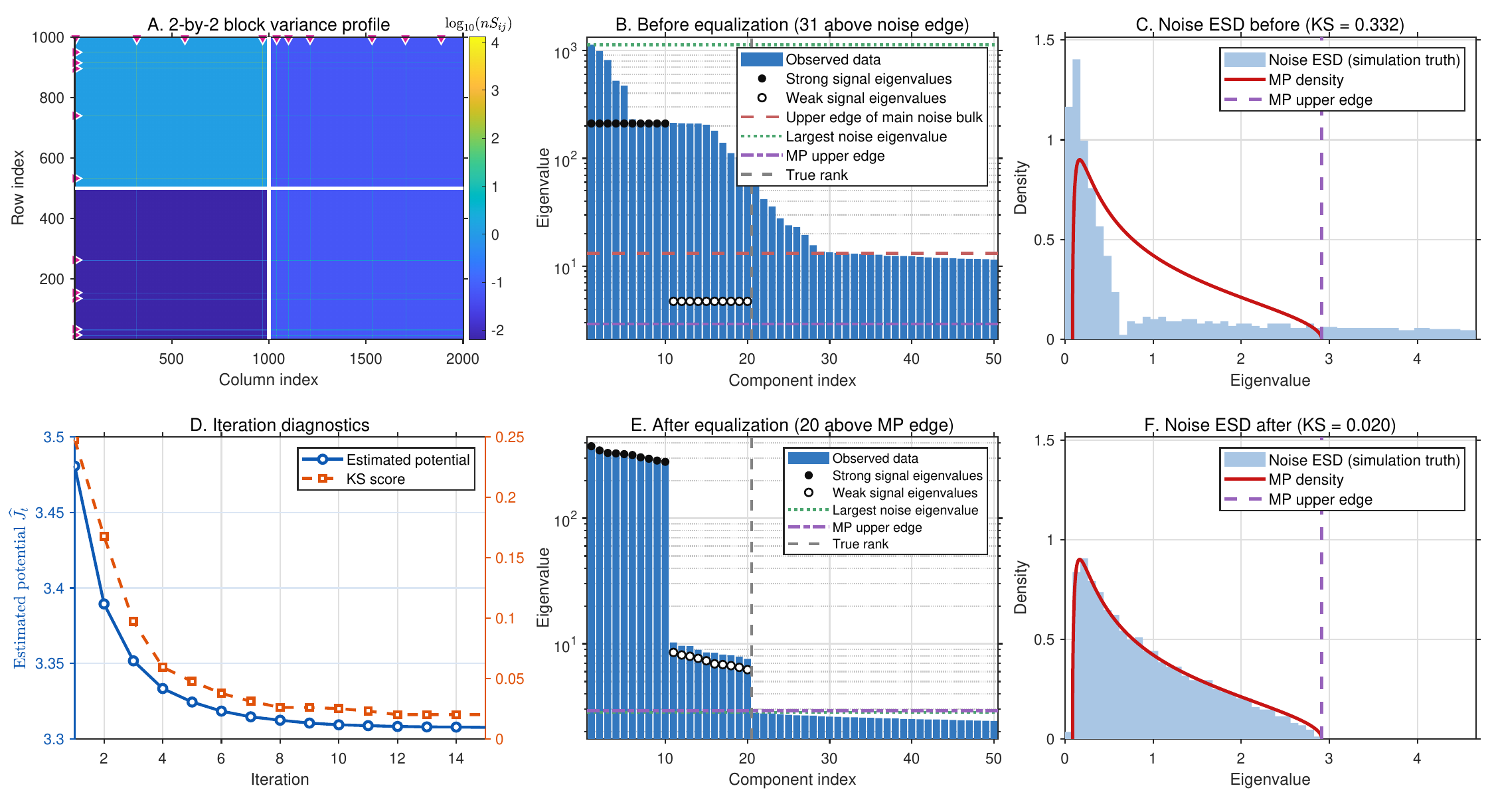}
    \caption{\small
    Stylized example with $m=1000$, $n=2000$, and a rank-$20$ signal containing ten strong and ten weak components. The noise is Gaussian with the variance profile shown in Panel~A: a $2\times2$ block structure with ten scattered high-variance rows and ten scattered high-variance columns (indicated by triangles). Panel~B shows the leading eigenvalues of $YY^T$. The blue bars are the observed eigenvalues, while the filled and open black markers show the eigenvalues of $XX^T$ associated with the strong and weak signal components, respectively. The horizontal lines mark the MP upper edge (dashed purple), the upper edge of the main noise bulk (dashed red), and the largest eigenvalue of the noise matrix $EE^T$ (dotted green). The vertical line marks the true rank. Panel~C compares the empirical spectral distribution of the noise with the MP law. Panels~D--F illustrate the benefits of our proposed method. Panel~D shows the estimated potential and the Kolmogorov-Smirnov (KS) statistic for goodness-of-fit to the MP law over the iterations. Panel~E shows the leading eigenvalues after applying the proposed normalization, with the MP upper edge shown by the horizontal line. Panel~F compares the empirical spectral distribution of the normalized noise with the MP law.
    }
    \label{fig:stylized_intro}
\end{figure}

Figure~\ref{fig:stylized_intro} illustrates these challenges in a stylized simulation with dimensions $m=1000$, $n=2000$, and a rank-$20$ signal consisting of ten strong and ten weak components. The variance profile is shown in Panel~A: it has a $2\times2$ block structure with variances $[0.1, 1; 20, 1]$ up to a global scalar normalization, along with ten scattered high-variance rows and ten scattered high-variance columns; complete simulation details are provided in Appendix~\ref{app:stylized_simulation}.

Panel~B shows the leading eigenvalues of the observed Gram matrix $Y Y^T$ (blue bars), together with the signal-only eigenvalues associated with the strong and weak components (filled and open black circles, respectively). Notably, the upper edge of the main noise bulk is more than threefold larger than the MP upper edge due to the heterogeneity of the variance profile. In addition, the largest eigenvalue of the noise $E E^T$ matches the largest eigenvalue of the observed Gram matrix $Y Y^T$ due to the high-variance rows and columns, which produce large outlier components that are reflected in the observed spectrum. Consequently, 31 observed eigenvalues lie above the main-bulk edge, substantially exceeding the true signal rank of 20. In this setting, the ten strong signal-only eigenvalues lie above the main-bulk edge, but the ten weak ones lie substantially below it and are masked by the noise. Panel~C further shows that the noise spectral distribution differs substantially from the MP law: it is more spread out and extends to larger eigenvalues than the MP distribution.

A natural strategy for handling heteroskedasticity is to modify the data or its empirical \sloppy Gram/covariance matrix prior to spectral analysis. For example, HeteroPCA~\cite{zhang2022heteroskedastic,zhou2025deflated} estimates the low-rank signal subspace by correcting the bias in the main diagonal of the empirical Gram matrix $Y Y^T$. In particular, it performs iterative imputation of the main diagonal of $Y Y^T$ via its low-rank approximation, assuming knowledge of the signal's rank $r$. An alternative approach is to directly stabilize the noise in the data before applying spectral procedures. Under one-sided heteroskedasticity, where the noise variance varies only across the rows or columns, several works study whitening, weighting, and variance-homogenization procedures~\cite{leeb2021optimal,hong2018asymptotic,gavish2022matrix,hong2018optimally,liu2018pca}. They show that appropriately transforming or weighting the heteroskedastic dimension of the data can substantially improve low-rank signal detection and recovery.

The heteroskedastic setting we consider here is two-sided: the variance profile $S_{ij}$ may vary jointly across both rows and columns. To address this setting, recent works~\cite{leeb2021matrix,landa2022biwhitening,landa2025dyson,zhang2024matrix} proposed transforming the rows and columns simultaneously. Here, we consider diagonal scaling of the form
\begin{equation}
\widetilde Y
=
\bigl(D\{\mathbf{x}\}\bigr)^{1/2}
Y
\bigl(D\{\mathbf{y}\}\bigr)^{1/2}
=
\widetilde X+\widetilde E,
\label{eq: Y_tilde def}
\end{equation}
for positive vectors $\mathbf{x}\in\mathbb{R}^m$ and $\mathbf{y}\in\mathbb{R}^n$, where $D\{\mathbf{v}\}$ denotes the diagonal matrix with $\mathbf{v}$ on its diagonal, and
\[
\widetilde X
=
\bigl(D\{\mathbf{x}\}\bigr)^{1/2}
X
\bigl(D\{\mathbf{y}\}\bigr)^{1/2},
\qquad
\widetilde E
=
\bigl(D\{\mathbf{x}\}\bigr)^{1/2}
E
\bigl(D\{\mathbf{y}\}\bigr)^{1/2}.
\]
Since the two diagonal scaling matrices are invertible, this transformation preserves the rank of the signal, i.e., $\operatorname{rank}(\widetilde X)
=
\operatorname{rank}(X)
=
r,
$
while modifying the noise variance profile according to
\begin{equation}
\widetilde S_{ij}
=
\mathbb{E}\!\left[\widetilde E_{ij}^{\,2}\right]
=
x_i S_{ij}y_j.
\label{eq:intro_normalized_variance}
\end{equation}

One special case is the rank-one variance profile $S=\mathbf{u}\mathbf{v}^T/n$, where $\mathbf{u}\in\mathbb{R}^m$ and $\mathbf{v}\in\mathbb{R}^n$ are positive vectors. Choosing $\mathbf{x}=\mathbf{u}^{-1}$ and $\mathbf{y}=\mathbf{v}^{-1}$ in~\eqref{eq: Y_tilde def} yields $\widetilde S_{ij}=1/n$ for all $i,j$, so that the normalized noise becomes completely homoskedastic. Leeb~\cite{leeb2021matrix} utilized this normalization for matrix denoising: an optimally tuned spectral denoiser is applied to the normalized matrix $\widetilde{Y}$, and the result is then converted back to an estimate of the original signal matrix $X$ by applying the inverse row and column transforms. It was also shown that, for sufficiently generic signal matrices, this normalization can increase the ratios between the signal singular values and the operator norm of the noise, making weak signal components easier to detect and recover. Related recent work~\cite{zhang2024matrix} considered signal recovery under more general left and right linear transformations, and derived optimal estimators in that setting, assuming the rank-one variance profile is known. Since the variance profile $S=\mathbf{u}\mathbf{v}^T/n$ is typically unknown,~\cite{leeb2021matrix} proposed to estimate $\mathbf{u}$ and $\mathbf{v}$ directly from the empirical variances of the data $Y$ across the rows and columns, respectively. However, this approach can be highly inaccurate if strong signal components are present, in which case the empirical variances (and consequently the normalizing factors) are strongly biased by the signal.

To address this limitation,~\cite{landa2025dyson} developed a method that estimates the required row and column scaling factors from the resolvent of the data (see eq.~\eqref{eq: definition of R_cal and Y_cal_tilde} in Section~\ref{sec: resolvent and Dyson equation}), using its connection to the Dyson equation~\cite{alt2017local,erdHos2019random} (see eq.~\eqref{eq: Dyson eq} in Section~\ref{sec: resolvent and Dyson equation}). This approach supports broad classes of noise distributions and is robust to strong low-rank signal components. In particular, the estimated scaling factors can remain accurate even when some signal components are arbitrarily large, provided suitable regularity conditions hold. However, the approach in~\cite{landa2025dyson} relies on the particular form of the Dyson equation under the rank-one variance profile $S=\mathbf{u}\mathbf{v}^T/n$, exploiting the special structure therein to provide closed-form estimates of $\mathbf{u}$ and $\mathbf{v}$. While some theoretical guarantees in~\cite{landa2025dyson} extend to higher-rank variance patterns that are sufficiently incoherent, the method and its guarantees do not cover general variance profiles $S$ with arbitrary positive structure.

When the variance profile $S$ has a general positive structure, the normalization~\eqref{eq: Y_tilde def} cannot make the noise exactly homoskedastic with identical variances across all entries. To address this case,~\cite{landa2022biwhitening} proposed to choose the scaling factors $\mathbf{x}$ and $\mathbf{y}$ in~\eqref{eq: Y_tilde def} so that the average noise variance is equalized across both dimensions, i.e., the average entry of $\widetilde S$ in each row and in each column is $1/n$. This condition is weaker than making all entries of $\widetilde S$ equal, but was nonetheless shown to be sufficient, under mild conditions, to recover the standard spectral behavior of homoskedastic noise, namely the MP law for $\widetilde E\widetilde E^T$ and convergence of the largest eigenvalue to the MP upper edge $\beta_+$~\cite{landa2022biwhitening}.
This provides a principled spectral threshold for signal detection and rank estimation, improving the detection of weak signal components~\cite{landa2022biwhitening} and, when combined with spectral denoising, the accuracy of low-rank signal recovery~\cite{landa2025dyson}.
While no closed-form expression exists for the required scaling factors in general, they can be found by the classical Sinkhorn--Knopp algorithm~\cite{sinkhorn1967concerning,sinkhorn1967diagonal} if $S$ is known. This procedure alternates between normalizing the row and column sums of $S$ repeatedly, until convergence to their prescribed values.

Since the general variance profile $S$ is typically unknown,~\cite{landa2022biwhitening} derived a data-driven procedure for estimating the required $\mathbf{x}$ and $\mathbf{y}$ under the assumption that $S_{ij}$ is a quadratic polynomial in the mean of $Y_{ij}$. Under this assumption, the Sinkhorn--Knopp algorithm can be applied to an unbiased estimator of the entrywise noise variances. This assumption holds for several prototypical count and nonnegative distributions, such as Poisson, binomial, negative binomial, and gamma distributions~\cite{morris1982natural}. Subsequently,~\cite{stanley2025principled} combined the resulting procedure with singular value shrinkage, demonstrating accurate rank estimation and improved recovery of signal structure across diverse biological datasets. However, these approaches require the existence of a mean--variance relationship with a prescribed form in the data. Such a relationship may fail to hold after common preprocessing steps and nonlinear transformations of count data, or for general non-count data distributions.

\subsection{Our approach and contributions}

In this work, we consider the problem of blind noise variance equalization in data matrices: finding the row and column scaling factors $\mathbf{x}$ and $\mathbf{y}$ in~\eqref{eq: Y_tilde def} that equalize the average noise variance across both dimensions. We show that this problem can be solved reliably under a broad class of high-dimensional signal-plus-noise models, where the unknown variance profile $S$ can have an arbitrary positive pattern. Our approach accommodates general noise distributions and low-rank signal structures, including mixtures of strong and weak components, and requires neither a mean--variance relationship nor prior knowledge of the signal or noise structure. Thus, our approach overcomes key limitations of existing methods. Under suitable conditions, the resulting normalization restores the MP law for the noise and can amplify signal components, enabling existing spectral methods for signal detection and recovery to be applied more effectively and reliably to the normalized data.

To accomplish this, we develop BRINE (Blind Resolvent-based Iterative Noise Equalizer), a data-driven method that iteratively rescales the rows and columns of the data in the spirit of the Sinkhorn--Knopp algorithm, but without the need to know or estimate the variance profile $S$. A key idea underpinning our approach is that, in a suitable high-dimensional regime, the diagonal entries of the data resolvent approach known values under the correct noise variance equalization. These values provide targets for our iterative procedure. To derive the updates, we construct a potential function that augments the classical logarithmic barrier function---whose alternating minimization yields the Sinkhorn--Knopp algorithm. Our augmented potential preserves the same stationary points, but its minimization leads to updates that can be estimated directly from the data resolvent through its connection to the Dyson equation. Our procedure is summarized in Algorithm~\ref{alg:BRINE} in Section~\ref{sec: data-driven procedure}; a Python implementation is available \href{https://github.com/KlugerLab/BRINE}{here}.

Panels~D--F in Figure~\ref{fig:stylized_intro} illustrate the behavior of BRINE in the previous simulation setup. Panel~D shows that both the estimated potential and the MP goodness-of-fit score decrease and then stabilize along the iterations. Panel~E shows that after equalization, exactly twenty eigenvalues of the observed matrix lie clearly above the MP upper edge, matching the true signal rank, while the variance-induced components present before normalization no longer appear above the spectral threshold. Finally, the MP upper edge is now very close to the largest eigenvalue of the normalized noise, and its empirical spectral distribution also closely follows the MP law, with a substantially improved KS score; see Panel~F.

Figure~\ref{fig:comparison_previous_normalizations} in
Appendix~\ref{appendix: comparison to previous normalizations} compares BRINE with the
normalizations proposed in~\cite{landa2025dyson} and~\cite{leeb2021matrix} under the same simulation setup. Neither
of these two methods restores the MP law since the
variance profile lies outside the rank-one or
separable models for which these procedures were developed; moreover, the
strong signal components affect the direct row and column empirical variance
estimates used by~\cite{leeb2021matrix}, causing it to fail. We note that the procedure of~\cite{landa2022biwhitening} is not part of this comparison since it is not directly
applicable: the data-generating process does not impose a
mean--variance relationship that could be used to infer the noise variance
profile.

We now describe the organization of the paper and the main contributions in more detail. In Section~\ref{sec: preliminaries} we review the technical background needed for the rest of the paper: the matrix scaling equations, the Sinkhorn--Knopp algorithm, the logarithmic barrier function, the MP law, and the Dyson equation and its connection to the resolvent.
Section~\ref{sec: method and guarantees} develops the proposed method and its theoretical guarantees.

First, in Section~\ref{sec: population-level procedure}, we introduce the potential function at the core of our approach, describe the corresponding population-level iterative minimization procedure, and characterize its properties. In particular, we show that the critical points of this potential coincide with the desired scaling factors that equalize the average noise variance across rows and columns, and prove that the population-level procedure converges to such a scaling for every positive variance profile; see Lemma~\ref{lem: properties of potential function} and Theorem~\ref{thm:population_convergence}. This theorem also establishes geometric convergence of the potential function and the scaling factors in a suitable metric under uniform upper and lower bounds on the variances and a bounded aspect ratio (Assumption~\ref{assump: population bounds}).

Second, in Section~\ref{sec: data-driven procedure}, we derive the data-driven version of this procedure, in which the required quantities are estimated from the observed matrix. Under additional assumptions on the noise and signal (Assumption~\ref{assump: main assumption}), we establish high-probability bounds on the error of the estimated scaling factors relative to the true ones; see Theorem~\ref{thm: main theoretical guarantees}. These bounds depend explicitly on the matrix dimensions, the number of iterations, and the localization of the signal singular vectors. This result supports general noise distributions satisfying suitable moment bounds and allows for arbitrary magnitudes of signal components, provided that the signal singular vectors satisfy a mild delocalization condition and the rank does not grow too quickly with the dimension.

Third, in Section~\ref{sec: signal enhancement}, we analyze how the proposed normalization affects the signal itself. Our result shows that beyond standardizing the spectrum of the noise, the proposed normalization can enhance the signal components. Specifically, we show that for orthogonally invariant signal singular vectors, and if the global average noise variance is $1/n$, the singular values of the scaled signal matrix $\widetilde X$ are asymptotically no smaller than those of $X$, with a gain of at least $e^{\Delta(S)/2}$, where $\Delta(S)$ is the relevant logarithmic-barrier gap; see Theorem~\ref{thm: generic singular value gain}. We also provide an explicit lower bound on this gain in terms of the row and column sums of the variance profile $S$. This phenomenon is evident in Panels~B and~E of Figure~\ref{fig:stylized_intro}, where the weak signal-only eigenvalues increase after equalization.

Section~\ref{sec:experiments} provides a comprehensive empirical study of the proposed normalization. We first examine the convergence and estimation accuracy of the scaling factors in a controlled simulation, comparing BRINE with the population-level procedure and existing normalization methods; see Section~\ref{sec:experiments-convergence} and Figure~\ref{fig:convergence-simulation}. In Section~\ref{sec:experiments-mp}, we evaluate our approach across 11 real-world biological datasets spanning 8 distinct data modalities and over 20 dataset-transformation combinations, covering a wide range of standard preprocessing pipelines. In all cases, the empirical spectral distribution is transformed from an arbitrary, data-dependent shape to one closely following the MP law, without any tuning specific to the data type or preprocessing; see Figure~\ref{fig:mp-main}.
These results suggest that the low-rank signal-plus-noise model with independent entries and a general variance pattern provides a useful approximation for a broad range of datasets encountered in practice.

To further support this observation, in Section~\ref{sec:experiments-correlated} we perform controlled simulation studies in which the assumptions of independent noise and low-rank signal are gradually relaxed; see Figure~\ref{fig:sims-correlated-noise}. We show that agreement with the MP law deteriorates under substantial deviations from the independence assumption, but remains stable under moderate or weak deviations. These findings suggest that the MP fit can serve as a useful diagnostic of model mismatch in applications. We also examine the effect of increasing signal rank beyond our theoretical analysis. The primary effect is a gradual decrease in the largest eigenvalue of the transformed noise, which remains below the MP upper edge throughout the tested range. Consequently, when the MP upper edge serves as the detection threshold, the proposed normalization can support reliable rank estimation without introducing false detections even when the signal rank is relatively high, with only a mild loss of detection sensitivity.

Beyond validating the underlying model and proposed approach, we demonstrate the practical benefit of the proposed normalization for downstream analysis in Section~\ref{sec:experiments-scrna}; see Figure~\ref{fig:zheng_scores}. In particular, we demonstrate that when integrated into a standard preprocessing pipeline for single-cell RNA sequencing, our blind noise equalization can substantially improve the ability to identify and extract rare cell populations from the data.

Finally, Section~\ref{sec: discussion} further discusses a range of topics, including the scope of our contributions, practical considerations, computational scalability, limitations of the proposed framework, and directions for future research.

\section{Preliminaries} \label{sec: preliminaries}

\subsection{Matrix scaling and the Marchenko-Pastur law} \label{Sec: sinkhorn scaling the MP law}
Our goal is to find the scaling factors $\mathbf{x}$ and $\mathbf{y}$ in~\eqref{eq: Y_tilde def} that equalize the average noise variance across the rows and columns, i.e., those that make the average entry of the normalized variance matrix $\widetilde{S}$ from~\eqref{eq:intro_normalized_variance} to be identical across both dimensions. A key fact enabling this goal is that there exists a pair of positive scaling vectors $(\mathbf{x}^\star,\mathbf{y}^\star)$ that satisfies the system of equations
\begin{equation}
    \frac{1}{n}\sum_{j=1}^n \mathbf{x}^\star_i {S}_{i,j} \mathbf{y}^\star_j
    = \frac{1}{n}\sum_{j=1}^n \widetilde{S}^\star_{i,j} =  \frac{1}{n}, \qquad\qquad \frac{1}{m}\sum_{i=1}^m \mathbf{x}^\star_i {S}_{i,j} \mathbf{y}^\star_j
    = \frac{1}{m}\sum_{i=1}^m \widetilde{S}^\star_{i,j} = \frac{1}{n}, \label{eq: scaling equations}
\end{equation}
for all $i\in [m]$ and $j\in [n]$, where we defined $\widetilde{S}^\star := D\{\mathbf{x}^\star\} S D\{\mathbf{y}^\star\}$. In other words, the average noise variance across each row and each column after normalization is exactly $1/n$. This is a special case of the \textit{matrix scaling} problem~\cite{idel2016review}, where the goal is to scale the rows and columns of a given nonnegative matrix to attain prescribed row and column sums. In our setting, the pair $(\mathbf{x}^\star,\mathbf{y}^\star)$ that solves the equations~\eqref{eq: scaling equations} is unique up to a trivial scalar ambiguity, i.e., it can only be replaced with $(a \mathbf{x}^\star, a^{-1} \mathbf{y}^\star)$ for any $a>0$, and the matrix $\widetilde{S}^\star$ is unique~\cite{sinkhorn1967diagonal,marshall1968scaling} (see also Proposition 2.1 in~\cite{landa2022biwhitening}).

Given $S$, the scaling equations~\eqref{eq: scaling equations} can be solved by Sinkhorn's algorithm~\cite{sinkhorn1967diagonal}, which iteratively normalizes the rows and columns of $S$ to achieve the prescribed row and column sums, respectively, until convergence. Specifically, given the row scaling vector in iteration $t\geq 0$, denoted by $\mathbf{x}^{(t)}$, the column scaling vector for the same iteration is computed by $\mathbf{y}_j^{(t)} = m/(n \sum_{i=1}^{m} \mathbf{x}_i^{(t)} S_{i,j})$ for all $j\in [n]$. Then, the row scaling vector for the next iteration is given by $\mathbf{x}_i^{(t+1)} = 1/( \sum_{j=1}^{n} S_{i,j} \mathbf{y}_j^{(t)})$ for all $i \in [m]$. This procedure is repeated until the scaled matrix $D\{\mathbf{x}^{(t)}\} S D\{\mathbf{y}^{(t)}\}$ satisfies the prescribed row and column sums to a desired precision.

The scaling equations~\eqref{eq: scaling equations} and Sinkhorn's algorithm can also be derived via the minimization of the \textit{logarithmic barrier function}
\begin{equation}
    L(\mathbf{x},\mathbf{y}) = \mathbf{x}^T S \mathbf{y} - \sum_{i=1}^m \log (\mathbf{x}_i) - \frac{m}{n}\sum_{j=1}^n \log (\mathbf{y}_j), \label{eq: logarithmic barrier function}
\end{equation}
 over the positive vectors $\mathbf{x}$ and $\mathbf{y}$. In particular, the critical points of the logarithmic barrier function~\eqref{eq: logarithmic barrier function}, which are also its global minimizers, coincide with the solutions to the scaling equations~\eqref{eq: scaling equations}. Sinkhorn's algorithm can then be interpreted as an alternating minimization of~\eqref{eq: logarithmic barrier function}, i.e., alternating between minimizing~\eqref{eq: logarithmic barrier function} over $\mathbf{x}$ while keeping $\mathbf{y}$ fixed and vice-versa; see~\cite{idel2016review} and references therein.

After the normalization~\eqref{eq: Y_tilde def} with the scaling factors $(\mathbf{x},\mathbf{y}) = (\mathbf{x}^\star,\mathbf{y}^\star)$, under suitable conditions, the spectral distribution of $\widetilde{E} \widetilde{E}^T$ satisfies the Marchenko-Pastur (MP) law in the asymptotic regime $m,n\rightarrow\infty$ with $m/n \rightarrow \gamma \in (0,1]$~\cite{landa2022biwhitening}. Concretely, consider the Empirical Spectral Distribution (ESD) of $\Sigma:= \widetilde{E} \widetilde{E}^T$, given by
\begin{equation}
    \hat{F}_{\Sigma} (\tau)= \frac{1}{m} \sum_{i=1}^m \mathbbm{1}\left(\lambda_i \{ \Sigma \} \leq \tau \right), \label{eq:empirical spectral distribution}
\end{equation}
where $\mathbbm{1}(\cdot)$ is the indicator function, and $\lambda_i \{ \Sigma \}$ is the $i$'th largest eigenvalue of $\Sigma$. Then, as $m,n\rightarrow \infty$ and $m/n\rightarrow \gamma \in (0,1]$, under suitable boundedness conditions on the variances $S_{i,j}$ and the higher-order moments of the noise variables $E_{i,j}$, the ESD $\hat{F}_{\Sigma}(\tau)$ converges almost surely to the MP distribution $F_{\gamma}(\tau)$~\cite{marvcenko1967distribution}, which is the cumulative distribution function of the MP density
\begin{equation}
    f_{\gamma}(\tau) = \frac{\sqrt{(\beta_+ - \tau)(\tau - \beta_-)}}{2\pi \gamma \tau} \mathbbm{1}\left( \beta_- \leq \tau \leq \beta_+\right), \label{eq:MP density}
\end{equation}
where $\beta_{\pm} = (1\pm \sqrt{\gamma})^2$~\cite{landa2022biwhitening}. Furthermore, the largest eigenvalue $\lambda_1\{\Sigma\}$ converges almost surely to the upper edge $\beta_+$; see Proposition 4.1 in~\cite{landa2022biwhitening}.

Since the rank of the normalized signal $\widetilde{X}$ remains unchanged by the diagonal scaling in~\eqref{eq: Y_tilde def}, i.e., $\operatorname{Rank}\{\widetilde{X}\} = r$, the rank can be estimated for large matrix dimensions by counting how many eigenvalues of $\widetilde{Y} \widetilde{Y}^T$ exceed $(1 + \sqrt{m/n})^2$ (by an arbitrarily small fixed margin)~\cite{landa2022biwhitening,landa2025dyson}. This approach is guaranteed to identify only true signal components, i.e., the rank estimate is asymptotically a lower bound on the true rank with high probability; see, e.g., Theorem 2.3 in~\cite{landa2022biwhitening}.

\subsection{The resolvent and the Dyson equation} \label{sec: resolvent and Dyson equation}
Let $\widetilde{\mathcal{E}},\widetilde{\mathcal{S}} \in\mathbb{R}^{(m+n)\times (m+n)}$ be the symmetrized versions of the scaled noise matrix $\widetilde{E}$ and the scaled variance matrix $\widetilde{S}$, respectively,
\begin{equation}
    \widetilde{\mathcal{E}} =
    \begin{bmatrix}
    \mathbf{0}_{m\times m} & \widetilde{E} \\
    \widetilde{E}^T & \mathbf{0}_{n\times n}
    \end{bmatrix}, \qquad\qquad
    \widetilde{\mathcal{S}} =
    \begin{bmatrix}
    \mathbf{0}_{m\times m} & \widetilde{S} \\
    \widetilde{S}^T & \mathbf{0}_{n\times n}
    \end{bmatrix}, \label{eq: definition of E_cal and S_cal}
\end{equation}
where $\mathbf{0}_{m\times m}$ and $\mathbf{0}_{n\times n}$ are $m\times m$ and $n\times n$ zero matrices, respectively. The resolvent of the scaled noise $\widetilde{\mathcal{E}}$ is the complex-valued matrix $R(z) \in \mathbb{C}^{(m+n)\times(m+n)}$ defined as
\begin{equation}
    R(z) = (\widetilde{\mathcal{E}} - z I_{m+n})^{-1}, \label{eq: resolvent def}
\end{equation}
where $I_{m+n}$ is the $(m+n)\times (m+n)$ identity matrix and $z\in\mathbb{C}^+$ is a point in the complex upper half-plane $\mathbb{C}^+$.

For a wide range of noise distributions, if the matrix dimensions $m$ and $n$ are sufficiently large, then the main diagonal of the noise resolvent $R(z)$ approximates a deterministic vector $\mathbf{f}\in\mathbb{C}^{m+n}$ that satisfies the Dyson equation
\begin{equation}
    z + \sum_{\ell=1}^{m+n}\widetilde{\mathcal{S}}_{k,\ell}\mathbf{f}_\ell = -\frac{1}{\mathbf{f}_k}, \label{eq: Dyson eq}
\end{equation}
for all $k\in [m+n]$ and any $z\in \mathbb{C}^+$; see~\cite{erdHos2019random,alt2017local} and references therein. The vector $\mathbf{f}$ is the unique solution to the Dyson equation~\eqref{eq: Dyson eq} with a strictly positive imaginary part; see, e.g.,~\cite{ajanki2019quadratic}.
In the special case where $z$ is purely imaginary, specifically $z = \imath \eta$ for $\eta>0$, the solution $\mathbf{f}$ is also purely imaginary and can be expressed as $\mathbf{f} = \imath [ \mathbf{g}^T, \mathbf{h}^T]^T$, where $\mathbf{g} \in\mathbb{R}^m$ and $\mathbf{h} \in\mathbb{R}^n$ are positive column vectors that uniquely satisfy the system of real-valued Dyson equations
\begin{equation}
    \eta  + \sum_{j=1}^n \widetilde{S}_{i,j} \mathbf{h}_j = \frac{1}{\mathbf{g}_i}, \qquad \qquad \eta  + \sum_{i=1}^m \mathbf{g}_i \widetilde{S}_{i,j}  = \frac{1}{\mathbf{h}_j}, \label{eq: real-valued Dyson}
\end{equation}
for all $i\in [m]$ and $j\in [n]$. Hence, under suitable conditions on the noise (described in Section~\ref{sec: method and guarantees}), the quantities $\operatorname{Im}\{R_{i,i}(\imath \eta)\}$ and $\operatorname{Im}\{R_{m+j,m+j}(\imath \eta)\}$ approximate $\mathbf{g}_i$ and $\mathbf{h}_j$, respectively, with high probability, for sufficiently large $m$ and $n$, where $\operatorname{Im}\{\cdot\}$ denotes the imaginary part of a complex number.

Under suitable conditions on the low-rank signal $X$ in~\eqref{eq: signal-plus-noise}, the same approximation extends to the analogous data resolvent, i.e., when the resolvent is constructed from the data instead of the noise~\cite{landa2025dyson}. Here, we define the data resolvent $\mathcal{R}(z)$ as
\begin{equation}
    \mathcal{R}(z) = (\widetilde{\mathcal{Y}} - z I)^{-1}, \qquad \qquad
    \widetilde{\mathcal{Y}} =
    \begin{bmatrix}
    \mathbf{0}_{m\times m} & \widetilde{Y} \\
    \widetilde{Y}^T & \mathbf{0}_{n\times n}
    \end{bmatrix}, \label{eq: definition of R_cal and Y_cal_tilde}
\end{equation}
and subsequently, the quantities $\operatorname{Im}\{\mathcal{R}_{i,i}(\imath \eta)\}$ and $\operatorname{Im}\{\mathcal{R}_{m+j,m+j}(\imath \eta)\}$ are expected to approximate $\mathbf{g}_i$ and $\mathbf{h}_j$, respectively, for sufficiently large $m$ and $n$. Consequently, the vectors $\mathbf{g}$ and $\mathbf{h}$ that solve the Dyson equations~\eqref{eq: real-valued Dyson} can be estimated from the data by
\begin{align}
    \hat{\mathbf{g}}_i = \operatorname{Im}\{\mathcal{R}_{i,i}(\imath \eta)\} &= \eta \left[ \left( \widetilde{Y} \widetilde{Y}^T + \eta^2 I_{m}  \right)^{-1}\right]_{i,i}, \label{eq: right resolvent formula} \\
    \hat{\mathbf{h}}_j = \operatorname{Im}\{\mathcal{R}_{m+j,m+j}(\imath \eta)\} &= \eta \left[ \left( \widetilde{Y}^T \widetilde{Y} + \eta^2 I_{n}  \right)^{-1} \right]_{j,j}, \label{eq: left resolvent formula}
\end{align}
where $[\cdot]_{i,j}$ denotes the $(i,j)$th entry of a matrix, and we used Lemma~\ref{lem: block resolvent formulas} in Appendix~\ref{appendix: block resolvent formulas} for the final formulas of the resolvent's main diagonal. The estimates $\hat{\mathbf{g}}$ and $\hat{\mathbf{h}}$ are accurate even if the signal $X$ contains arbitrarily strong low-rank factors, as long as they are not concentrated in very small subsets of entries; see Assumptions~\ref{assump: population bounds} and~\ref{assump: main assumption} for the required conditions and Lemma~\ref{lem:resolvent_supnorm_concentration} in Appendix~\ref{appendix: lem:resolvent_supnorm_concentration} for the resulting concentration of $(\hat{\mathbf{g}},\hat{\mathbf{h}})$ around $({\mathbf{g}},{\mathbf{h}})$.

Importantly, if the scaling factors $\mathbf{x}$ and $\mathbf{y}$ are chosen as the solutions to the scaling equations~\eqref{eq: scaling equations}, i.e., $(\mathbf{x},\mathbf{y}) = (\mathbf{x}^\star,\mathbf{y}^\star)$ (and thus $\widetilde{S} = \widetilde{S}^\star$), then $\mathbf{g}$ and $\mathbf{h}$ have explicit formulas given by
\begin{equation}
    \mathbf{g}^{\star}_i = \frac{1-\alpha}{\eta}, \qquad \qquad \mathbf{h}^{\star}_j = \frac{1-\alpha \frac{m}{n}}{\eta}, \label{eq: formulas for g_star and h_star}
    \end{equation}
for all $i\in [m]$ and $j\in [n]$, where
\begin{align}
    \alpha = \frac{1 + \frac{m}{n} + \eta^2 - \sqrt{\left( 1 + \frac{m}{n} + \eta^2 \right)^2 - 4 \frac{m}{n}}}{2\frac{m}{n}}.
    \label{eq: alpha def}
\end{align}
These formulas can be verified directly by substituting~\eqref{eq: formulas for g_star and h_star} into~\eqref{eq: real-valued Dyson} and using the fact that $\widetilde{S}^\star$ has constant row and column sums, in which case the equations~\eqref{eq: real-valued Dyson} are reduced to a quadratic scalar equation for $\alpha$; see also the proof of Lemma~\ref{lem: properties of potential function} in Appendix~\ref{appendix: proof of lemma on properties of potential fun}. Consequently, under the true equalization of the variance profile $S$, the vectors solving the Dyson equations have constant entries with known values. Since the vectors $(\mathbf{g},\mathbf{h})$ solving the Dyson equations in general can be estimated from the main diagonal of the data resolvent (as explained above), this enables a mechanism for blind equalization of the noise variances.

\section{The proposed method and theoretical guarantees} \label{sec: method and guarantees}

The main limitation of the approach described in Section~\ref{Sec: sinkhorn scaling the MP law} is that
the scaling equations~\eqref{eq: scaling equations} cannot be solved directly. In particular, each step of Sinkhorn's algorithm requires knowing the row (column) sums of the matrix $S$ after its columns (rows) have been normalized in the previous iteration to attain the prescribed sums. For example, the first iteration of Sinkhorn's algorithm requires knowing the sum of the noise variances $S_{i,j}$ across the rows or columns. These sums cannot be estimated trivially from the data for general noise distributions due to the presence of the low-rank signal matrix $X$, especially if the signal contains strong components that dominate the empirical variance of the data across the rows or columns.

To overcome this obstacle, we develop a different iterative procedure for scaling the rows and columns of $S$, such that the required scaling factors at each step can be accurately estimated from the data, particularly by exploiting the data resolvent and its relation to the Dyson equation as explained in Section~\ref{sec: resolvent and Dyson equation}. This new procedure is not equivalent to Sinkhorn's algorithm, but it has the same stationary points, which are the solutions to the scaling equations~\eqref{eq: scaling equations}. We first develop the procedure at the population level, assuming that the variance matrix $S$ is known, and then derive its data-driven version, where we estimate the quantities required for each iteration directly from the data.

\subsection{The population-level procedure} \label{sec: population-level procedure}
To find the scaling factors $(\mathbf{x}^\star,\mathbf{y}^\star)$ that solve~\eqref{eq: scaling equations}, we propose to minimize a certain potential function that augments the logarithmic barrier function~\eqref{eq: logarithmic barrier function}.
Specifically, consider the function
\begin{align}
\begin{aligned}
    J({\mathbf{x}},{\mathbf{y}},\mathbf{g},\mathbf{h}) &=  \mathbf{g}^T D\{\mathbf{{\mathbf{x}}}\} S D\{{\mathbf{y}}\} \mathbf{h} \\
    &+ \sum_{i=1}^m \left( \eta \mathbf{g}_i - \log (\mathbf{g}_i) - \alpha \log ({\mathbf{x}}_i) \right) +  \sum_{j=1}^n (\eta \mathbf{h}_j - \log (\mathbf{h}_j) - \alpha \frac{m}{n} \log ({\mathbf{y}}_j) ),
    \end{aligned}
    \label{eq: potential fun}
\end{align}
where the variables ${\mathbf{x}}\in\mathbb{R}^m$, ${\mathbf{y}}\in\mathbb{R}^n$, $\mathbf{g}\in\mathbb{R}^m$, ${\mathbf{h}}\in\mathbb{R}^n$ are positive column vectors, $\eta>0$ is the resolvent parameter from Section~\ref{sec: resolvent and Dyson equation}, and $\alpha$ is from~\eqref{eq: alpha def}.
The augmentation consists of introducing the auxiliary variables $\mathbf{g}$ and $\mathbf{h}$, which modify the bilinear term and contribute additional terms to the objective. If we fix $\mathbf{g}=\mathbf{g}^\star$ and $\mathbf{h}=\mathbf{h}^\star$ from~\eqref{eq: formulas for g_star and h_star}, then the function~\eqref{eq: potential fun} reduces to the logarithmic barrier function~\eqref{eq: logarithmic barrier function} up to additive constants and the global scalar $\alpha$ (since $\mathbf{g}^\star_i\mathbf{h}^\star_j=\alpha$; see the proof of Lemma~\ref{lem: properties of potential function} in Appendix~\ref{appendix: proof of lemma on properties of potential fun}).

A key property of the potential function~\eqref{eq: potential fun} is that its minimization with respect to the pair $(\mathbf{g},\mathbf{h})$ leads to the Dyson equations~\eqref{eq: real-valued Dyson}, whose solution can be estimated accurately from the data in a broad range of settings. Subsequently, minimization with respect to $\mathbf{x}$ or $\mathbf{y}$ can be evaluated using these estimates, leading to a practical alternating minimization procedure resembling Sinkhorn's algorithm (which minimizes the standard logarithmic barrier function).

The following lemma characterizes the critical points of the potential function~\eqref{eq: potential fun} and outlines several properties that are key in the derivation of our proposed procedure.
\begin{lem} \label{lem: properties of potential function}
    Let $(\mathbf{x}^\star,\mathbf{y}^\star)$ be an arbitrary pair of positive vectors that solve the scaling equations~\eqref{eq: scaling equations}. Then, a point $(\mathbf{x},\mathbf{y},\mathbf{g},\mathbf{h})$ is a critical point of $J$ from~\eqref{eq: potential fun} if and only if it is of the form \sloppy $(a \mathbf{x}^\star, a^{-1} \mathbf{y}^\star, \mathbf{g}^\star, \mathbf{h}^\star)$ for some $a>0$, where $\mathbf{g}^\star$ and $\mathbf{h}^\star$ are from~\eqref{eq: formulas for g_star and h_star}. Every critical point of $J$ is a global minimizer. Furthermore, we have the following:
\begin{enumerate}
    \item \label{lem: potential function lemm property 1}
    Let $\overline{\mathbf{x}}\in\mathbb{R}^m$ and $\overline{\mathbf{y}}\in\mathbb{R}^n$ be fixed positive vectors. Then, the function $J(\overline{\mathbf{x}},\overline{\mathbf{y}},\mathbf{g},\mathbf{h})$ admits a unique minimizer over the pair of positive vectors $(\mathbf{g},\mathbf{h})$. This minimizer, denoted by $(\overline{\mathbf{g}},\overline{\mathbf{h}})$, is the unique solution to the Dyson equations~\eqref{eq: real-valued Dyson} with $\mathbf{x} = \overline{\mathbf{x}}$ and $\mathbf{y} = \overline{\mathbf{y}}$, i.e.,
    \begin{equation}
        \eta + \sum_{j=1}^n \overline{\mathbf{x}}_i S_{i,j} \overline{\mathbf{y}}_j \overline{\mathbf{h}}_j = \frac{1}{\overline{\mathbf{g}}_i}, \qquad \qquad \eta + \sum_{i=1}^m \overline{\mathbf{g}}_i \overline{\mathbf{x}}_i S_{i,j} \overline{\mathbf{y}}_j = \frac{1}{\overline{\mathbf{h}}_j}, \label{eq: Dyson equation}
    \end{equation}
    for all $i\in [m]$ and $j\in [n]$, where $\overline{\mathbf{g}}_i < {1}/{\eta}$ and $\overline{\mathbf{h}}_j < {1}/{\eta}$.
    \item \label{lem: potential function lemm property 2} Let $\overline{\mathbf{x}}$, $\overline{\mathbf{y}}$, $\overline{\mathbf{g}}$, $\overline{\mathbf{h}}$ be as described in property~\ref{lem: potential function lemm property 1}. Then, $J({\mathbf{x}},\overline{\mathbf{y}},\overline{\mathbf{g}},\overline{\mathbf{h}})$ admits a unique minimizer over the positive vector $\mathbf{x}$, given by
    \begin{equation}
        \mathbf{x}_i = \overline{\mathbf{x}}_i \frac{\alpha }{1 - \eta \overline{\mathbf{g}}_i} , \label{eq: formula for x_prime update}
    \end{equation}
    for all $i\in [m]$.
    \item \label{lem: potential function lemm property 3} Let $\overline{\mathbf{x}}$, $\overline{\mathbf{y}}$, $\overline{\mathbf{g}}$, $\overline{\mathbf{h}}$ be as described in property~\ref{lem: potential function lemm property 1}. Then, $J(\overline{\mathbf{x}},{\mathbf{y}},\overline{\mathbf{g}},\overline{\mathbf{h}})$ admits a unique minimizer over the positive vector $\mathbf{y}$, given by
    \begin{equation}
        \mathbf{y}_j = \overline{\mathbf{y}}_j  \frac{\alpha \frac{m}{n}}{1 - \eta \overline{\mathbf{h}}_j} , \label{eq: formula for y_prime update}
    \end{equation}
    for all $j\in [n]$.
\end{enumerate}
\end{lem}
The proof of Lemma~\ref{lem: properties of potential function} can be found in Appendix~\ref{appendix: proof of lemma on properties of potential fun}. Lemma~\ref{lem: properties of potential function} shows that the only critical points of the potential function~\eqref{eq: potential fun}, in terms of the vectors $\mathbf{x}$ and $\mathbf{y}$, are those that coincide with the desired scaling factors $\mathbf{x}^\star$ and $\mathbf{y}^\star$ that solve the scaling equations~\eqref{eq: scaling equations}. These critical points are global minimizers of $J$. Hence, we propose to find $\mathbf{x}^\star$ and $\mathbf{y}^\star$ by minimizing~\eqref{eq: potential fun} over the four variables $\mathbf{x}$, $\mathbf{y}$, $\mathbf{g}$, and $\mathbf{h}$. To this end, properties~\ref{lem: potential function lemm property 1}--\ref{lem: potential function lemm property 3} in Lemma~\ref{lem: properties of potential function} suggest a natural procedure to iteratively minimize~\eqref{eq: potential fun}. Given any positive vectors $\overline{\mathbf{x}}$ and $\overline{\mathbf{y}}$, we first minimize $J$ over $(\mathbf{g},\mathbf{h})$ according to property~\ref{lem: potential function lemm property 1} (via the Dyson equations~\eqref{eq: Dyson equation}) and then minimize $J$ over $\mathbf{x}$ or $\mathbf{y}$ according to properties~\ref{lem: potential function lemm property 2} or~\ref{lem: potential function lemm property 3}, respectively, via the formulas~\eqref{eq: formula for x_prime update} or~\eqref{eq: formula for y_prime update}. Note that these formulas are well-posed since $\overline{\mathbf{g}}_i < {1}/{\eta}$ and $\overline{\mathbf{h}}_j < {1}/{\eta}$ according to property~\ref{lem: potential function lemm property 1}.

Specifically, we sequentially update four vectors: $\mathbf{x}^{(t)}\in\mathbb{R}^m$, $\mathbf{y}^{(t)}\in\mathbb{R}^n$, $\mathbf{g}^{(t)}\in\mathbb{R}^m$, and $\mathbf{h}^{(t)}\in\mathbb{R}^n$, where $t$ is an integer iteration index. For each iteration $t>0$, given the scaling vectors $\mathbf{x}^{(t-1)}$ and $\mathbf{y}^{(t-1)}$ from the previous iteration, we obtain the vectors $({\mathbf{g}^{(t)}},{\mathbf{h}^{(t)}})$ as the pair of positive vectors $(\overline{\mathbf{g}},\overline{\mathbf{h}})$ that solve the Dyson equations~\eqref{eq: Dyson equation} for $\overline{\mathbf{x}} = \mathbf{x}^{(t-1)}$ and $\overline{\mathbf{y}} = \mathbf{y}^{(t-1)}$.
After obtaining $\mathbf{g}^{(t)}$ and $\mathbf{h}^{(t)}$, we update either the row scaling vector $\mathbf{x}^{(t)}\in\mathbb{R}^m$ according to property~\ref{lem: potential function lemm property 2} in Lemma~\ref{lem: properties of potential function} while keeping the column scaling vector $\mathbf{y}^{(t)}\in\mathbb{R}^n$ unchanged, i.e.,
\begin{equation}
    \mathbf{x}^{(t)}_i = \mathbf{x}^{(t-1)}_i \frac{\alpha }{1 - \eta {\mathbf{g}}^{(t)}_i}, \qquad\qquad \mathbf{y}^{(t)}_j = \mathbf{y}^{(t-1)}_j, \label{eq: x_t update}
\end{equation}
or vice versa according to property~\ref{lem: potential function lemm property 3} in Lemma~\ref{lem: properties of potential function}, that is
\begin{equation}
    \mathbf{y}^{(t)}_j = \mathbf{y}^{(t-1)}_j \frac{\alpha \frac{m}{n}}{1 - \eta {\mathbf{h}}^{(t)}_j}, \qquad\qquad \mathbf{x}^{(t)}_i = \mathbf{x}^{(t-1)}_i, \label{eq: y_t update}
\end{equation}
for all $i\in [m]$ and $j\in [n]$, recalling that $\alpha$ was defined in~\eqref{eq: alpha def}.

To determine whether to update the row scaling vector according to~\eqref{eq: x_t update} or the column scaling vector according to~\eqref{eq: y_t update}, we identify which of these two updates reduces the potential function~\eqref{eq: potential fun} more. To this end, we show in Appendix~\ref{appendix: derivation of J_t formula} that the following identity holds for a normalized version of the potential function:
\begin{align}
    J_t &:= \frac{J(\mathbf{x}^{(t)},\mathbf{y}^{(t)},\mathbf{g}^{(t)},\mathbf{h}^{(t)})}{m} - \alpha \nonumber \\
    &= \frac{1}{m}\sum_{i=1}^m \left( \eta \mathbf{g}^{(t)}_i - \log (\mathbf{g}^{(t)}_i) - \alpha \log (\mathbf{x}^{(t)}_i) \right) + \frac{1}{m} \sum_{j=1}^n \left(\eta \mathbf{h}^{(t)}_j - \log (\mathbf{h}^{(t)}_j) - \alpha \frac{m}{n} \log (\mathbf{y}^{(t)}_j) \right). \label{eq: normalized potential fun}
\end{align}
According to~\eqref{eq: normalized potential fun}, updating the row scaling vector (via~\eqref{eq: x_t update}) will decrease $J_t$ more than updating the column scaling vector (via~\eqref{eq: y_t update}) if and only if
\begin{equation}
    \frac{1}{m} \sum_{i=1}^m \log \left( \frac{\alpha }{1 - \eta {\mathbf{g}}^{(t)}_i} \right) > \frac{1}{n} \sum_{j=1}^n \log \left( \frac{\alpha \frac{m}{n}}{1 - \eta {\mathbf{h}}^{(t)}_j} \right). \label{eq: condition for updating row scaling vector}
\end{equation}
Hence, we use~\eqref{eq: x_t update} if~\eqref{eq: condition for updating row scaling vector} holds and use~\eqref{eq: y_t update} otherwise.

Overall, we derived a procedure for iterative minimization of the potential function $J$ by the sequence of updates
\[
(\mathbf{x}^{(t)},\mathbf{y}^{(t)}) \rightarrow (\mathbf{g}^{(t+1)},\mathbf{h}^{(t+1)}) \rightarrow (\mathbf{x}^{(t+1)},\mathbf{y}^{(t+1)}),
\]
where we first obtain $(\mathbf{g}^{(t+1)},\mathbf{h}^{(t+1)})$ from $(\mathbf{x}^{(t)},\mathbf{y}^{(t)})$ by solving the Dyson equations, and then obtain $(\mathbf{x}^{(t+1)},\mathbf{y}^{(t+1)})$ from $(\mathbf{x}^{(t)},\mathbf{y}^{(t)},\mathbf{g}^{(t+1)},\mathbf{h}^{(t+1)})$ using either~\eqref{eq: x_t update} or~\eqref{eq: y_t update}, depending on the branching condition~\eqref{eq: condition for updating row scaling vector}.
We refer to this procedure as the \textit{population-level procedure}.

To quantify the convergence rate, we consider the following boundedness conditions on the variance profile and the aspect ratio.
\begin{assump}\label{assump: population bounds}
We have the following:
\begin{enumerate}
    \item \textbf{Bounded aspect ratio.} There exist global constants $\gamma_-,\gamma_+\in(0,1]$ such that $\gamma_-\le m/n\le\gamma_+$. \label{assump: bounded aspet ratio}
    \item \textbf{Variance bounds.} There exist global constants $0<c\le C<\infty$ such that
    \begin{equation}\label{eq: uniform variance bounds}
        \frac{c}{n}\le S_{ij}\le\frac{C}{n}
        \qquad\text{for all }i\in[m],\ j\in[n].
    \end{equation}
    \label{assump: variance boundedness}
\end{enumerate}
\end{assump}
These conditions allow general positive structure in the variance profile $S$ while controlling its overall scale and the ratio between its largest and smallest entries. The conditions are needed only to establish a convergence rate that is uniform over the matrix dimensions and variance profiles; convergence itself holds for every positive $S$ (see part~\ref{thm: population qualitative} of Theorem~\ref{thm:population_convergence} below).

To account for the scalar ambiguity in the scaling equations~\eqref{eq: scaling equations}, we measure the discrepancy between any two positive scaling pairs by
\begin{equation}\label{eq: averaged logarithmic distance}
\begin{aligned}
d\bigl((\mathbf x,\mathbf y),(\mathbf u,\mathbf v)\bigr)
:= \sqrt{\inf_{a>0}\left\{
\frac1m\left\|\log\!\left(\frac{a\mathbf x}{\mathbf u}\right)\right\|_2^2
+\frac1n\left\|\log\!\left(\frac{a^{-1}\mathbf y}{\mathbf v}\right)\right\|_2^2
\right\}},
\end{aligned}
\end{equation}
where logarithms and divisions are taken entrywise. The logarithms measure multiplicative discrepancies in the scaling factors, and the infimum removes the trivial scalar ambiguity.

We now have the following theorem, which establishes convergence of the population-level procedure and, under Assumption~\ref{assump: population bounds}, a geometric convergence rate. Its proof is given in Appendix~\ref{appendix:proof_population_convergence}.

\begin{thm}
\label{thm:population_convergence}
Let $\{(\mathbf{x}^{(t)},\mathbf{y}^{(t)})\}_{t\ge0}$ be generated by the population-level procedure with $\eta>0$, and let $(\mathbf{x}^\star,\mathbf{y}^\star)$ solve the scaling equations~\eqref{eq: scaling equations}.
\begin{enumerate}[label=(\roman*)]
    \item \label{thm: population qualitative}
    For every positive variance profile $S$ and positive initialization $(\mathbf{x}^{(0)},\mathbf{y}^{(0)})$, we have $J_{t+1}\le J_t$ for all $t\ge1$, with strict inequality whenever $(\mathbf{x}^{(t)},\mathbf{y}^{(t)})$ is not a solution of~\eqref{eq: scaling equations}. Moreover, there exist scalars $a_t>0$ such that
    \[
    (a_t\mathbf{x}^{(t)},a_t^{-1}\mathbf{y}^{(t)})
    \longrightarrow(\mathbf{x}^\star,\mathbf{y}^\star)
    \qquad\text{as }t\to\infty.
    \]
    \item \label{thm: population quantitative}
    Suppose, in addition, that Assumption~\ref{assump: population bounds} holds and $\mathbf{x}^{(0)}=\mathbf1_m$, $\mathbf{y}^{(0)}=\mathbf1_n$. Define
    \begin{equation}\label{eq: optimal normalized potential}
        J_\star:=\frac1m J(\mathbf{x}^\star,\mathbf{y}^\star,\mathbf{g}^\star,\mathbf{h}^\star)-\alpha.
    \end{equation}
    There exist constants $K>0$ and $\rho\in(0,1)$, depending only on $\eta,c,C,\gamma_-,\gamma_+$, such that, for all $t\ge1$,
    \begin{equation}\label{eq: population geometric rates}
    \begin{aligned}
        0\le J_t-J_\star&\le K\rho^{2t}, \qquad
        d\bigl((\mathbf{x}^{(t)},\mathbf{y}^{(t)}),
               (\mathbf{x}^\star,\mathbf{y}^\star)\bigr)&\le K\rho^t.
    \end{aligned}
    \end{equation}
\end{enumerate}
\end{thm}

Part~\ref{thm: population qualitative} establishes convergence for arbitrary positive variance profiles $S$. Part~\ref{thm: population quantitative} shows that, under the boundedness requirements of Assumption~\ref{assump: population bounds}, the potential gap $J_t - J_\star$ and the error in the scaling factors decrease geometrically, with constants independent of the dimensions and the particular variance pattern. The same rate holds for any positive initialization whose entries are bounded above and below by global constants.

\subsection{The data-driven procedure} \label{sec: data-driven procedure}
A key aspect of the population-level procedure is that the update mechanism of $\mathbf{x}^{(t)}$ and $\mathbf{y}^{(t)}$ and the normalized potential $J_t$ depends only on the vectors $\mathbf{g}^{(t)}$ and $\mathbf{h}^{(t)}$, which in turn solve the Dyson equations~\eqref{eq: Dyson equation} for $\overline{\mathbf{x}} = {\mathbf{x}^{(t-1)}}$ and $\overline{\mathbf{y}} = {\mathbf{y}^{(t-1)}}$. Therefore, the pair $(\mathbf{g}^{(t)},\mathbf{h}^{(t)})$ can be estimated from the resolvent of the scaled data matrix $\widetilde{Y} =( D\{\mathbf{x}^{(t-1)}\} )^{1/2} Y (D\{\mathbf{y}^{(t-1)}\})^{1/2}$ as explained in Section~\ref{sec: resolvent and Dyson equation}, using the formulas~\eqref{eq: right resolvent formula} and~\eqref{eq: left resolvent formula} with $\mathbf{x} = \mathbf{x}^{(t-1)}$ and $\mathbf{y} = \mathbf{y}^{(t-1)}$. Since $\mathbf{x}^{(t-1)}$ and $\mathbf{y}^{(t-1)}$ cannot be computed directly if $S$ is unknown, we replace them with their estimated counterparts in each iteration, which are updated sequentially using the formulas of the population-level procedure. Importantly, the Dyson equations do not need to be solved explicitly at any point of our data-driven procedure---we only need the diagonal entries of the data resolvent.

Concretely, suppose that for some iteration $t \geq 1$, we have estimates of $\mathbf{x}^{(t-1)}$ and $\mathbf{y}^{(t-1)}$, denoted by $\hat{\mathbf{x}}^{(t-1)}$ and $\hat{\mathbf{y}}^{(t-1)}$, respectively. The corresponding normalized data matrix for iteration $t-1$ is
\begin{equation}
    \hat{Y}^{(t-1)} = \left( D\{\hat{\mathbf{x}}^{(t-1)}\} \right)^{1/2} Y  \left( D\{\hat{\mathbf{y}}^{(t-1)}\} \right)^{1/2}. \label{eq: Y_hat def}
\end{equation}
We then estimate the pair $(\mathbf{g}^{(t)},\mathbf{h}^{(t)})$ by $(\hat{\mathbf{g}}^{(t)},\hat{\mathbf{h}}^{(t)})$, which is defined via~\eqref{eq: right resolvent formula} and~\eqref{eq: left resolvent formula} when replacing $\widetilde{Y}$ with $\hat{Y}^{(t-1)}$, i.e.,
\begin{align}
    \hat{\mathbf{g}}_i^{(t)} &= \eta \left[ \left( \hat{Y}^{(t-1)} \left(\hat{Y}^{(t-1)}\right)^T + \eta^2 I_{m}  \right)^{-1}\right]_{i,i} \label{eq: g_hat formula}, \\
    \hat{\mathbf{h}}_j^{(t)} &= \eta \left[ \left( \left( \hat{Y}^{(t-1)} \right)^T \hat{Y}^{(t-1)} + \eta^2 I_{n}  \right)^{-1} \right]_{j,j}, \label{eq: h_hat formula}
\end{align}
for all $i\in [m]$ and $j\in [n]$.
Subsequently, we compute $\hat{\mathbf{x}}^{(t)}$ and $\hat{\mathbf{y}}^{(t)}$ according to~\eqref{eq: x_t update} and~\eqref{eq: y_t update}, respectively, when replacing ${\mathbf{x}}^{(t-1)}$, ${\mathbf{y}}^{(t-1)}$, ${\mathbf{g}}^{(t)}$, ${\mathbf{h}}^{(t)}$ in these equations with $\hat{\mathbf{x}}^{(t-1)}$, $\hat{\mathbf{y}}^{(t-1)}$, $\hat{\mathbf{g}}^{(t)}$, $\hat{\mathbf{h}}^{(t)}$, respectively. Similarly, we define $\hat{J}_t$ analogously to $J_t$ from~\eqref{eq: normalized potential fun} and use the condition in~\eqref{eq: condition for updating row scaling vector} to determine whether to update the estimated row scaling vector $\hat{\mathbf{x}}^{(t)}$ or the estimated column scaling vector $\hat{\mathbf{y}}^{(t)}$, when replacing ${\mathbf{x}}^{(t)}$, ${\mathbf{y}}^{(t)}$, ${\mathbf{g}}^{(t)}$, ${\mathbf{h}}^{(t)}$ with $\hat{\mathbf{x}}^{(t)}$, $\hat{\mathbf{y}}^{(t)}$, $\hat{\mathbf{g}}^{(t)}$, $\hat{\mathbf{h}}^{(t)}$, respectively.

For $t=0$, the vectors $\hat{\mathbf{x}}^{(0)}$ and $\hat{\mathbf{y}}^{(0)}$ are typically initialized as the all-ones vectors. Alternatively, they can also be initialized via other methods discussed in the introduction (e.g.,~\cite{landa2025dyson,landa2022biwhitening}), which may provide useful starting points and expedite the convergence in practice, even if the assumptions of those methods do not strictly hold. After scaling the rows and columns of the data with the initial scaling factors $\hat{\mathbf{x}}^{(0)}$ and $\hat{\mathbf{y}}^{(0)}$, and before proceeding with the proposed iterative procedure, we perform a one-time scalar normalization where we divide the matrix by its median singular value. This scalar normalization ensures, under mild conditions, that the global scale of the noise conforms to the theoretical requirements of the Dyson equation and our subsequent guarantees, in particular that the variances $S_{ij}$ are of the scale $1/n$.
The resolvent parameter $\eta>0$ sets the spectral scale at which the resolvent is evaluated. Although the population-level procedure targets the same noise equalizer for every fixed $\eta>0$, its convergence rate and finite-sample estimation accuracy may depend on this choice. We use $\eta=1$ by default: after the median-singular-value normalization, this places $\eta$ on the scale of a typical singular value of the data. This choice is a practical convention throughout all of our experiments; we discuss its role further in Section~\ref{sec: discussion}.

Note that for the data-driven procedure, $Y$ should not have any rows or columns that are entirely zero, as otherwise the corresponding entries in $\hat{\mathbf{g}}^{(t)}$ and $\hat{\mathbf{h}}^{(t)}$ would be exactly $1/\eta$, in which case the formulas in~\eqref{eq: x_t update} and~\eqref{eq: y_t update} are ill-posed. As long as $Y$ does not have any zero rows and columns, the entries of $\hat{\mathbf{g}}^{(t)}$ and $\hat{\mathbf{h}}^{(t)}$ are guaranteed to be positive and smaller than $1/\eta$. Overall, our proposed data-driven procedure is summarized in Algorithm~\ref{alg:BRINE}.

\begin{algorithm}
\caption{BRINE (Blind Resolvent-based Iterative Noise Equalizer)}\label{alg:BRINE}
\begin{algorithmic}[1]
\Statex{\textbf{Input:} Data matrix $Y\in \mathbb{R}^{m\times n}$ with $m\leq n$ and no rows or columns that are entirely zero; Initial scaling factors $\hat{\mathbf{x}}^{(0)}$ and $\hat{\mathbf{y}}^{(0)}$ (default: $\hat{\mathbf{x}}^{(0)} = \mathbf{1}_m$, $\hat{\mathbf{y}}^{(0)} = \mathbf{1}_n$); Resolvent parameter $\eta>0$ (default: $\eta = 1$)}; and maximum number of iterations $T$.
\State \label{alg:step 5}\textbf{Initialize:} $\overline{Y}^{(0)} = \left(D\left\{ \hat{\mathbf{x}}^{(0)} \right\}\right)^{1/2} Y \left(D\left\{ \hat{\mathbf{y}}^{(0)} \right\}\right)^{1/2}$; $\hat{Y}^{(0)} = \overline{Y}^{(0)}/\operatorname{Median}\{\sigma_1,\ldots,\sigma_m\}$, where $\sigma_1,\ldots,\sigma_m$ are the singular values of $\overline{Y}^{(0)}$.
\State For $t = 1,\ldots,T$ do
\begin{enumerate}[label=(\alph*)]
        \item Compute $\hat{\mathbf{g}}^{(t)}\in\mathbb{R}^m$ and $\hat{\mathbf{h}}^{(t)}\in\mathbb{R}^n$ from $\hat{Y}^{(t-1)}$ according to~\eqref{eq: g_hat formula} and~\eqref{eq: h_hat formula}, respectively.
        \item Compute $\mathbf{x}^{\text{update}} \in \mathbb{R}^m$ and $\mathbf{y}^{\text{update}} \in \mathbb{R}^n$ according to
        \begin{equation} \label{eq: empirical update formulas for x and y}
            \mathbf{x}_i^{\text{update}} = \frac{\alpha }{1 - \eta \hat{\mathbf{g}}^{(t)}_i}, \qquad\qquad \mathbf{y}_j^{\text{update}} = \frac{\alpha \frac{m}{n}}{1 - \eta \hat{\mathbf{h}}^{(t)}_j},
        \end{equation}
        for all $i\in [m]$ and $j\in [n]$, where $\alpha$ is defined in~\eqref{eq: alpha def}.
        \item If $\frac{1}{m} \sum_{i=1}^m \log(\mathbf{x}_i^{\text{update}}) > \frac{1}{n}\sum_{j=1}^n \log (\mathbf{y}_j^{\text{update}})$, update the row scaling factors:
        \begin{equation}
            \hat{\mathbf{x}}^{(t)} = \hat{\mathbf{x}}^{(t-1)} \circ \mathbf{x}^{\text{update}}, \qquad \hat{\mathbf{y}}^{(t)} = \hat{\mathbf{y}}^{(t-1)}, \qquad \hat{Y}^{(t)} = \left(D\{\mathbf{x}^{\text{update}}\}\right)^{1/2} \hat{Y}^{(t-1)},
        \end{equation}
        where $\circ$ is the Hadamard (element-wise) product. Otherwise, update the column factors:
        \begin{equation}
            \hat{\mathbf{y}}^{(t)} = \hat{\mathbf{y}}^{(t-1)} \circ \mathbf{y}^{\text{update}}, \qquad \hat{\mathbf{x}}^{(t)} = \hat{\mathbf{x}}^{(t-1)}, \qquad \hat{Y}^{(t)} = \hat{Y}^{(t-1)} \left(D\{\mathbf{y}^{\text{update}}\}\right)^{1/2}.
        \end{equation}
        \item Estimate the normalized potential function at iteration $t$ as
        \begin{equation}
            \hat{J}_t = \frac{1}{m}\sum_{i=1}^m \left( \eta \hat{\mathbf{g}}^{(t)}_i - \log (\hat{\mathbf{g}}^{(t)}_i) - \alpha \log (\hat{\mathbf{x}}^{(t)}_i) \right) + \frac{1}{m} \sum_{j=1}^n \left(\eta \hat{\mathbf{h}}^{(t)}_j - \log (\hat{\mathbf{h}}^{(t)}_j) - \alpha \frac{m}{n} \log (\hat{\mathbf{y}}^{(t)}_j) \right).
        \end{equation}
        \item If $t>1$ and $\hat{J}_t \geq \hat{J}_{t-1}$, stop the procedure and return $\hat{\mathbf{x}}^{(t-1)}$, $\hat{\mathbf{y}}^{(t-1)}$, $\hat{Y}^{(t-1)}$, $\hat{J}_{t-1}$. \label{alg: early stopping}
    \end{enumerate}
\State Return $\hat{\mathbf{x}}^{(t)}$, $\hat{\mathbf{y}}^{(t)}$, $\hat{Y}^{(t)}$, $\hat{J}_{t}$.
\end{algorithmic}
\end{algorithm}

Observe that according to the resulting data-driven update formulas~\eqref{eq: empirical update formulas for x and y}, the empirical procedure reaches saturation when $\mathbf{x}_i^{\text{update}} \approx 1$ and $\mathbf{y}_j^{\text{update}} \approx 1$ for all $i\in[m]$ and $j\in [n]$, or equivalently,
\begin{equation}
    \hat{\mathbf{g}}^{(t)} \approx \frac{1-\alpha}{\eta} \mathbf{1}_m = \mathbf{g}^{\star} , \qquad \hat{\mathbf{h}}^{(t)} \approx \frac{1-\alpha\frac{m}{n}}{\eta} \mathbf{1}_n = \mathbf{h}^{\star},
\end{equation}
where $(\mathbf{g}^{\star},\mathbf{h}^{\star})$ are from~\eqref{eq: formulas for g_star and h_star} and correspond to the solution of the Dyson equation under true equalization. Therefore, the data-driven procedure can be interpreted as an iterative rescaling procedure for equalizing the main diagonal entries of the data resolvent, i.e., trying to match them to their expected population values~\eqref{eq: formulas for g_star and h_star} under true variance equalization.

We now proceed to analyze the behavior of the data-driven procedure. Let us denote the (compact) SVD of the signal matrix $X$ as
\begin{equation}
    X = \widetilde{U}  D \{s_1,\ldots,s_r\} \widetilde{V}^T, \label{eq: SVD of X}
\end{equation}
where $\widetilde{U}\in\mathbb{R}^{m\times r}$, $\widetilde{V}\in\mathbb{R}^{n\times r}$, and $ s_1\geq \ldots \geq s_r$ stand for the left singular vectors, right singular vectors, and singular values of $X$, respectively. We also define the quantity
\begin{equation}
    \kappa_n = \max\left\{\max_{i\in [m]}\left\{ \sum_{k=1}^r \widetilde{U}_{ik}^2\right\}, \max_{j\in [n]}\left\{\sum_{k=1}^r \widetilde{V}_{jk}^2\right\}\right\},
\end{equation}
which is related to the localization level of the signal's singular vectors. For instance, if one of the singular vectors is supported on a finite number of entries independently of $m$ or $n$, then $\kappa_n$ is of the order of a constant and is not decaying with $m$ or $n$, indicating high localization. On the other extreme, we have $\kappa_n \sim 1/m$ when all singular vectors are fully delocalized (all entries have comparable magnitudes) and the rank $r$ is fixed.

We now describe the additional conditions on the noise and signal required for the data-driven guarantees.
\begin{assump}\label{assump: main assumption}
We have the following:
\begin{enumerate}
    \item \textbf{Moment upper bounds.} There exist global constants $\{\mu_q\}_{q=1}^\infty$ such that $\mathbb{E}|\sqrt n E_{ij}|^q\le\mu_q$ for all $i\in[m]$, $j\in[n]$, and $q\in\mathbb N$. \label{assump: noise moment bound}
    \item \textbf{Bounded rank growth.} There exist global constants $\widetilde C_0>0$ and $\delta_0\in(0,1/2]$ such that $r\le\widetilde C_0 n^{1/2-\delta_0}$. \label{assump: rank growth rate}
    \item \textbf{Power-law singular vector delocalization.} There exist global constants $\widetilde C_1,\delta_1>0$ such that $\kappa_n\le\widetilde C_1 n^{-\delta_1}$. \label{assump: singular vector localization}
\end{enumerate}
\end{assump}

The first condition in Assumption~\ref{assump: main assumption} requires uniform bounds on all moments of $\sqrt{n}E_{ij}$,
thereby excluding very heavy-tailed noise while still allowing
a broad class of non-Gaussian noise distributions. Its case $q=2$ also implies the upper variance bound in~\eqref{eq: uniform variance bounds}, with $C=\mu_2$.
The next two conditions control the complexity of the low-rank signal. Assumption~\ref{assump: main assumption}.\ref{assump: rank growth rate} allows the signal rank to grow with the dimension, but requires it to remain $o(\sqrt n)$. Assumption~\ref{assump: main assumption}.\ref{assump: singular vector localization} is a power-law delocalization requirement on the singular vectors. It can be interpreted as requiring that the singular-vector matrices $\widetilde U$ and $\widetilde V$ are not too concentrated across their rows. In particular, it rules out situations in which the signal matrix $X$ is localized on only a fixed number of rows or columns, or even on only logarithmically many rows or columns. At the same time, the condition still permits substantial sparsity: the signal may be supported on a vanishing fraction of coordinates, provided that this support remains sufficiently large in a polynomial sense and that the rank does not grow too quickly. For example, when the rank is fixed, the singular vectors may be supported on a fraction $n^{-1/4}$ of the coordinates, as long as their mass is spread reasonably uniformly over that support.

We now have the following theorem, which quantifies the error of the estimated scaling factors $(\hat{\mathbf x}^{(t)},\hat{\mathbf y}^{(t)})$ for each iteration $t$ relative to the desired factors $(\mathbf x^\star,\mathbf y^\star)$, using the same logarithmic error as in Theorem~\ref{thm:population_convergence}. For this result we omit the preliminary median normalization and the stopping rule in Algorithm~\ref{alg:BRINE}, so that the iterates are defined through a prescribed number of steps.

\begin{thm}\label{thm: main theoretical guarantees}
Suppose that Assumptions~\ref{assump: population bounds}
and~\ref{assump: main assumption} hold and
$\hat{\mathbf x}^{(0)}=\mathbf1_m$,
$\hat{\mathbf y}^{(0)}=\mathbf1_n$.
Let $(\mathbf x^\star,\mathbf y^\star)$ solve the scaling
equations~\eqref{eq: scaling equations}.
Fix $A>0$, $\varepsilon\in(0,\delta_0)$, and $\tau>0$, and define
\begin{equation}\label{eq: statistical error and iteration horizon}
    \omega_n:=n^{\varepsilon-1/2}+\kappa_n,
    \qquad
    T_n:=\lfloor A\log n\rfloor.
\end{equation}
There exist constants $K_1,K_2>0$ and $\widetilde\rho\in(0,1)$
such that, for all sufficiently large $m,n$, with probability
at least $1-K_2n^{-\tau}$, simultaneously for $0\le t\le T_n$,
\begin{equation}\label{eq: total scaling error}
    d\bigl((\hat{\mathbf x}^{(t)},\hat{\mathbf y}^{(t)}),
           (\mathbf x^\star,\mathbf y^\star)\bigr)
    \le K_1\widetilde\rho^{\,t}+K_2t\omega_n.
\end{equation}
Here $K_1$ and $\widetilde\rho$ depend only on
$\eta,c,C,\gamma_-,\gamma_+$, while $K_2$ and the required lower
bound on the dimensions may additionally depend on
$A,\varepsilon,\tau$ and the global constants in
Assumption~\ref{assump: main assumption}.
\end{thm}

The first term in~\eqref{eq: total scaling error} stems from the convergence of the population-level procedure, with a possibly slower rate than in Theorem~\ref{thm:population_convergence} to account for differences in the row and column update choices made in the empirical procedure. The second term bounds the accumulated statistical error, which depends on the matrix dimensions and the localization level $\kappa_n$ of the signal singular vectors, but does not require any bound on the signal singular values. Under Assumption~\ref{assump: main assumption}.\ref{assump: singular vector localization}, we have $\omega_n\le n^{\varepsilon-1/2}+\widetilde C_1n^{-\delta_1}$. Taking $t=T_n$ therefore makes both terms in~\eqref{eq: total scaling error} vanish as $m,n\to\infty$, requiring only a logarithmic number of iterations in $n$. This suggests that a relatively small number of iterations may be enough to achieve satisfactory performance in practice, and also motivates the stopping rule in Step~\ref{alg: early stopping} of Algorithm~\ref{alg:BRINE}. Namely, if the estimated potential \(\hat J_t\) becomes non-decreasing (recalling that the population potential $J_t$ must decrease), it may indicate that the number of iterations is already too large for the finite-dimensional empirical iterates to remain well approximated by the corresponding asymptotic population-level procedure. The proof of Theorem~\ref{thm: main theoretical guarantees} is given in Appendix~\ref{appendix:proof_thm_main}.

\subsection{Signal enhancement via noise equalization} \label{sec: signal enhancement}
Let $(\mathbf{x}^\star,\mathbf{y}^\star)$ denote a solution of the
scaling equations~\eqref{eq: scaling equations} and consider the normalized data~\eqref{eq: Y_tilde def} using the scaling factors $(\mathbf{x},\mathbf{y}) = (\mathbf{x}^\star,\mathbf{y}^\star)$. Hence, the scaled signal matrix is given by
\begin{equation}
\widetilde X
=
\bigl(D\{\mathbf{x}^\star\}\bigr)^{1/2}
X
\bigl(D\{\mathbf{y}^\star\}\bigr)^{1/2}. \label{eq: signal scaling with true scaling factors}
\end{equation}
Throughout this section, we assume that the average entry of the variance matrix $S$ is $1/n$, i.e.,
\begin{equation}
    \sum_{i=1}^m\sum_{j=1}^n S_{i,j}=m. \label{eq: average noise variance assumption}
\end{equation}
This assumption is non-restrictive and only meant to fix the global noise scale. According to the scaling equations~\eqref{eq: scaling equations}, the total variance of the noise is preserved under the equalizing transformation, since $\sum_{i=1}^m\sum_{j=1}^n \widetilde S_{i,j}=m$. Therefore, the proposed noise equalization acts only by redistributing the noise variance across rows and columns, rather than by changing its overall level.

Let $L$ be the logarithmic barrier function from~\eqref{eq: logarithmic barrier function}, and define the potential gap
\begin{equation}
\Delta(S)
:= \frac{1}{m}\Big( L(\mathbf{1}_m,\mathbf{1}_n)-L(\mathbf{x}^\star,\mathbf{y}^\star)\Big).
\label{eq: Delta S def}
\end{equation}
Since $(\mathbf{x}^\star,\mathbf{y}^\star)$ is a global minimizer of $L$, we have $\Delta(S)\ge 0$.
We also define the row and column variance sums
\begin{equation}
r_i := \sum_{j=1}^n S_{i,j},
\qquad
c_j := \sum_{i=1}^m S_{i,j}.
\label{eq: row col sums def}
\end{equation}
Note that if $S$ already satisfies the scaling equations~\eqref{eq: scaling equations} with $\mathbf{x}^\star = \mathbf{1}_m$ and $\mathbf{y}^\star = \mathbf{1}_n$, in which case $\Delta(S) = 0$, then $r_i = 1$ and $c_j = m/n$ for all $i\in [m]$ and $j\in [n]$.

We now have the following theorem, whose proof can be found in Appendix~\ref{appendix: proof of theorem on singular value gain}.

\begin{thm}
\label{thm: generic singular value gain}
Let the SVD of \(X\) be as in~\eqref{eq: SVD of X} with fixed rank \(r\), where
\(\widetilde U\in\mathbb{R}^{m\times r}\) and \(\widetilde V\in\mathbb{R}^{n\times r}\)
are independent Haar-distributed random matrices with orthonormal columns, while the noise variance matrix \(S\)
and the signal singular values \(s_1,\ldots,s_r>0\) are fixed. In addition, suppose that~\eqref{eq: average noise variance assumption} holds, $m/n \geq \gamma_{-}$ for a constant $\gamma_{-} >0$, and there exist constants \(0<c\le C<\infty\) such that
\[
\frac{c}{n}\le S_{ij}\le \frac{C}{n}
\qquad \forall i\in[m],\ \forall j\in[n].
\]
Then, for every fixed \(k\in[r]\), almost surely,
\begin{equation}
\liminf_{m,n\to\infty}
\frac{\sigma_k(\widetilde X)}{s_k \, e^{\Delta(S)/2}}
\ge 1,
\label{eq: liminf singular value lower bound Delta}
\end{equation}
where $\sigma_k(\widetilde X)$ is the $k$'th singular value of $\widetilde X$ and $\Delta(S)$ is defined in~\eqref{eq: Delta S def}. Moreover,
\begin{equation}
    e^{\Delta(S)/2}
\ge
\sqrt{\max\left\{ \left( \prod_{i=1}^m r_i \right)^{-1/m}, \left( \prod_{j=1}^n \left( \frac{n}{m} \right) c_j \right)^{-1/n} \right\} }
\ge 1. \label{eq: explicit lower bound for singular value growth}
\end{equation}
\end{thm}
Theorem~\ref{thm: generic singular value gain} shows that, for generic singular vectors that are independent of the variance profile $S$, noise equalization asymptotically cannot decrease the singular values of the signal. More precisely, for every fixed $k\in[r]$, the scaled singular value $\sigma_k(\widetilde X)$ is asymptotically bounded below by $s_k e^{\Delta(S)/2}$ up to a vanishing relative error. In particular, if the potential gap $\Delta(S)$ from~\eqref{eq: Delta S def} is bounded away from zero asymptotically, then the noise equalization amplifies each signal singular value. The larger the potential gap $\Delta(S)$ is, the stronger the guaranteed amplification.

Moreover,~\eqref{eq: explicit lower bound for singular value growth} gives an explicit lower bound on this gain in terms of the row sums $r_i$ and column sums $c_j$ from~\eqref{eq: row col sums def}. In particular, the guaranteed amplification increases as the geometric means of these normalized row and column variance sums decrease, that is, as the marginal noise variances become more unevenly distributed across rows and/or columns. Hence, stronger heterogeneity in the total row/column variances leads to a larger guaranteed increase in the signal singular values after equalization. In other words, more severe heteroskedasticity across rows/columns leads to increased amplification of the signal singular values after equalization.

\section{Experiments}\label{sec:experiments}

\subsection{Recovery of the true scaling factors}\label{sec:experiments-convergence}
We first examine the accuracy of the estimated scaling factors in a controlled simulation where the true variance profile $S$ is known. We use the same signal-plus-noise generative model as in Figure~\ref{fig:stylized_intro}, with a $2\times 2$ block variance pattern, $20$ scattered high-variance rows and columns, and a rank-$20$ signal with ten strong and ten weak components (see Appendix~\ref{app:stylized_simulation}). We measure recovery using the logarithmic distance $d$ from~\eqref{eq: averaged logarithmic distance}, which accounts for the scalar ambiguity of the scaling factors. The true scaling factors $(\mathbf{x}^\star,\mathbf{y}^\star)$ are computed by applying Sinkhorn's algorithm to the true variance profile. We compare the data-driven BRINE algorithm with its population-level counterpart, computed using the true variance profile $S$ and the corresponding Dyson equations, as well as the rank-one Dyson equalizer of~\cite{landa2025dyson} and the empirical-variance equalizer of~\cite{leeb2021matrix}.

\begin{figure}[t]
    \centering
    \includegraphics[width=1\linewidth]{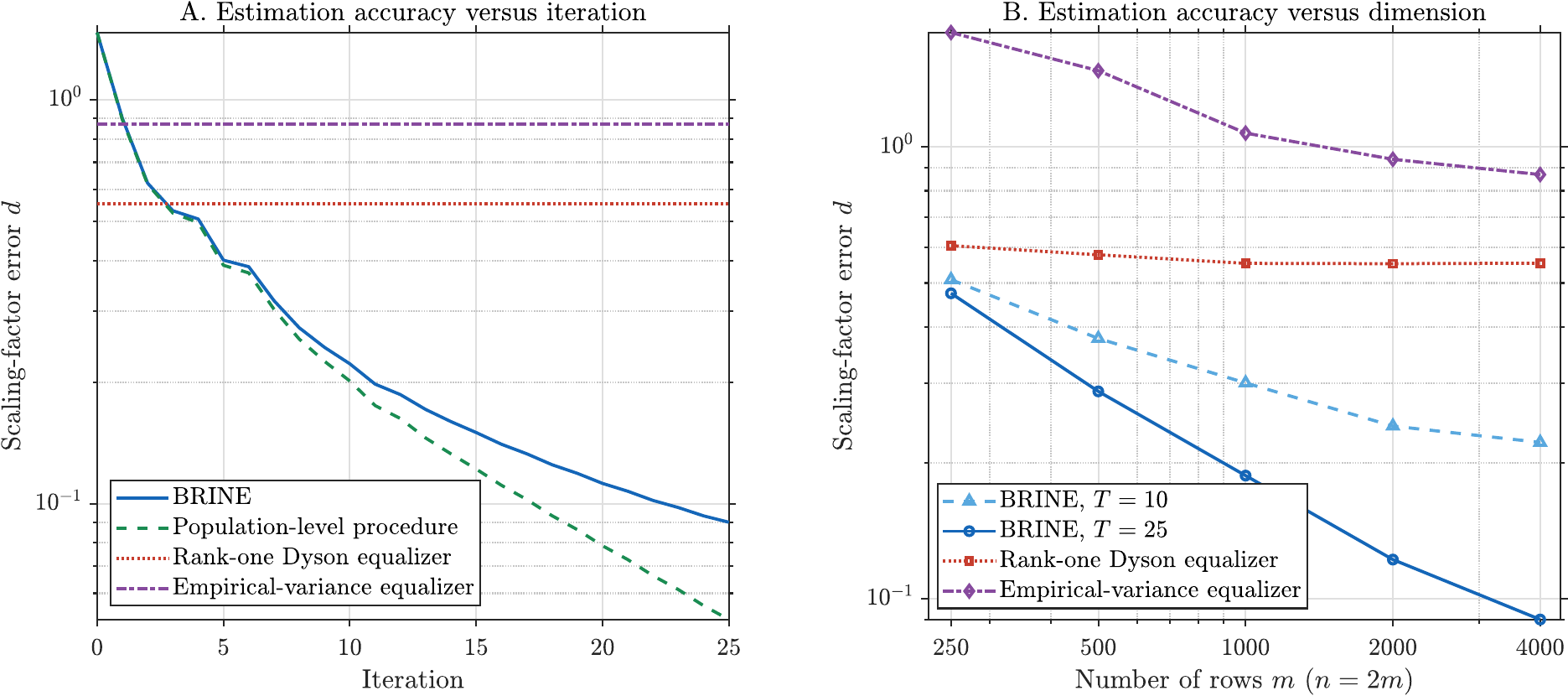}
    \caption{Scaling-factor estimation accuracy in the same signal-plus-noise setup as for Figure~\ref{fig:stylized_intro}. The vertical axis in both panels is the logarithmic error $d$, defined in~\eqref{eq: averaged logarithmic distance}, between the estimated factors and the true scaling factors $(\mathbf{x}^\star,\mathbf{y}^\star)$. \textbf{Panel A.} Error versus iteration for $m=4000$, $n=8000$, obtained from the data-driven procedure (BRINE) and the population-level procedure (computed via the true variance profile), together with the errors of the rank-one Dyson equalizer~\cite{landa2025dyson} and the empirical-variance equalizer~\cite{leeb2021matrix} (which are independent of iterations). The trajectories are displayed through iteration $25$. \textbf{Panel B.} Error versus the number of rows $m$, with $n=2m$, for BRINE with maximum iteration caps $T=10$ and $25$, and the same two alternative methods as in Panel A.}
    \label{fig:convergence-simulation}
\end{figure}

Panel~A of Figure~\ref{fig:convergence-simulation} shows a steady reduction in the error of both the population-level and data-driven BRINE iterations. The error of the data-driven procedure closely follows the population-level procedure over the initial iterations, with a larger discrepancy appearing later, since the error becomes more affected by the accumulated statistical error while the population error becomes small. This behavior is consistent with the geometric convergence established for the population procedure in Theorem~\ref{thm:population_convergence} and the two terms in the error bound of Theorem~\ref{thm: main theoretical guarantees}.
Panel~B shows that BRINE becomes more accurate as the dimensions increase, and that allowing $25$ iterations instead of $10$ leads to lower error, particularly at larger dimensions. For $T=25$ and dimensions $m=250,500,1000,2000$, the error decreases faster than $n^{-1/2}$, which is the dominant term in the bound of Theorem~\ref{thm: main theoretical guarantees} (since $\kappa_n$ is of order $1/m$ in this regime up to logarithmic factors). In contrast, the two alternative methods of~\cite{landa2025dyson} and~\cite{leeb2021matrix} retain substantially larger errors over the tested range. These results illustrate the value of the iterations when the statistical error is sufficiently small, and the benefit of BRINE over alternative approaches, especially when the variance profile has a general pattern beyond rank-one and when strong signal components are also present in the data.

\subsection{Spectral behavior across biological modalities and preprocessing steps}\label{sec:experiments-mp}
A natural question is whether the spectral standardization guaranteed by our theory under the general heteroskedastic model extends to real-world data, where the model assumptions may not hold exactly. To investigate this, we applied Algorithm~\ref{alg:BRINE} with default parameters to a broad collection of biological datasets and evaluated the resulting adherence to the MP law.

\paragraph{Data Collection and Preprocessing.}
We assembled 11 datasets spanning 8 distinct biological modalities: single-cell RNA sequencing, CITE-seq, spatial transcriptomics from four different platforms (SeqFISH+, CosMx, DBiT-seq, and Visium), single-cell ATAC-seq, genotyping, single-cell methylomics, and calcium imaging. We further include a multimodal (multiome) dataset that concatenates RNA expression and chromatin accessibility into a single matrix, combining two fundamentally different data distributions. For each dataset, we considered the preprocessing transformations standard in its respective domain, yielding over 20 dataset-transformation combinations in total. Full details of datasets, sources, and transformations are provided in Appendix \ref{app:all_mp_fits}.

\begin{figure}[t]
  \centering
  \includegraphics[width=\linewidth]{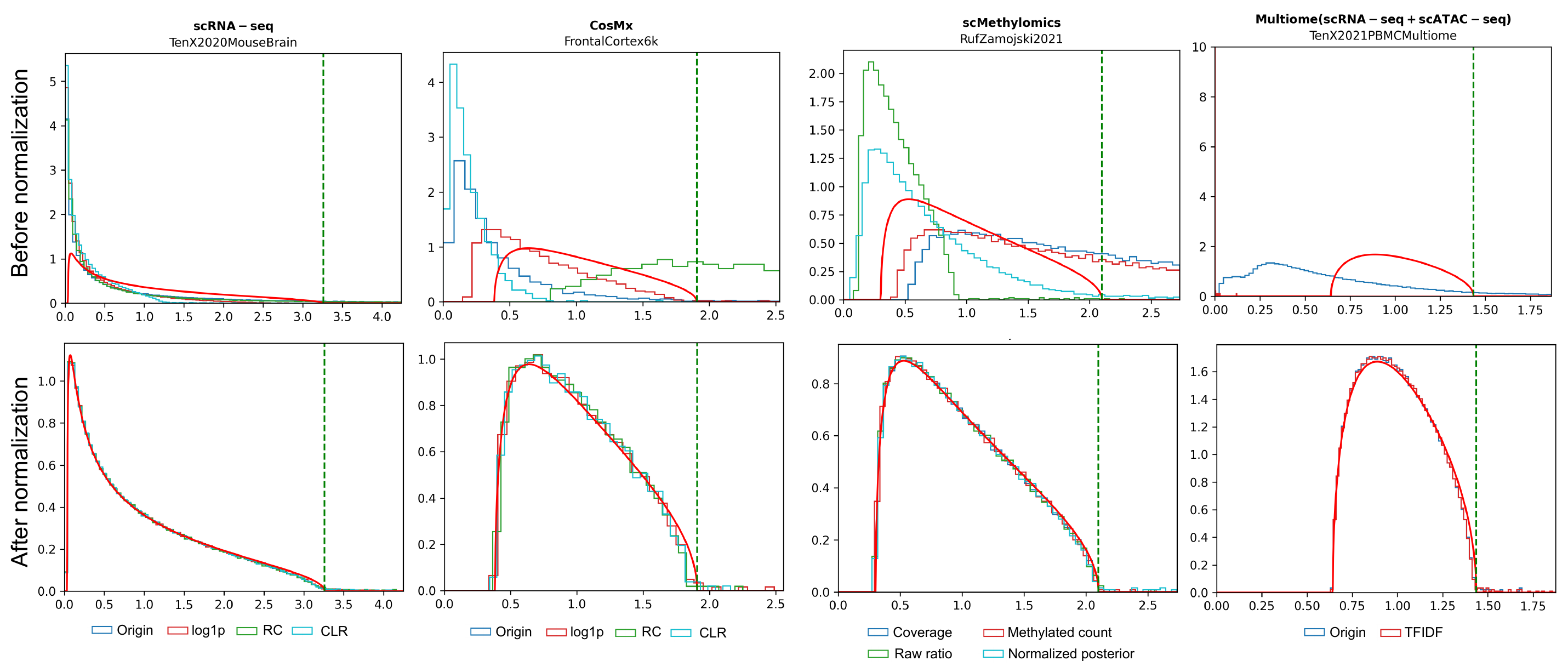}
  \caption{Effect of BRINE normalization on the empirical spectral distribution across four representative biological modalities. \textbf{Top row:} before normalization. \textbf{Bottom row:} after normalization. Each panel overlays the scaled empirical spectral distribution (histograms) with the theoretical MP density (red curve); green dashed lines mark the MP upper edge. Different histogram colors correspond to different preprocessing transformations applied to the raw data. Complete results for all 11 datasets are provided in Figure \ref{fig:before_dyson} and Figure \ref{fig:after_dyson} in the appendix.}
  \label{fig:mp-main}
\end{figure}

\paragraph{Results.}
Three findings emerge consistently across all datasets and transformations:

\subparagraph{Spectral standardization.}
Before normalization, empirical spectral distributions exhibit heavy tails, irregular shapes, and modality-specific artifacts characteristic of heteroskedastic noise (Figure \ref{fig:mp-main}, top row). After applying Algorithm \ref{alg:BRINE}, the empirical spectra align consistently with the theoretical MP distribution (Figure \ref{fig:mp-main}, bottom row). This alignment holds regardless of the input data distribution or the preprocessing applied. Complete spectral fits for all 11 datasets and all transformations are provided in Appendix \ref{app:all_mp_fits} (Figures \ref{fig:before_dyson} and \ref{fig:after_dyson}).

\subparagraph{Robustness to modality and transformation.}
Datasets subjected to different transformations (e.g., log1p vs. CLR vs. relative counts) achieve comparably tight fits to the MP law after normalization. Notably, the multiome dataset, in which continuous RNA expression counts are concatenated with sparse binary ATAC accessibility features, is also successfully standardized, demonstrating that the method accommodates heterogeneous block structure within a single matrix.

\subparagraph{Convergence behavior.}
To quantitatively assess the normalization, we track three diagnostics across iterations: the Kolmogorov--Smirnov (KS) statistic, which measures the maximum deviation between the empirical spectral distribution and the theoretical MP cumulative distribution function (smaller is better); the associated KS nominal $p$-value; and the estimated potential $\hat{J}_t$ from Step~2(d) of Algorithm~\ref{alg:BRINE}. In all tested cases, the KS statistic drops rapidly, indicating improved spectral agreement with the MP law, while the estimated potential $\hat{J}_t$ decreases along the retained iterations and stabilizes. Detailed traces for all datasets are shown in Appendix~\ref{app:all_mp_fits} (Figure~\ref{fig:dyson_convergence}).

\subsection{Departures from independence and low-rank assumptions}
\label{sec:experiments-correlated}

The results of Section~\ref{sec:experiments-mp} show that BRINE produces spectra closely following the MP law across a broad collection of real biological datasets and preprocessing transformations. This agreement provides empirical support for the usefulness of the underlying model, which assumes independent noise entries with a general variance profile and low-rank signal structure. We next examine the robustness of our approach to departures from these assumptions. In particular, we study how dependence between noise entries affects the MP fit, and how increasing signal rank affects the estimated normalization and the resulting largest noise eigenvalue.

\paragraph{Setup.}
To study the effect of noise dependence, we generate a square signal matrix $X\in\mathbb{R}^{m\times m}$ with $m=1000$ and rank $r=10$ as $X=X_0X_1^T$, where $X_0,X_1\in\mathbb{R}^{m\times r}$ have i.i.d.\ standard Gaussian entries. To simulate the noise, we first generate an i.i.d.\ Gaussian noise matrix $E_0$, and then mix its rows by forming the product $R_L E_0$, where $R_L\in\mathbb{R}^{m\times m}$ is a mixing matrix given by
\begin{equation}
(R_L)_{ij} =
\begin{cases}
1, & i=j,\\
h\,\mathbf{1}\{U_{ij}<f\}, & i\neq j,
\end{cases}
\qquad
U_{ij}\overset{\mathrm{i.i.d.}}{\sim}\mathrm{Uniform}(0,1).
\label{eq: correlated noise mixing operator}
\end{equation}
Here, $f\in[0,1]$ controls the expected fraction of nonzero off-diagonal entries, $h\geq 0$ controls their magnitude, and each row of $R_L$ is normalized to unit Euclidean norm. The parameters $f$ and $h$ control the noise dependence; at $h=0$, the mixing operator is the identity and the noise entries are independent. Increasing $f$ mixes each row with more rows on average, while increasing $h$ strengthens the mixing. After this mixing, the noise entries are still homoskedastic with variance $1$ for each entry. We thus rescale the resulting noise matrix entrywise to obtain the block variance profile of Figure~\ref{fig:stylized_intro}, and set the average noise variance to be $1/100$ across all entries. We evaluate BRINE on a $20\times20$ grid of $(f,h)$ values and record the KS statistic measuring the agreement between the normalized spectrum and the MP law.

To study the effect of signal rank, we instead fix the noise model and vary $r$. We set $m=5000$ and $n=10000$, generate independent heteroskedastic noise $E$ with the same block variance profile as Figure~\ref{fig:stylized_intro}, and construct a rank-$r$ signal $X_r$ as above. Specifically, we set $X=aX_r$ while normalizing the signal and noise such that ${\|X_r\|_F^2}/{mn}={\|E\|_F^2}/{mn}=1$, so that $a$ controls the signal-to-noise ratio. We vary $r$ from $1$ to $1000$ and use $a\in\{0.1,1,10,100,1000\}$, with five independent realizations for each setting.

To assess the effect of the signal on noise equalization, we apply the scaling factors estimated by Algorithm~\ref{alg:BRINE} from $Y$ to the corresponding pure noise matrix $E$ and record its largest eigenvalue. We compare this quantity with the MP upper edge $\beta_+$, which is determined by the matrix aspect ratio.

\begin{figure}[t]
    \centering
    \includegraphics[width=\linewidth]{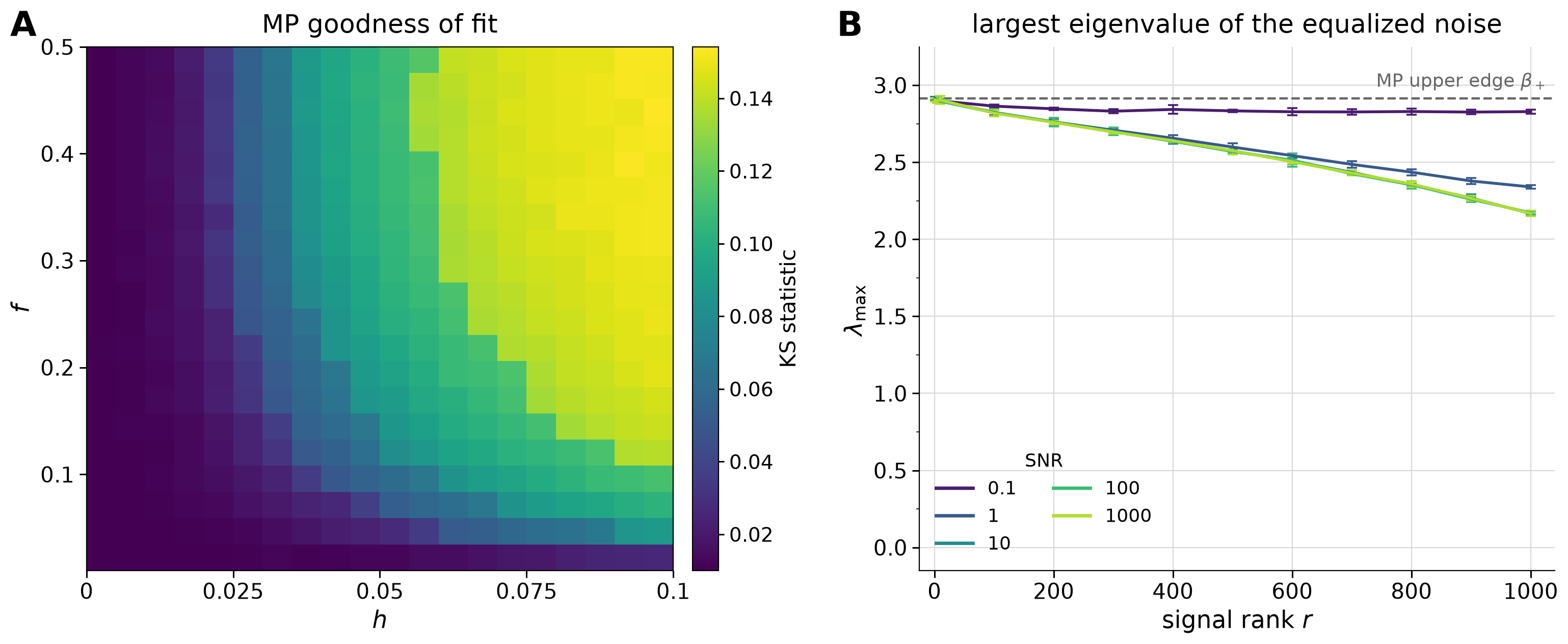}
    \caption{\small Simulated departures from the assumptions of independent noise and low-rank signal structure. \textbf{Panel A.} The effect of correlated noise on the MP fit, for $m=n=1000$ and a rank-$10$ signal. The heatmap shows the Kolmogorov--Smirnov statistic over a $20\times20$ grid of the mixing parameters in~\eqref{eq: correlated noise mixing operator}, with the expected proportion $f$ of nonzero off-diagonal entries on the vertical axis and their intensity $h$ on the horizontal axis. The bottom-left corner corresponds to i.i.d.\ noise. \textbf{Panel B.} The largest eigenvalue of $\widetilde{E} \widetilde{E}^T$, using scaling factors estimated from $Y$, as a function of signal rank for several signal-to-noise ratios (SNRs). Here, $m=5000$ and $n=10000$, and the dashed line marks the MP upper edge $\beta_+$. Curves show the mean over five realizations, with error bars extending two standard deviations on either side of the mean.
    }
    \label{fig:sims-correlated-noise}
\end{figure}

\paragraph{Results.}
Figure~\ref{fig:sims-correlated-noise}A shows that the MP fit remains accurate under weak mixing but deteriorates as the dependence becomes stronger. For a fixed $f$, the KS statistic generally increases with $h$, and the onset of this deterioration occurs at smaller values of $h$ as $f$ increases. Thus, the effect of dependence is determined jointly by the number of rows involved in the mixing and the strength of their contributions. These results show that moderate departures from independence need not substantially affect spectral equalization, whereas sufficiently strong dependence produces a detectable departure from the MP law.

Figure~\ref{fig:sims-correlated-noise}B shows the largest eigenvalue of the equalized noise Gram matrix as the signal rank increases. Across all five signal-to-noise ratios, the largest eigenvalue remains below the MP upper edge $\beta_+$ throughout the tested range. For $a=0.1$ and $a=1$, it remains close to $\beta_+$ even at high ranks, indicating that the estimated normalization is largely unaffected by the signal. For the higher SNRs, the largest eigenvalue gradually moves below the MP upper edge as the rank increases. The effect becomes evident for sufficiently strong signals, but changes relatively little with further increases in signal-to-noise ratio.

Importantly, the estimated normalization does not produce spurious noise eigenvalues above the MP upper edge in the tested settings. The MP edge therefore remains a conservative threshold for distinguishing signal from noise, even when the signal rank is substantially larger than assumed by the theory. At high signal ranks and signal-to-noise ratios, the downward shift of the noise spectrum entails only a mild loss of spectral sensitivity. For example, at $r=500$, the largest noise eigenvalue is approximately $2.6$ for the stronger-signal curves, compared with the MP upper edge of approximately $2.9$. Thus, relative to a threshold based on the actual noise edge, signal components whose observed eigenvalues fall between these two values may be missed, corresponding to a threshold gap of about $10\%$ of the MP edge.

Taken together, the two experiments suggest that our approach can tolerate moderate departures from the independent-noise and low-rank assumptions. Strong noise dependence leads to a detectable deterioration of the MP fit, whereas increasing signal rank has a relatively mild effect on the largest equalized noise eigenvalue, which remains below the MP upper edge throughout the tested range.

\subsection{Identification of rare cell populations in single-cell transcriptomics}\label{sec:experiments-scrna}

Single-cell RNA sequencing (scRNA-seq) measures gene activity in individual cells, allowing researchers to identify cell types and study differences between them. The resulting cell-by-gene RNA count matrix is sparse and affected by technical noise, including variability in RNA capture and sequencing efficiency~\cite{grun2014validation}. A particularly challenging application is the detection of rare cell populations, which may be biologically important despite comprising only a small fraction of a sample~\cite{shekhar2016comprehensive}. Their distinguishing expression patterns are represented in few cells and contribute weak low-rank components that can be obscured by noise. Here, we investigate whether BRINE can improve the ability to identify such populations.

\paragraph{Setup.}
We use six populations from a published scRNA-seq dataset~\cite{abdelaal2019comparison,zheng2017massively}: CD4+ T Helper2, CD8+ Cytotoxic T, CD14+ Monocyte, CD19+ B, CD34+, and CD56+ NK. The population labels were established before sequencing by fluorescence-activated cell sorting (FACS), which separates cells using surface protein markers. We initially draw 1{,}000 cells from each population, retain genes detected in at least 100 cells, and rescale each cell's counts to sum to $10^4$ to adjust for differences in sequencing depth --- a common step known as library normalization~\cite{cole2019performance}. To control rarity, we retain only $s$ randomly selected cells from a target population while keeping the other five populations fixed.

We compare five preprocessing pipelines: \texttt{unprocessed}, which uses the library-normalized counts without further transformation; \texttt{log1p}, which applies the transformation $\log(1+x)$ entrywise to the library-normalized counts; and three BRINE-based pipelines that apply BRINE alone (\texttt{BRINE}), before log1p (\texttt{BRINElog1p}), or after log1p (\texttt{log1pBRINE}). In each BRINE-based pipeline, noise equalization is immediately followed by low-rank reconstruction using the MP-selected rank, as described in Appendix~\ref{sec: low-rank approximation}, and negative entries are clipped to zero. For \texttt{unprocessed} and \texttt{log1p}, we instead compute a truncated SVD with a fixed rank of 50. The \texttt{log1p} pipeline follows the common practice in single-cell analysis, with 50 components matching the default number of components used by Seurat's PCA routine~\cite{stuart2019comprehensive}.

We then construct a $k$-nearest-neighbor graph using Euclidean distances between cells and apply Leiden clustering~\cite{traag2019louvain}. Recovery is evaluated against the FACS labels. Let $\mathcal{T}$ be the set of target (rare population) cells and $C^\star$ the cluster containing the largest number of cells from $\mathcal{T}$. We measure
\begin{equation}
\text{precision} = \frac{|\mathcal{T}\cap C^\star|}{|C^\star|},
\qquad
\text{recall} = \frac{|\mathcal{T}\cap C^\star|}{|\mathcal{T}|},
\qquad
\text{specificity} = \frac{|\mathcal{T}^c \cap (C^\star)^c|}{|\mathcal{T}^c|},
\label{eq: recovery metrics}
\end{equation}
where complements are taken within the analyzed sample.

\begin{figure}
\centering
\includegraphics[width=1.0\linewidth]{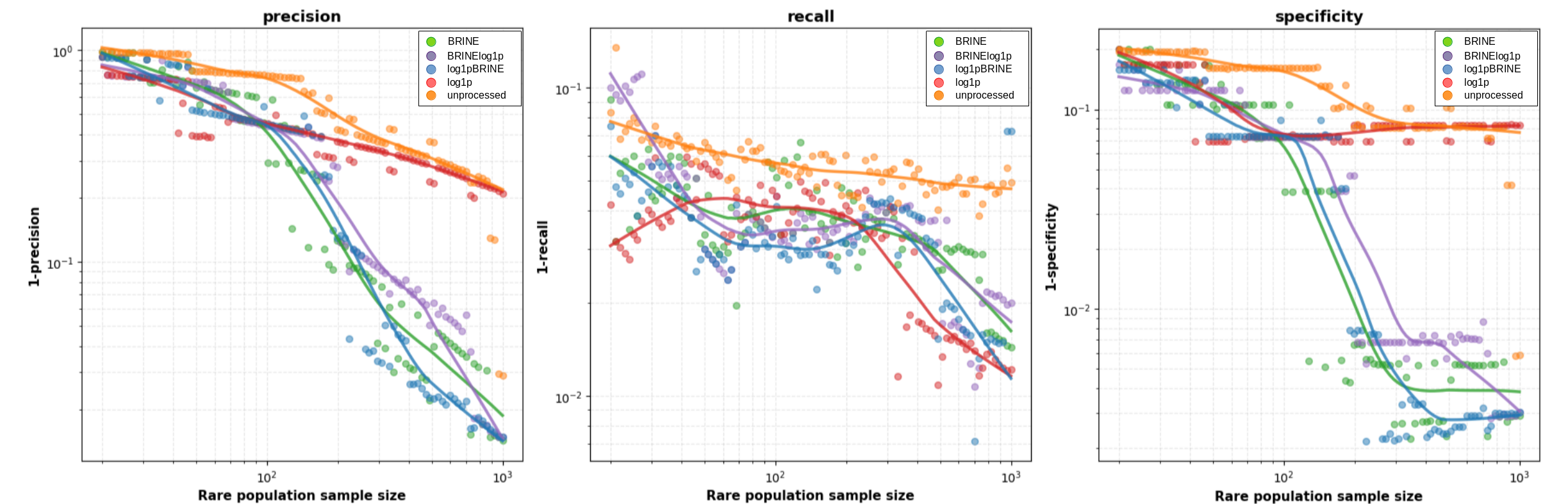}
\caption{Recovery of a rare cell population as a function of the target size $s$ for the five preprocessing pipelines. \textbf{Left to right:} precision, recall, and specificity, as defined in~\eqref{eq: recovery metrics}. Each panel plots one minus the metric, so lower values indicate better recovery. Both axes are logarithmic, and LOWESS curves show the trends.}
\label{fig:zheng_scores}
\end{figure}

\paragraph{Results.}
Figure~\ref{fig:zheng_scores} plots one minus precision, recall, and specificity as a function of the target size $s$, so that lower values indicate better recovery. The five pipelines perform similarly when the target population is very rare, but an advantage for the BRINE-based pipelines clearly emerges around $s=100$ samples from the rare population, approximately $10\%$ of its original size, and becomes more pronounced as the target size increases. The improvement is driven by precision and specificity: all five transforms recover comparable fractions of the target cells within a single cluster, as measured by recall, but the BRINE-based transforms more effectively exclude cells from the remaining five populations. In contrast, the lower precision and specificity of the \texttt{unprocessed} and \texttt{log1p} pipelines indicate that their recovered clusters remain more heavily contaminated by non-target cells.

Overall, this experiment suggests that integrating BRINE into standard processing pipelines in single cell transcriptomics can improve the identification and recovery of rare cell populations by providing a comparatively purer representative subset of such cells, which is important for reliable downstream biological analysis.

\section{Discussion}
\label{sec: discussion}

In this work, we address the problem of blind noise equalization in data matrices: finding row and column scaling factors that equalize the average noise variances across both dimensions without knowing the underlying variance profile. Our primary goal is to make this normalization available under general heteroskedasticity, thereby providing a foundation for signal detection and recovery methods that rely on the spectral behavior of homogeneous noise. Previous work has established benefits of noise normalization for signal detection, matrix denoising, and downstream analysis under various structural or distributional assumptions~\cite{leeb2021matrix,landa2022biwhitening,landa2025dyson,stanley2025principled}. Our contribution substantially expands the settings in which the required scaling factors can be determined reliably from the data. Accordingly, our experiments emphasize spectral normalization across data modalities and preprocessing transformations, with the rare-cell analysis illustrating one downstream application. We note that noise equalization may benefit a broad range of data analysis and inference techniques that were originally developed under homoskedastic noise assumptions. However, a systematic investigation of such benefits is left for future work.

The resolvent parameter $\eta$ in our approach controls the spectral scale used to estimate our proposed normalization. For every fixed $\eta>0$, the population procedure converges to the same noise equalizer, although its convergence rate and the statistical error bounds may depend on $\eta$. In finite samples, this parameter affects the weighting of signal and noise contributions to the resolvent. Our default choice $\eta=1$, following normalization by the median singular value (see the initialization step in Algorithm~\ref{alg:BRINE}), places $\eta$ exactly at the median singular value of the data. This choice balances sensitivity to fluctuations in small singular values against the influence of large signal components. In particular, taking $\eta$ very small emphasizes the lower end of the spectrum, whereas taking it very large makes the updates approach scaling based on row and column energies. The choice $\eta=1$ provides a simple, robust, and scale-adaptive convention. In our experiments, mild variations in the choice of this parameter did not lead to any noticeable differences in the resulting equalization. Characterizing the finite-sample tradeoffs and developing optimal data-driven choices of $\eta$ are of potential interest for future investigation.

Our procedure chooses between row and column updates using a greedy rule that selects the larger decrease in the population potential. A natural alternative is to alternate row and column updates, as in the classical Sinkhorn--Knopp algorithm. A fixed alternating scheme would simplify parts of the statistical analysis by eliminating the need to track discrepancies between the empirical and population update choices. We expect that similar convergence guarantees could be established for this variant. Our preference for the greedy rule is motivated by the potentially different levels of noise heterogeneity across the two dimensions. For example, noise variation may be more pronounced across features than across samples. The greedy rule allows the update schedule to adapt to this imbalance by picking the dimension offering the larger estimated decrease, potentially reducing the number of costly resolvent evaluations required to achieve accurate equalization.

The computational cost of BRINE is dominated by the repeated evaluation of the diagonal entries of the data resolvent~\eqref{eq: definition of R_cal and Y_cal_tilde}. These entries can be computed without matrix inversion via one SVD of the scaled data matrix $\hat{Y}^{(t)}$ at each iteration. For a dense $m\times n$ matrix with $m\leq n$, this requires $O(m^2n)$ operations per iteration. Since this may become expensive for very large datasets, it is of interest to investigate more scalable implementations. For sparse data, selected inversion~\cite{lin2011selinv} can evaluate the required entries efficiently using an LDL factorization, with computational savings depending on the arrangement of the nonzero entries. Alternatively, randomized diagonal estimators~\cite{bekas2007estimator,baston2022stochastic}, combined with conjugate-gradient solvers, can approximate all required diagonal entries by solving a collection of linear systems. A basic implementation using at most $K$ matrix--vector products with the scaled data matrix and its transpose per iteration has total cost $O(TKq)$, where $q$ is the number of nonzero entries of $Y$ and $K$ depends on the data, $\eta$, and the desired accuracy. For dense data, $q=mn$, so the repeated evaluations are cheaper when $K\ll m$. Developing and evaluating these implementations, and understanding how approximation errors affect the accuracy and convergence of the normalization, are left for future work.

Our theory suggests several other potential research directions. One is to accommodate dependence between noise entries. In such settings, it would be useful to characterize which forms of dependence preserve the accuracy of the resolvent-based updates and when variance equalization must be combined with a correction for noise correlations. A second direction concerns sparse data and localized signals. Rare cell populations and signals supported on small subsets of features motivate studying how equalization accuracy depends on signal support size and the sparsity of the observations, beyond the present uniform variance and delocalization assumptions. Finally, our signal amplification result describes signal directions independent of the variance profile. In count data, signal and noise variance are often related, making it natural to investigate how their alignment affects detection and recovery after equalization. Such results could clarify when normalization is beneficial and how it should be combined with subsequent spectral estimation.

\section{Acknowledgments}
The authors acknowledge the use of AI models (Claude Fable and ChatGPT 5.1 Sol) for language polishing, proofreading, coding assistance, technical review of proofs, and refinement of technical arguments. This work was supported by NIH grants UM1PA051410, U54AG076043, U54AG079759, U01DA053628, P50CA121974, and R33DA047037.

\begin{appendices}

\section{Simulation details for Figure~\ref{fig:stylized_intro}}
\label{app:stylized_simulation}

We generate a signal-plus-noise matrix $Y=X+E$ as follows.
We first construct a $2\times2$ block variance profile. The rows are divided into two groups of size $m/2$, and the columns are divided into two groups of size $n/2$. Before the final global rescaling, the relative variances of the four blocks are
\[
B=
\begin{pmatrix}
0.1 & 1\\
20 & 1
\end{pmatrix}.
\]
Thus, if $a(i)\in\{1,2\}$ and $b(j)\in\{1,2\}$ denote the row and column groups containing indices $i$ and $j$, respectively, the initial variance profile is $S^{(0)}_{ij}=B_{a(i),b(j)}/n$. We then sample, uniformly without replacement, a set $\mathcal I\subset[m]$ of ten rows and a set $\mathcal J\subset[n]$ of ten columns. The variances in each selected row and column are multiplied by $100$, so entries in $\mathcal I\times\mathcal J$ receive both multipliers. Finally, the entire variance profile is multiplied by a common scalar so that
\[
\frac{1}{mn}\sum_{i=1}^m\sum_{j=1}^n S_{ij}=\frac{1}{n}.
\]
Conditional on $S$, the noise entries are generated independently according to
\[
E_{ij}=\sqrt{S_{ij}}\,Z_{ij},
\qquad
Z_{ij}\overset{\mathrm{i.i.d.}}{\sim}\mathcal N(0,1).
\]

Let $\lambda_1(EE^T)\geq\cdots\geq\lambda_m(EE^T)$ denote the ordered noise eigenvalues. Since the variance profile contains ten exceptional rows and ten exceptional columns, these features may generate up to $20$ large noise eigenvalues. We therefore use
\[
\tau_{\mathrm{bulk}}
:=
\lambda_{21}(EE^T)
\]
as an estimate of the upper edge of the main noise bulk. The horizontal line labeled ``Upper edge of main noise bulk'' in Panel~B of Figure~\ref{fig:stylized_intro} is based on $\tau_{\mathrm{bulk}}$, while the largest noise eigenvalue $\lambda_1(EE^T)$ is shown separately. These quantities use the simulated noise matrix and are included only to interpret the example; they are not supplied to the proposed algorithm.

The signal has rank $r=20$, with ten strong and ten weak components. We generate independent Gaussian matrices $G_U\in\mathbb R^{m\times r}$ and $G_V\in\mathbb R^{n\times r}$ with i.i.d.\ standard normal entries, and obtain matrices $U\in\mathbb R^{m\times r}$ and $V\in\mathbb R^{n\times r}$ with orthonormal columns from their QR decompositions. All strong signal singular values are set to $s_{\mathrm{strong}}=4\sqrt{\tau_{\mathrm{bulk}}}$, and all weak signal singular values are set to $s_{\mathrm{weak}}=0.6\sqrt{\tau_{\mathrm{bulk}}}$. Writing $\mathbf 1_{10}$ for the length-ten vector of ones, the signal matrix is
\[
X
=
U\,D\!\left\{
s_{\mathrm{strong}}\mathbf 1_{10},
s_{\mathrm{weak}}\mathbf 1_{10}
\right\}V^T.
\]

For the display in Panels~B and C, we also normalize the data by a scalar to match the median of the noise eigenvalues to the median of the MP distribution. This adjustment removes visual discrepancies attributable to global scaling, providing a shape-focused comparison of the noise spectrum to the MP law.

For Algorithm~\ref{alg:BRINE}, we use the default settings (initialization and resolvent parameter), with $T=15$ iterations. Panel~D reports the estimated potential $\widehat J_t$ described in Section~\ref{sec: data-driven procedure} together with the data-based KS score after each update.

\section[Comparison with previous normalization procedures]
{Comparison of BRINE to~\cite{landa2025dyson} and~\cite{leeb2021matrix} under the simulation setup of Figure~\ref{fig:stylized_intro}}
\label{appendix: comparison to previous normalizations}

\begin{figure}[t]
    \centering
    \includegraphics[width=\textwidth]
    {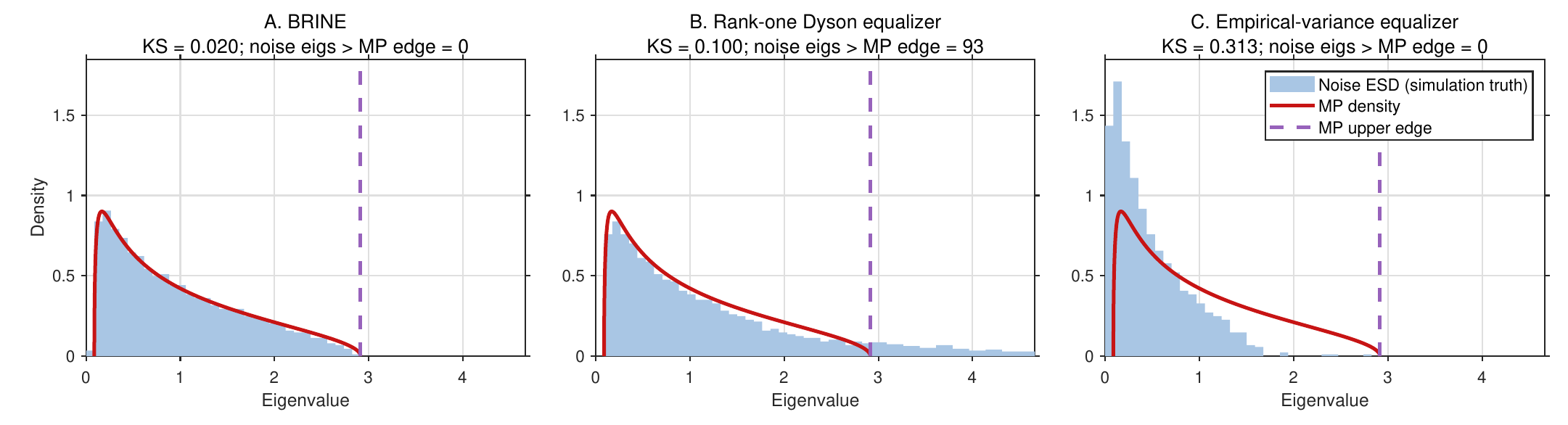}
    \caption{
    Comparison of BRINE to~\cite{landa2025dyson} and~\cite{leeb2021matrix} under the simulation setup of
    Figure~\ref{fig:stylized_intro}. Each panel shows the empirical spectral
    distribution of the noise after applying the scaling factors estimated
    from the observed signal-plus-noise matrix, together with the MP density
    and its upper edge.
    \textbf{A.} BRINE closely restores the MP law, with KS distance $0.02$
    and no noise eigenvalues above the MP edge.
    \textbf{B.} The rank-one Dyson equalizer of~\cite{landa2025dyson},
    derived under a rank-one variance-profile model, leaves a visible upper
    tail, with KS distance $0.100$ and $93$ noise eigenvalues above the MP
    edge.
    \textbf{C.} The empirical row/column variance equalization procedure of~\cite{leeb2021matrix} produces no noise eigenvalues
    above the MP edge but substantially compresses the noise spectrum due to the presence of strong signals in the data,
    resulting in a KS distance of $0.313$.    }
    \label{fig:comparison_previous_normalizations}
\end{figure}

In Figure~\ref{fig:comparison_previous_normalizations} we compare BRINE with the rank-one Dyson equalizer
of~\cite{landa2025dyson} and the empirical variance equalizer of Leeb~\cite{leeb2021matrix}, using the same simulation setup as
in Figure~\ref{fig:stylized_intro}.
For each procedure, the scaling factors are directly from the
observed matrix. These estimated factors are then applied to the true noise matrix $E$ to
evaluate the resulting spectral fit. BRINE is implemented as in
Figure~\ref{fig:stylized_intro}. For the rank-one Dyson equalizer, we convert its output to the variance-$1/n$
normalization used here. For~\cite{leeb2021matrix}, we use the data-driven
row and column variance estimators of Proposition~6.2 therein. Specifically, writing $R_i=\sum_j Y_{ij}^2$, $C_j=\sum_i Y_{ij}^2$, and $T=\sum_{i,j}Y_{ij}^2$ for the squared norms of the preliminarily normalized data matrix, we use scaling factors $x_i=1/R_i$ and $y_j=T/(nC_j)$

\section{Proof of Lemma~\ref{lem: properties of potential function}} \label{appendix: proof of lemma on properties of potential fun}
The partial derivatives of the function $J$ from~\eqref{eq: potential fun} are given by
\begin{align}
    \frac{\partial J}{\partial {\mathbf{x}^{\prime}}_i} &= - \frac{\alpha}{{\mathbf{x}^{\prime}}_i} + \sum_{j=1}^n \mathbf{g}_i S_{i,j} {\mathbf{y}^{\prime}}_j \mathbf{h}_j, \label{eq: derivative wrt to x_prime} \\
    \frac{\partial J}{\partial {\mathbf{y}^{\prime}}_j} &= - \frac{\alpha m}{n {\mathbf{y}^{\prime}}_j} + \sum_{i=1}^m \mathbf{g}_i {\mathbf{x}^{\prime}}_i  S_{i,j} \mathbf{h}_j, \\
    \frac{\partial J}{\partial {\mathbf{g}}_i} &= \eta - \frac{1}{\mathbf{g}}_i + \sum_{j=1}^n {\mathbf{x}^{\prime}}_i S_{i,j} {\mathbf{y}^{\prime}}_j \mathbf{h}_j, \label{eq: derivative wrt to g}\\
    \frac{\partial J}{\partial {\mathbf{h}}_j} &= \eta - \frac{1}{\mathbf{h}}_j + \sum_{i=1}^m {\mathbf{x}^{\prime}}_i \mathbf{g}_i S_{i,j} {\mathbf{y}^{\prime}}_j. \label{eq: derivative wrt to h}
\end{align}
Setting the partial derivatives to zero and after a trivial manipulation, it follows that any critical point $(\mathbf{x}^\star, \mathbf{y}^\star, \mathbf{g}^\star, \mathbf{h}^\star)$ must satisfy the equations
\begin{align}
        \alpha &= \sum_{j=1}^n \mathbf{g}^{\star}_i {\mathbf{x}}^{\star}_i S_{i,j} {\mathbf{y}}^{\star}_j \mathbf{h}^{\star}_j, \qquad \qquad
        \alpha  \frac{m}{n}  = \sum_{i=1}^m \mathbf{g}^{\star}_i {\mathbf{x}}^{\star}_i S_{i,j} {\mathbf{y}}^{\star}_j \mathbf{h}^{\star}_j, \label{eq: critical point analysis 1} \\
        1-\eta \mathbf{g}^{\star}_i &= \sum_{j=1}^n \mathbf{g}^{\star}_i {\mathbf{x}}^{\star}_i S_{i,j} {\mathbf{y}}^{\star}_j \mathbf{h}^{\star}_j, \qquad\qquad
        1-\eta \mathbf{h}^{\star}_j = \sum_{i=1}^m \mathbf{g}^{\star}_i {\mathbf{x}}^{\star}_i S_{i,j} {\mathbf{y}}^{\star}_j \mathbf{h}^{\star}_j, \label{eq: critical point analysis 2}
\end{align}
for all $i\in [m]$ and $j\in [n]$. Comparing~\eqref{eq: critical point analysis 1} and~\eqref{eq: critical point analysis 2}, we obtain
\begin{equation}
    \mathbf{g}^{\star}_i = \frac{1-\alpha}{\eta}, \qquad \qquad \mathbf{h}^{\star}_j = \frac{1-\alpha \frac{m}{n}}{\eta}, \label{eq: critical point analysis 6}
\end{equation}
for all $i\in [m]$ and $j\in [n]$. Substituting these expressions into~\eqref{eq: critical point analysis 1}, we get the system of equations
\begin{equation}
    \sum_{j=1}^n {\mathbf{x}}^{\star}_i S_{i,j} {\mathbf{y}}^{\star}_j = \frac{\alpha \eta^2}{(1-\alpha)(1-\alpha \frac{m}{n})}, \qquad \qquad
    \sum_{i=1}^m {\mathbf{x}}^{\star}_i S_{i,j} {\mathbf{y}}^{\star}_j = \frac{\alpha \eta^2 \frac{m}{n}}{(1-\alpha)(1-\alpha \frac{m}{n})}. \label{eq: critical point analysis 4}
\end{equation}
Therefore, it remains to show that
\begin{equation}
    \frac{\alpha \eta^2}{(1-\alpha)(1-\alpha \frac{m}{n})} = 1. \label{eq: critical point analysis 3}
\end{equation}
To see this, observe that~\eqref{eq: critical point analysis 3} is equivalent to the quadratic equation in the variable $\alpha$
\begin{equation}
    \frac{m}{n} \alpha^2 - (1 + \frac{m}{n} + \eta^2) \alpha + 1 = 0, \label{eq: critical point analysis 5}
\end{equation}
whose smaller root is
\begin{equation}
    \alpha = \frac{1 + \frac{m}{n} + \eta^2 - \sqrt{\left( 1 + \frac{m}{n} + \eta^2 \right)^2 - 4 \frac{m}{n}}}{2\frac{m}{n}},
\end{equation}
as defined in~\eqref{eq: alpha def}.
Thus, the system of equations in~\eqref{eq: critical point analysis 4} coincides with the scaling equations~\eqref{eq: scaling equations}, whose solution is the pair $(a \mathbf{x}^\star,a^{-1} \mathbf{y}^\star)$ for any $a>0$.

To show that the critical points are global minimizers, we first consider the minimization of $J$ over $(\mathbf{x},\mathbf{y})$ while keeping $\mathbf{g}$ and $\mathbf{h}$ fixed. This minimization is equivalent to minimizing the logarithmic barrier function~\eqref{eq: logarithmic barrier function} with the modified variance matrix $D\{\mathbf{g}\} S D\{\mathbf{h}\}/\alpha$ instead of $S$, up to additive constants. Hence, the minimizers are the solutions to the scaling equations~\eqref{eq: scaling equations} when replacing $S_{i,j}$ with $\mathbf{g}_i S_{i,j} \mathbf{h}_j /\alpha$, and denoting the solution as $(\mathbf{x}^\prime,\mathbf{y}^\prime)$. Subsequently, the solution to these modified scaling equations must be
\begin{equation}
    \mathbf{x}_i^\prime = a \frac{\alpha  \mathbf{x}^\star_i}{\mathbf{g}_i}, \qquad \qquad \mathbf{y}_j^\prime = a^{-1} \frac{\mathbf{y}^\star_j}{\mathbf{h}_j},
\end{equation}
for all $i\in [m]$ and $j\in [n]$ and any $a>0$.
Substituting these expressions back into the potential function $J$, and using the scaling equations
$\sum_{j=1}^n \mathbf{x}^\star_i S_{i,j}\mathbf{y}^\star_j = 1$ for all $i$ (hence
$\sum_{i=1}^m\sum_{j=1}^n \mathbf{x}^\star_i S_{i,j}\mathbf{y}^\star_j = m$), we obtain
\begin{equation}
   J(\mathbf{x}^\prime,\mathbf{y}^\prime,\mathbf{g},\mathbf{h})
   =  C
   + \sum_{i=1}^m \left( \eta \mathbf{g}_i + (\alpha-1) \log \mathbf{g}_i \right)
   + \sum_{j=1}^n \left( \eta \mathbf{h}_j + \left(\alpha \frac{m}{n}-1\right) \log \mathbf{h}_j \right),
   \label{eq: potential function after mimnimizing over g and h}
\end{equation}
where
\begin{equation}\label{eq:def_constant_C}
C = \alpha m - \alpha m \log \alpha - \alpha \sum_{i=1}^m \log(\mathbf{x}^\star_i)
- \alpha \frac{m}{n}\sum_{j=1}^n \log(\mathbf{y}^\star_j),
\end{equation}
which does not depend on $(\mathbf{g},\mathbf{h})$.
We note that $\alpha<1$, since the quadratic function in the left-hand side of~\eqref{eq: critical point analysis 5} is positive for $\alpha=0$ and negative for $\alpha=1$, hence the smaller root must be in $(0,1)$. Therefore, the function~\eqref{eq: potential function after mimnimizing over g and h} is strictly convex in $(\mathbf{g},\mathbf{h})$ due to the logarithmic terms. Taking the partial derivatives with respect to $\mathbf{g}_i$ and $\mathbf{h}_j$ and setting to zero, we obtain that the minimizer is given by $\mathbf{g} = \mathbf{g}^\star$ and $\mathbf{h} = \mathbf{h}^\star$.

We now prove property~\ref{lem: potential function lemm property 1} in Lemma~\ref{lem: properties of potential function}. If we fix the positive vectors $\overline{\mathbf{x}}$ and $\overline{\mathbf{y}}$, minimizing $J(\overline{\mathbf{x}},\overline{\mathbf{y}},\mathbf{g},\mathbf{h})$ over $(\mathbf{g},\mathbf{h})$ is equivalent to minimizing the function
\begin{equation}
    J_1(\mathbf{g},\mathbf{h}) = \mathbf{g}^T D\{\mathbf{\overline{\mathbf{x}}}\} S D\{\overline{\mathbf{y}}\} \mathbf{h} + \sum_{i=1}^m \left( \eta \mathbf{g}_i - \log (\mathbf{g}_i) \right) +  \sum_{j=1}^n (\eta \mathbf{h}_j - \log (\mathbf{h}_j) ),
\end{equation}
where we removed constant terms that do not affect the minimization. Let us consider the change of variables $\mathbf{g}_i = e^{\zeta_i}$ and $\mathbf{h}_j = e^{\xi_j}$ for $\zeta_1,\ldots,\zeta_m\in\mathbb{R}$ and $\xi_1,\ldots,\xi_n\in\mathbb{R}$. Under this change of variables, minimizing $J_1$ over the positive vectors $(\mathbf{g},\mathbf{h})$ is equivalent to minimizing the function
\begin{equation}
    J_2(\zeta_1,\ldots,\zeta_m,\xi_1,\ldots,\xi_n) = \sum_{i=1}^m \sum_{j=1}^n \overline{\mathbf{x}}_i S_{i,j} \overline{\mathbf{y}}_j e^{\zeta_i + \xi_j} + \sum_{i=1}^m \left( \eta e^{\zeta_i} - \zeta_i \right) + \sum_{j=1}^n \left( \eta e^{\xi_j} - \xi_j \right),
\end{equation}
which is strictly convex in the variables $\zeta_1,\ldots,\zeta_m,\xi_1,\ldots,\xi_n$ (since the first term is convex while the second and third terms together are strictly convex by the strict convexity of the exponential terms). Moreover, $J_2$ is coercive, since $u\mapsto\eta e^u-u$ tends to $\infty$ as $|u|\to\infty$ and the first term is nonnegative. Hence, $J_2$ admits a unique minimizer, and by consequence so does $J(\overline{\mathbf{x}},\overline{\mathbf{y}},\mathbf{g},\mathbf{h})$ over $(\mathbf{g},\mathbf{h})$, which is denoted by $(\overline{\mathbf{g}},\overline{\mathbf{h}})$. The connection between this minimizer and the solution to~\eqref{eq: Dyson equation} follows immediately from the fact that~\eqref{eq: Dyson equation} is the system of equations that describes the stationary points of $J_1(\mathbf{g},\mathbf{h})$. Lastly, the fact that $\overline{\mathbf{g}}_i < \frac{1}{\eta}$ and $\overline{\mathbf{h}}_j < \frac{1}{\eta}$ follows immediately from~\eqref{eq: Dyson equation} using the fact that $\sum_{j=1}^n \overline{\mathbf{x}}_i S_{i,j} \overline{\mathbf{y}}_j \mathbf{h}_j > 0$ and $ \sum_{i=1}^m \mathbf{g}_i \overline{\mathbf{x}}_i S_{i,j} \overline{\mathbf{y}}_j > 0$ (since all quantities therein are strictly positive in our setting).

As for properties~\ref{lem: potential function lemm property 2} and~\ref{lem: potential function lemm property 3} in Lemma~\ref{lem: properties of potential function}, if we fix the positive vectors $\overline{\mathbf{x}}, \overline{\mathbf{y}},\overline{\mathbf{g}},\overline{\mathbf{h}}$ as described in property~\ref{lem: potential function lemm property 1}, then the function $J({\mathbf{x}^{\prime}},\overline{\mathbf{y}},\overline{\mathbf{g}},\overline{\mathbf{h}})$ is strictly convex with respect to $\mathbf{x}^{'}$ due to the strict convexity of the term $- \alpha \sum_{i=1}^m \log (\mathbf{x}^{'}_i)$ in~\eqref{eq: potential fun} (where the other terms are convex). Hence, the minimizer of $J({\mathbf{x}^{\prime}},\overline{\mathbf{y}},\overline{\mathbf{g}},\overline{\mathbf{h}})$ over the positive vector $\mathbf{x}^{'}$ is unique. To find this minimizer, we set~\eqref{eq: derivative wrt to x_prime} to zero (after replacing $\mathbf{g},\mathbf{h},\mathbf{y}^{'}$ with $\overline{\mathbf{g}},\overline{\mathbf{h}},\overline{\mathbf{y}}$, respectively), obtaining
\begin{equation}
    \mathbf{x}^{'}_i = \frac{\alpha}{\sum_{j=1}^n \overline{\mathbf{g}}_i S_{i,j} \overline{\mathbf{y}}_j \overline{\mathbf{h}}_j} = \overline{\mathbf{x}}_i \left( \frac{\alpha }{1 - \eta \overline{\mathbf{g}}_i} \right),
\end{equation}
where we used~\eqref{eq: Dyson equation} in the last equality.
The proof of property~\ref{lem: potential function lemm property 3} follows analogously.

\section{Resolvent formulas for symmetrized matrices}
\label{appendix: block resolvent formulas}

\begin{lem}
\label{lem: block resolvent formulas}
Let $A\in\mathbb{R}^{m\times n}$, and define its symmetrized matrix
\[
\mathcal A :=
\begin{bmatrix}
0 & A\\
A^T & 0
\end{bmatrix}
\in\mathbb{R}^{(m+n)\times(m+n)}.
\]
Fix $\eta>0$ and let $\mathcal R_A(\imath\eta) := (\mathcal A-\imath\eta I_{m+n})^{-1}$. Then
\begin{equation}
\label{eq: block resolvent formula full}
\mathcal R_A(\imath\eta)
=
\begin{bmatrix}
\imath\eta(AA^T+\eta^2 I_m)^{-1}
&
A(A^T A+\eta^2 I_n)^{-1}
\\[0.5em]
A^T(AA^T+\eta^2 I_m)^{-1}
&
\imath\eta(A^T A+\eta^2 I_n)^{-1}
\end{bmatrix}.
\end{equation}
In particular,
\begin{align}
\operatorname{Im}\{(\mathcal R_A(\imath\eta))_{i,i}\}
&=
\eta\big[(AA^T+\eta^2 I_m)^{-1}\big]_{i,i},
\qquad i\in[m], \label{eq: block resolvent diag formula left}
\\
\operatorname{Im}\{(\mathcal R_A(\imath\eta))_{m+j,m+j}\}
&=
\eta\big[(A^T A+\eta^2 I_n)^{-1}\big]_{j,j},
\qquad j\in[n]. \label{eq: block resolvent diag formula right}
\end{align}
\end{lem}

\begin{proof}
Observe that
\[
(\mathcal A-\imath\eta I_{m+n})(\mathcal A+\imath\eta I_{m+n})
=
\mathcal A^2+\eta^2 I_{m+n}.
\]
Therefore, we have the identity
\[
R_A(\imath\eta) = (\mathcal A-\imath\eta I_{m+n})^{-1}
=
(\mathcal A+\imath\eta I_{m+n})(\mathcal A^2+\eta^2 I_{m+n})^{-1}.
\]
Since
\[
\mathcal A^2
=
\begin{bmatrix}
AA^T & 0\\
0 & A^T A
\end{bmatrix},
\]
we can write
\[
(\mathcal A^2+\eta^2 I_{m+n})^{-1}
=
\begin{bmatrix}
(AA^T+\eta^2 I_m)^{-1} & 0\\
0 & (A^T A+\eta^2 I_n)^{-1}
\end{bmatrix}.
\]
Overall, we obtain
\begin{align*}
\mathcal R_A(\imath\eta)
&=
\begin{bmatrix}
\imath\eta I_m & A\\
A^T & \imath\eta I_n
\end{bmatrix}
\begin{bmatrix}
(AA^T+\eta^2 I_m)^{-1} & 0\\
0 & (A^T A+\eta^2 I_n)^{-1}
\end{bmatrix} \\
&=
\begin{bmatrix}
\imath\eta(AA^T+\eta^2 I_m)^{-1}
&
A(A^T A+\eta^2 I_n)^{-1}
\\[0.5em]
A^T(AA^T+\eta^2 I_m)^{-1}
&
\imath\eta(A^T A+\eta^2 I_n)^{-1}
\end{bmatrix},
\end{align*}
which proves~\eqref{eq: block resolvent formula full}.
Finally, the matrices $(AA^T+\eta^2 I_m)^{-1}$ and $(A^T A+\eta^2 I_n)^{-1}$ are real-valued, so taking the imaginary part of the diagonal entries of the
top-left and bottom-right blocks in~\eqref{eq: block resolvent formula full}
immediately yields~\eqref{eq: block resolvent diag formula left}
and~\eqref{eq: block resolvent diag formula right}.
\end{proof}

\section{Proof of Theorem~\ref{thm:population_convergence}}\label{appendix:proof_population_convergence}

\subsection{Supporting lemmas}

\subsubsection{Uniform bounds for the population-level procedure}

\begin{lem}\label{lem: finite-time bounds population}
Suppose that Assumption~\ref{assump: population bounds} holds and $\mathbf x^{(0)}=\mathbf1_m$, $\mathbf y^{(0)}=\mathbf1_n$. There exist constants $0<c_0\leq C_0<\infty$, $b_0>0$, and $r_0\in(0,1)$, depending only on $\eta,c,C,\gamma_-,\gamma_+$, such that
\begin{equation}\label{eq: uniform population bounds}
\begin{aligned}
c_0\le x_i^{(t)},y_j^{(t)}\le C_0
&\qquad(t\ge0),\\
b_0\le g_i^{(t)},h_j^{(t)}\le\frac{1-r_0}{\eta}
&\qquad(t\ge1),
\end{aligned}
\end{equation}
for all $i\in[m]$, $j\in[n]$. These bounds hold for any sequence of choices between the row and column updates.
\end{lem}

\begin{proof}
Normalize a solution of~\eqref{eq: scaling equations} so that $\sum_i x_i^\star=m$. The column equations and~\eqref{eq: uniform variance bounds} give
\[
y_j^\star=\frac{m/n}{\sum_i x_i^\star S_{ij}}
\in[1/C,1/c].
\]
Substitution into the row equations then yields
\begin{equation}\label{eq: bounded optimal scalings}
    \frac{c}{C}\le x_i^\star\le\frac{C}{c},
    \qquad
    \frac1C\le y_j^\star\le\frac1c.
\end{equation}

We next show that the population iterates remain bounded between
fixed scalar multiples of the desired scaling factors. For this
purpose, we use the variables
\[
\mathbf z=\mathbf y^{-1},
\qquad
\mathbf p=(\mathbf x,\mathbf z),
\qquad
\mathbf p^\star=
\bigl(\mathbf x^\star,(\mathbf y^\star)^{-1}\bigr),
\]
where vector inverses are taken entrywise. Let $\mathcal T$ denote
either the prescribed row update or the prescribed column update
from~\eqref{eq: x_t update}--\eqref{eq: y_t update}, expressed in
the variable $\mathbf p$. We will show that each of these two maps
preserves entrywise inequalities, commutes with multiplication by
a positive scalar, and fixes $\mathbf p^\star$. Together, these
properties will allow us to propagate upper and lower bounds given
by scalar multiples of $\mathbf p^\star$.

We first establish the comparison property needed for the updates.
For a positive pair $(\mathbf x,\mathbf z)$, let
$(\mathbf g,\mathbf h)$ be the unique positive solution
of~\eqref{eq: Dyson equation} with scaling vectors
$\mathbf x$ and $\mathbf y=\mathbf z^{-1}$, and define
\[
\mathbf a=\mathbf x\circ\mathbf g,
\qquad
\mathbf b=\mathbf z^{-1}\circ\mathbf h.
\]
Thus, $\mathbf a$ and $\mathbf b$ are determined by the scaling
pair $(\mathbf x,\mathbf z)$ through its Dyson solution.
In these variables,~\eqref{eq: Dyson equation} becomes
\begin{equation}\label{eq: transformed population Dyson}
\frac1{a_i}=\frac{\eta}{x_i}+(S\mathbf b)_i,
\qquad
\frac1{b_j}=\eta z_j+(S^T\mathbf a)_j.
\end{equation}

Consider another positive pair
$(\widetilde{\mathbf x},\widetilde{\mathbf z})$, and let
$(\widetilde{\mathbf a},\widetilde{\mathbf b})$ be the corresponding
vectors defined from its Dyson solution. Suppose that
\[
\mathbf x\le\widetilde{\mathbf x},
\qquad
\mathbf z\le\widetilde{\mathbf z},
\]
where all inequalities between vectors are understood entrywise.
We claim that
\[
\mathbf a\le\widetilde{\mathbf a},
\qquad
\mathbf b\ge\widetilde{\mathbf b}.
\]
To prove the first inequality, suppose otherwise and set
\[
\theta:=\max_{i\in[m]}\frac{a_i}{\widetilde a_i}>1.
\]
Then $\mathbf a\le\theta\widetilde{\mathbf a}$ and
$\mathbf z\le\widetilde{\mathbf z}
\le\theta\widetilde{\mathbf z}$. Since $S$ has positive entries,
the second equation in~\eqref{eq: transformed population Dyson}
gives, for every $j\in[n]$,
\[
b_j
=\frac1{\eta z_j+(S^T\mathbf a)_j}
\ge
\frac1{\theta\bigl(\eta\widetilde z_j+
                  (S^T\widetilde{\mathbf a})_j\bigr)}
=\frac{\widetilde b_j}{\theta}.
\]
Choose an index $i_0$ at which the maximum defining $\theta$ is
attained. The first equation
in~\eqref{eq: transformed population Dyson} then yields
\[
\begin{aligned}
\frac1{a_{i_0}}
&=\frac{\eta}{x_{i_0}}+(S\mathbf b)_{i_0}\\
&\ge
\frac{\eta}{\widetilde x_{i_0}}
+\frac1\theta(S\widetilde{\mathbf b})_{i_0}\\
&>
\frac1\theta
\left(
\frac{\eta}{\widetilde x_{i_0}}
+(S\widetilde{\mathbf b})_{i_0}
\right)\\
&=\frac1{\theta\widetilde a_{i_0}}
=\frac1{a_{i_0}},
\end{aligned}
\]
which leads to a contradiction. Hence,
$\mathbf a\le\widetilde{\mathbf a}$. Returning to the second
equation in~\eqref{eq: transformed population Dyson}, we now obtain
\[
b_j
=\frac1{\eta z_j+(S^T\mathbf a)_j}
\ge
\frac1{\eta\widetilde z_j+
       (S^T\widetilde{\mathbf a})_j}
=\widetilde b_j.
\]
Here both terms in the denominator increase when passing from
$(\mathbf x,\mathbf z)$ to
$(\widetilde{\mathbf x},\widetilde{\mathbf z})$:
the first because $\mathbf z\le\widetilde{\mathbf z}$, and the
second because $\mathbf a\le\widetilde{\mathbf a}$.

We now use these comparisons to show that each prescribed update
preserves entrywise inequalities. Write $(\mathbf x',\mathbf z)$
for the result of the row update and $(\mathbf x,\mathbf z')$
for the result of the column update. Using
\eqref{eq: transformed population Dyson}, the update
formulas~\eqref{eq: x_t update}--\eqref{eq: y_t update} become
\begin{equation}\label{eq: monotone population updates}
\begin{aligned}
x_i'
&=\frac{\alpha x_i}{1-\eta g_i}
=\alpha\left(x_i+\frac{\eta}{(S\mathbf b)_i}\right),\\
z_j'
&=\frac{(1-\eta h_j)z_j}{\alpha m/n}
=\frac1{\alpha m/n}
\left(
\frac1{z_j}+\frac{\eta}{(S^T\mathbf a)_j}
\right)^{-1}.
\end{aligned}
\end{equation}
For the two ordered scaling pairs considered above, we have
$S\mathbf b\ge S\widetilde{\mathbf b}$ and
$S^T\mathbf a\le S^T\widetilde{\mathbf a}$. Consequently,
\[
\begin{aligned}
x_i'
&\le
\alpha\left(
\widetilde x_i+\frac{\eta}{(S\widetilde{\mathbf b})_i}
\right)
=\widetilde x_i',\\
z_j'
&\le
\frac1{\alpha m/n}
\left(
\frac1{\widetilde z_j}
+\frac{\eta}{(S^T\widetilde{\mathbf a})_j}
\right)^{-1}
=\widetilde z_j'.
\end{aligned}
\]
The coordinates left unchanged by each update also retain their
ordering. Thus, for either prescribed choice of $\mathcal T$,
\begin{equation}\label{eq: population update monotonicity}
\mathbf p\le\widetilde{\mathbf p}
\quad\Longrightarrow\quad
\mathcal T(\mathbf p)\le\mathcal T(\widetilde{\mathbf p}).
\end{equation}

Next, multiplying $\mathbf p=(\mathbf x,\mathbf z)$ by
$\lambda>0$ replaces the original scaling pair
$(\mathbf x,\mathbf y)$ by
$(\lambda\mathbf x,\lambda^{-1}\mathbf y)$.
This transformation leaves the scaled variance matrix unchanged:
\[
D\{\lambda\mathbf x\}S
D\{(\lambda\mathbf z)^{-1}\}
=
D\{\mathbf x\}S D\{\mathbf z^{-1}\}.
\]
Therefore, the unique Dyson solution $(\mathbf g,\mathbf h)$
is unchanged as well. The multipliers of $x_i$ and $z_j$ in
\eqref{eq: monotone population updates} consequently remain
unchanged, so each update satisfies
\begin{equation}\label{eq: population update homogeneity}
\mathcal T(\lambda\mathbf p)
=\lambda\mathcal T(\mathbf p),
\qquad \lambda>0.
\end{equation}
Moreover,~\eqref{eq: formulas for g_star and h_star} gives
\[
\frac{\alpha}{1-\eta g_i^\star}=1,
\qquad
\frac{1-\eta h_j^\star}{\alpha m/n}=1.
\]
Hence, both prescribed updates fix the desired pair:
$\mathcal T(\mathbf p^\star)=\mathbf p^\star$.

We can now propagate the desired bounds. By
\eqref{eq: bounded optimal scalings}, every entry of
$\mathbf p^\star$ belongs to $[L_-,L_+]$, where
\[
L_-:=\min\{c/C,c\}\le1,
\qquad
L_+:=\max\{C/c,C\}\ge1.
\]
Write $\mathbf z^{(t)}=(\mathbf y^{(t)})^{-1}$ and
$\mathbf p^{(t)}=(\mathbf x^{(t)},\mathbf z^{(t)})$.
Since $\mathbf p^{(0)}=\mathbf1_{m+n}$, we have
\[
L_+^{-1}\mathbf p^\star
\le\mathbf p^{(0)}
\le L_-^{-1}\mathbf p^\star.
\]
Suppose that these inequalities hold at iteration $t$, and let
$\mathcal T$ be the update selected at iteration $t+1$.
Equations~\eqref{eq: population update monotonicity}
and~\eqref{eq: population update homogeneity}, together with
$\mathcal T(\mathbf p^\star)=\mathbf p^\star$, give
\[
\begin{aligned}
\mathbf p^{(t+1)}
&=\mathcal T(\mathbf p^{(t)})
\ge\mathcal T(L_+^{-1}\mathbf p^\star)
=L_+^{-1}\mathbf p^\star,\\
\mathbf p^{(t+1)}
&=\mathcal T(\mathbf p^{(t)})
\le\mathcal T(L_-^{-1}\mathbf p^\star)
=L_-^{-1}\mathbf p^\star.
\end{aligned}
\]
Induction therefore yields
\begin{equation}\label{eq: invariant population bounds}
L_+^{-1}\mathbf p^\star
\le\mathbf p^{(t)}
\le L_-^{-1}\mathbf p^\star
\qquad\text{for every }t\ge0,
\end{equation}
regardless of the sequence of selected updates. Since the entries
of $\mathbf p^\star$ lie in $[L_-,L_+]$, this implies
\[
\frac{L_-}{L_+}
\le x_i^{(t)},z_j^{(t)}
\le\frac{L_+}{L_-}.
\]
Taking reciprocals of the bounds for $z_j^{(t)}$ gives the same
bounds for $y_j^{(t)}$. We may therefore take
$c_0=L_-/L_+$ and $C_0=L_+/L_-$.

Finally, the Dyson equations and $g_i^{(t)},h_j^{(t)}<1/\eta$ give
\[
g_i^{(t)},h_j^{(t)}
\ge b_0:=\left(\eta+\frac{CC_0^2}{\eta}\right)^{-1}.
\]
Multiplying the Dyson equations by $g_i^{(t)}$ or $h_j^{(t)}$, respectively, and using the lower bounds on the scalings and variances yields
\[
1-\eta g_i^{(t)}\ge c c_0^2 b_0^2,
\qquad
1-\eta h_j^{(t)}\ge\gamma_- c c_0^2 b_0^2.
\]
Taking $r_0=\min\{1/2,\gamma_-c c_0^2 b_0^2\}$ proves~\eqref{eq: uniform population bounds}. The same bounds apply to the normalized optimal scalings in~\eqref{eq: bounded optimal scalings} and their Dyson vectors.
\end{proof}

\subsubsection{Curvature of the potential}

We use the potential~\eqref{eq: potential fun} in logarithmic coordinates. For positive vectors $\mathbf x,\mathbf g\in\mathbb R^m$ and $\mathbf y,\mathbf h\in\mathbb R^n$, write
\[
\boldsymbol\theta=(\log\mathbf x,\log\mathbf y,\log\mathbf g,\log\mathbf h),
\qquad
\Phi(\boldsymbol\theta)=J(\mathbf x,\mathbf y,\mathbf g,\mathbf h),
\]
where logarithms are taken entrywise. Choose a solution $(\mathbf x^\star,\mathbf y^\star)$ of~\eqref{eq: scaling equations} with $\sum_i x_i^\star=m$, as in the proof of Lemma~\ref{lem: finite-time bounds population}, and let $(\mathbf g^\star,\mathbf h^\star)$ be given by~\eqref{eq: formulas for g_star and h_star}. Define
$\boldsymbol\theta^\star=(\log\mathbf x^\star,\log\mathbf y^\star,
\log\mathbf g^\star,\log\mathbf h^\star)$ and
$\Phi_\star=\Phi(\boldsymbol\theta^\star)$. By Lemma~\ref{lem: properties of potential function}, $\Phi_\star$ is the global minimum of $\Phi$; by~\eqref{eq: optimal normalized potential}, $\Phi_\star=m(J_\star+\alpha)$.

\begin{lem}[Curvature of the potential]\label{lem: population potential curvature}
Suppose that Assumption~\ref{assump: population bounds} holds. Let $c_0,C_0,b_0$ be the constants constructed in the proof of Lemma~\ref{lem: finite-time bounds population}. There is a constant $\mu\in(0,1)$, depending only on $\eta,c,C,\gamma_-,\gamma_+$, such that, for every choice of positive vectors $\mathbf x,\mathbf g\in\mathbb R^m$ and $\mathbf y,\mathbf h\in\mathbb R^n$ satisfying
\begin{equation}\label{eq: population curvature hypotheses}
\begin{aligned}
c_0\le x_i\le C_0,\qquad g_i\ge b_0
&\qquad\text{for all }i\in[m],\\
c_0\le y_j\le C_0,\qquad h_j\ge b_0
&\qquad\text{for all }j\in[n],
\end{aligned}
\end{equation}
we have
\begin{equation}\label{eq: logarithmic gradient gap}
\|\nabla\Phi(\boldsymbol\theta)\|_2^2
\ge 2\mu\bigl(\Phi(\boldsymbol\theta)-\Phi_\star\bigr)
\end{equation}
and
\begin{equation}\label{eq: distance controlled by potential}
d\bigl((\mathbf x,\mathbf y),(\mathbf x^\star,\mathbf y^\star)\bigr)^2
\le \frac{2}{m\mu}\bigl(\Phi(\boldsymbol\theta)-\Phi_\star\bigr).
\end{equation}
The vectors $\mathbf g,\mathbf h$ need not solve the Dyson equations~\eqref{eq: Dyson equation} for $(\mathbf x,\mathbf y)$.
\end{lem}

\begin{proof}
Equation~\eqref{eq: bounded optimal scalings} and the construction of $c_0,C_0$ give $c_0\le x_i^\star\le C_0$ and $c_0\le y_j^\star\le C_0$ for all $i\in[m]$, $j\in[n]$. Using these upper bounds and $g_i^\star,h_j^\star<1/\eta$ in the Dyson equations~\eqref{eq: Dyson equation}, together with $S_{ij}\le C/n$ and $m/n\le1$, gives
\[
g_i^\star\ge b_0,\qquad h_j^\star\ge b_0,
\qquad b_0=\left(\eta+\frac{CC_0^2}{\eta}\right)^{-1}.
\]

To compare $(\mathbf x,\mathbf y)$ with $(\mathbf x^\star,\mathbf y^\star)$, we use the reciprocal rescaling $(\mathbf x,\mathbf y)\mapsto(\lambda\mathbf x,\lambda^{-1}\mathbf y)$, $\lambda>0$. This replacement preserves each product $x_i y_j$, and the changes in the logarithmic terms of~\eqref{eq: potential fun} cancel. Thus $J$, its gradient in logarithmic coordinates, and $d((\mathbf x,\mathbf y),(\mathbf x^\star,\mathbf y^\star))$ are unchanged.

Choose $\lambda$ so that, after the replacement,
\begin{equation}\label{eq: logarithmic scalar normalization}
\sum_i\log x_i-\sum_j\log y_j
=\sum_i\log x_i^\star-\sum_j\log y_j^\star.
\end{equation}
Such a $\lambda$ exists uniquely because rescaling adds $(m+n)\log\lambda$ to the left-hand side. Retain the symbols $\mathbf x,\mathbf y$ for the rescaled vectors and set $\boldsymbol\theta=(\log\mathbf x,\log\mathbf y,\log\mathbf g,\log\mathbf h)$, with $\mathbf g,\mathbf h$ unchanged. The bounds $x_i y_j\ge c_0^2$, $g_i\ge b_0$, and $h_j\ge b_0$ remain valid. It suffices to prove~\eqref{eq: logarithmic gradient gap} and~\eqref{eq: distance controlled by potential} with this choice of $\mathbf x,\mathbf y$, because both sides of each inequality are invariant under the rescaling.

Define the vectors along the segment from $\boldsymbol\theta$ to $\boldsymbol\theta^\star$ by
\[
\boldsymbol\theta_s
:=(\log\mathbf x(s),\log\mathbf y(s),\log\mathbf g(s),\log\mathbf h(s))
=(1-s)\boldsymbol\theta+s\boldsymbol\theta^\star,
\qquad 0\le s\le1.
\]
Set $A_0=c c_0^2b_0^2$ and $B_0=\eta b_0$. The bounds on $x_i y_j$, $g_i$, and $h_j$ above, the corresponding bounds on $x_i^\star,y_j^\star,g_i^\star,h_j^\star$, and $S_{ij}\ge c/n$ imply, for every $s\in[0,1]$,
\begin{equation}\label{eq: population Hessian coefficient bounds}
\begin{aligned}
x_i(s)S_{ij}y_j(s)g_i(s)h_j(s)
&=(x_iS_{ij}y_jg_ih_j)^{1-s}
  (x_i^\star S_{ij}y_j^\star g_i^\star h_j^\star)^s
  \ge\frac{A_0}{n},\\
\eta g_i(s)&=(\eta g_i)^{1-s}(\eta g_i^\star)^s\ge B_0,\\
\eta h_j(s)&=(\eta h_j)^{1-s}(\eta h_j^\star)^s\ge B_0.
\end{aligned}
\end{equation}

Let $\boldsymbol\xi=(\mathbf a,\mathbf b,\mathbf c,\mathbf d)$, where $\mathbf a,\mathbf c\in\mathbb R^m$ and $\mathbf b,\mathbf d\in\mathbb R^n$. The logarithmic terms of $J$ are affine in $\boldsymbol\theta$. Differentiating $u\mapsto\Phi(\boldsymbol\theta_s+u\boldsymbol\xi)$ twice at $u=0$ therefore gives
\begin{equation}\label{eq: logarithmic potential Hessian}
\begin{aligned}
\boldsymbol\xi^T\nabla^2\Phi(\boldsymbol\theta_s)\boldsymbol\xi
&=\sum_{i,j}x_i(s)S_{ij}y_j(s)g_i(s)h_j(s)(a_i+b_j+c_i+d_j)^2\\
&\quad+\eta\sum_i g_i(s)c_i^2+\eta\sum_j h_j(s)d_j^2.
\end{aligned}
\end{equation}
To bound~\eqref{eq: logarithmic potential Hessian} below, restrict $\boldsymbol\xi$ to directions preserving~\eqref{eq: logarithmic scalar normalization}, namely $\sum_i a_i=\sum_j b_j$. Let $\mathbf e=(\mathbf1_m,-\mathbf1_n)$, $\boldsymbol\xi_1=(\mathbf a,\mathbf b)$, and $\boldsymbol\xi_2=(\mathbf c,\mathbf d)$. For $\boldsymbol\xi_1\perp\mathbf e$,
\begin{equation}\label{eq: bipartite quadratic lower bound}
\begin{aligned}
\frac1n\sum_{i,j}(a_i+b_j)^2
&=\|\mathbf a\|_2^2+\frac mn\|\mathbf b\|_2^2
+\frac2n\left(\sum_i a_i\right)\left(\sum_j b_j\right)\\
&\ge\gamma_-\|\boldsymbol\xi_1\|_2^2.
\end{aligned}
\end{equation}
Let $\boldsymbol\xi_{2,\perp}$ be the orthogonal projection of $\boldsymbol\xi_2$ onto $\mathbf e^\perp$. The difference $\boldsymbol\xi_2-\boldsymbol\xi_{2,\perp}$ is a multiple of $\mathbf e$, so $(\boldsymbol\xi_{2,\perp})_i+(\boldsymbol\xi_{2,\perp})_{m+j}=c_i+d_j$. Applying~\eqref{eq: bipartite quadratic lower bound} to $\boldsymbol\xi_1+\boldsymbol\xi_{2,\perp}\in\mathbf e^\perp$ and using~\eqref{eq: population Hessian coefficient bounds} in~\eqref{eq: logarithmic potential Hessian} yields
\[
\boldsymbol\xi^T\nabla^2\Phi(\boldsymbol\theta_s)\boldsymbol\xi
\ge A_0\gamma_-\|\boldsymbol\xi_1+\boldsymbol\xi_{2,\perp}\|_2^2
+B_0\|\boldsymbol\xi_2\|_2^2.
\]
Together with
\[
\|\boldsymbol\xi_1\|_2^2+\|\boldsymbol\xi_2\|_2^2
\le2\|\boldsymbol\xi_1+\boldsymbol\xi_{2,\perp}\|_2^2
+3\|\boldsymbol\xi_2\|_2^2,
\]
this gives, for every $s\in[0,1]$ and every $\boldsymbol\xi$ with $\sum_i a_i=\sum_j b_j$,
\begin{equation}\label{eq: uniform logarithmic curvature}
\boldsymbol\xi^T\nabla^2\Phi(\boldsymbol\theta_s)\boldsymbol\xi
\ge\mu\|\boldsymbol\xi\|_2^2,
\qquad
\mu:=\min\left\{\frac{A_0\gamma_-}{2},\frac{B_0}{3}\right\}\in(0,1).
\end{equation}
Here $\mu<1$ because $B_0=\eta b_0<1$.

We now apply~\eqref{eq: uniform logarithmic curvature} to the
direction $\boldsymbol\xi = \boldsymbol\delta:=\boldsymbol\theta-\boldsymbol\theta^\star$.
Denote the first two blocks of $\boldsymbol\delta$ by
$\mathbf a$ and $\mathbf b$, so that
\[
a_i=\log x_i-\log x_i^\star,
\qquad
b_j=\log y_j-\log y_j^\star.
\]
The normalization~\eqref{eq: logarithmic scalar normalization} gives
\[
\begin{aligned}
\sum_i a_i-\sum_j b_j
&=
\left(\sum_i\log x_i-\sum_j\log y_j\right)
-
\left(\sum_i\log x_i^\star-\sum_j\log y_j^\star\right)\\
&=0.
\end{aligned}
\]
Thus $\sum_i a_i=\sum_j b_j$, and
\eqref{eq: uniform logarithmic curvature} applies with
$\boldsymbol\xi=\boldsymbol\delta$ at every point
$\boldsymbol\theta_s$, $0\le s\le1$.

Recall that
$\boldsymbol\theta_s=(1-s)\boldsymbol\theta+s\boldsymbol\theta^\star$,
so $\boldsymbol\theta_0=\boldsymbol\theta$ and
$\boldsymbol\theta_1=\boldsymbol\theta^\star$.
Taylor's theorem with integral remainder, applied from
$\boldsymbol\theta^\star$ to $\boldsymbol\theta$ and using
$\nabla\Phi(\boldsymbol\theta^\star)=\mathbf0$, therefore gives
\begin{equation}\label{eq: logarithmic quadratic growth}
\begin{aligned}
\Phi(\boldsymbol\theta)-\Phi_\star
&=
\int_0^1 s\,
\boldsymbol\delta^T
\nabla^2\Phi(\boldsymbol\theta_s)
\boldsymbol\delta\,ds\\
&\ge
\mu\|\boldsymbol\delta\|_2^2\int_0^1 s\,ds
=
\frac{\mu}{2}
\|\boldsymbol\theta-\boldsymbol\theta^\star\|_2^2.
\end{aligned}
\end{equation}

To prove~\eqref{eq: distance controlled by potential}, evaluate the infimum in~\eqref{eq: averaged logarithmic distance} at $a=1$ for the rescaled vectors $(\mathbf x,\mathbf y)$. Using $m\le n$ and~\eqref{eq: logarithmic quadratic growth} gives
\[
\begin{aligned}
d((\mathbf x,\mathbf y),(\mathbf x^\star,\mathbf y^\star))^2
&\le\frac1m\left\|\log\frac{\mathbf x}{\mathbf x^\star}\right\|_2^2
+\frac1n\left\|\log\frac{\mathbf y}{\mathbf y^\star}\right\|_2^2\\
&\le\frac1m\|\boldsymbol\theta-\boldsymbol\theta^\star\|_2^2
\le\frac{2}{m\mu}\bigl(\Phi(\boldsymbol\theta)-\Phi_\star\bigr).
\end{aligned}
\]
Since $\boldsymbol\theta^\star-\boldsymbol\theta
=-\boldsymbol\delta$, Taylor's theorem with integral remainder,
followed by~\eqref{eq: uniform logarithmic curvature}, gives
\[
\begin{aligned}
\Phi_\star
&=\Phi(\boldsymbol\theta)
-\nabla\Phi(\boldsymbol\theta)^T\boldsymbol\delta\\
&\quad+
\int_0^1(1-s)\,
\boldsymbol\delta^T
\nabla^2\Phi(\boldsymbol\theta_s)
\boldsymbol\delta\,ds\\
&\ge
\Phi(\boldsymbol\theta)
-\nabla\Phi(\boldsymbol\theta)^T\boldsymbol\delta
+\frac{\mu}{2}\|\boldsymbol\delta\|_2^2.
\end{aligned}
\]
Rearranging and completing the square yields
\[
\begin{aligned}
\Phi(\boldsymbol\theta)-\Phi_\star
&\le
\nabla\Phi(\boldsymbol\theta)^T\boldsymbol\delta
-\frac{\mu}{2}\|\boldsymbol\delta\|_2^2
=
\frac{1}{2\mu}\|\nabla\Phi(\boldsymbol\theta)\|_2^2
-\frac{\mu}{2}
\left\|
\boldsymbol\delta-\frac{1}{\mu}\nabla\Phi(\boldsymbol\theta)
\right\|_2^2
\le
\frac{1}{2\mu}\|\nabla\Phi(\boldsymbol\theta)\|_2^2,
\end{aligned}
\]
which proves~\eqref{eq: logarithmic gradient gap}.\qedhere

\end{proof}

\subsubsection{Scalar estimates for candidate decreases}

Define
\begin{equation}\label{eq: scalar decrease definition}
\varphi(u):=u-1-\log u,\qquad u>0.
\end{equation}
The following lemma collects the scalar estimates used in the
population convergence proof and in the comparison of population
and empirical candidate decreases.

\begin{lem}[Scalar estimates for candidate decreases]\label{lem: scalar candidate decrease estimates}
For all $u,v>0$,
\begin{equation}\label{eq: quadratic candidate decrease}
v\varphi(u/v)\ge\frac{(u-v)^2}{2\max\{u,v\}},
\end{equation}
and
\begin{equation}\label{eq:thm2_scalar_decrease}
\varphi(u)\ge(\sqrt u-1)^2\ge0,
\qquad
\varphi(u)=0\quad\Longleftrightarrow\quad u=1.
\end{equation}
For all $u,\widetilde u,v\in(0,1]$,
\begin{equation}\label{eq: quadratic decrease comparison}
v\varphi(u/v)
\le2v\varphi(\widetilde u/v)
+\left[\log\frac{u}{\widetilde u}\right]^2.
\end{equation}
\end{lem}

\begin{proof}
By the definition~\eqref{eq: scalar decrease definition} and the
substitution $s=v+t(u-v)$,
\begin{equation}\label{eq: scalar decrease integral}
\begin{aligned}
v\varphi(u/v)
&=v\left(\frac{u}{v}-1-\log(u/v)\right)\\
&=u-v-v\log(u/v)
=\int_v^u\frac{s-v}{s}\,ds\\
&=(u-v)^2\int_0^1\frac{t}{v+t(u-v)}\,dt.
\end{aligned}
\end{equation}
For $0\le t\le1$,
\begin{equation}\label{eq: scalar interpolation bound}
0<v+t(u-v)\le\max\{u,v\}.
\end{equation}
Substituting~\eqref{eq: scalar interpolation bound} into
\eqref{eq: scalar decrease integral} and using
$\int_0^1t\,dt=1/2$ proves~\eqref{eq: quadratic candidate decrease}.
Taking $v=1$ in~\eqref{eq: quadratic candidate decrease} shows that
$\varphi(u)>0$ for $u\ne1$, while
\eqref{eq: scalar decrease definition} gives $\varphi(1)=0$.
Moreover, the same definition gives
\begin{equation}\label{eq: scalar square root identity}
\varphi(u)-(\sqrt u-1)^2
=2\varphi(\sqrt u)\ge0,
\end{equation}
where the inequality follows from
\eqref{eq: quadratic candidate decrease} with $v=1$ and $u$ replaced
by $\sqrt u$. This proves~\eqref{eq:thm2_scalar_decrease}.

Now let $u,\widetilde u,v\in(0,1]$ and put
\begin{equation}\label{eq: scalar logarithmic difference}
\delta:=\log(u/\widetilde u),
\qquad u=\widetilde u e^\delta.
\end{equation}
Expanding both occurrences of $\varphi$ by
\eqref{eq: scalar decrease definition} and using
\eqref{eq: scalar logarithmic difference} gives the exact identity
\begin{equation}\label{eq: scalar perturbation identity}
v\varphi(u/v)
=v\varphi(\widetilde u/v)
+(\widetilde u-v)\delta
+\widetilde u(e^\delta-1-\delta).
\end{equation}
Integration by parts gives
\begin{equation}\label{eq: scalar exponential remainder}
\begin{aligned}
\widetilde u(e^\delta-1-\delta)
&=\delta^2\int_0^1(1-t)\widetilde u e^{t\delta}\,dt\\
&=\delta^2\int_0^1(1-t)\widetilde u^{1-t}u^t\,dt
\le\frac{\delta^2}{2}.
\end{aligned}
\end{equation}
The second equality uses~\eqref{eq: scalar logarithmic difference};
the inequality uses $u,\widetilde u\le1$, so that
$\widetilde u^{1-t}u^t\le1$ for $0\le t\le1$.

To bound $(\widetilde u-v)\delta$ in
\eqref{eq: scalar perturbation identity}, expand the nonnegative
square $(\widetilde u-v-\delta)^2$ and apply
\eqref{eq: quadratic candidate decrease} with $u=\widetilde u$.
Since $\max\{\widetilde u,v\}\le1$, this yields
\begin{equation}\label{eq: scalar product bound}
\begin{aligned}
2(\widetilde u-v)\delta
&=(\widetilde u-v)^2+\delta^2
-(\widetilde u-v-\delta)^2\\
&\le(\widetilde u-v)^2+\delta^2
\le2v\varphi(\widetilde u/v)+\delta^2.
\end{aligned}
\end{equation}
Combining~\eqref{eq: scalar perturbation identity},
\eqref{eq: scalar exponential remainder}, and
\eqref{eq: scalar product bound} gives
\eqref{eq: quadratic decrease comparison}, with
$\delta$ as defined in~\eqref{eq: scalar logarithmic difference}.
\end{proof}

\subsubsection{Potential decrease under row and column updates}

\begin{lem}[Potential decrease under row and column updates]\label{lem: population candidate decreases}
Fix $t\ge0$ and a positive scaling pair $(\mathbf x^{(t)},\mathbf y^{(t)})$. Let the vectors $(\mathbf g^{(t+1)},\mathbf h^{(t+1)})$ solve~\eqref{eq: Dyson equation} for $(\mathbf x^{(t)},\mathbf y^{(t)})$. The candidate vectors from~\eqref{eq: x_t update} and~\eqref{eq: y_t update} at iteration $t+1$ are
\begin{equation}\label{eq:thm2_candidate_vectors}
x_i'=\frac{\alpha x_i^{(t)}}{1-\eta g_i^{(t+1)}},
\qquad
y_j'=\frac{(\alpha m/n)y_j^{(t)}}{1-\eta h_j^{(t+1)}}.
\end{equation}
With the Dyson vectors $(\mathbf g^{(t+1)},\mathbf h^{(t+1)})$ held fixed, define the row and column decreases by
\begin{equation}\label{eq:thm2_decrease_definitions}
\begin{aligned}
\Delta_{\mathrm{row}}^{(t+1)}
&:=J(\mathbf x^{(t)},\mathbf y^{(t)},\mathbf g^{(t+1)},\mathbf h^{(t+1)})
-J(\mathbf x',\mathbf y^{(t)},\mathbf g^{(t+1)},\mathbf h^{(t+1)}),\\
\Delta_{\mathrm{col}}^{(t+1)}
&:=J(\mathbf x^{(t)},\mathbf y^{(t)},\mathbf g^{(t+1)},\mathbf h^{(t+1)})
-J(\mathbf x^{(t)},\mathbf y',\mathbf g^{(t+1)},\mathbf h^{(t+1)}).
\end{aligned}
\end{equation}
With $\varphi$ defined in~\eqref{eq: scalar decrease definition}, we have
\begin{equation}\label{eq:thm2_candidate_decreases}
\begin{aligned}
\Delta_{\mathrm{row}}^{(t+1)}
&=\alpha\sum_i\varphi\left(\frac{1-\eta g_i^{(t+1)}}{\alpha}\right),\\
\Delta_{\mathrm{col}}^{(t+1)}
&=\frac{\alpha m}{n}\sum_j\varphi\left(\frac{1-\eta h_j^{(t+1)}}{\alpha m/n}\right),
\end{aligned}
\end{equation}
satisfying $\Delta_{\mathrm{row}}^{(t+1)},\Delta_{\mathrm{col}}^{(t+1)}\ge0$, with simultaneous equality if and only if $(\mathbf x^{(t)},\mathbf y^{(t)})$ satisfies~\eqref{eq: scaling equations}. Moreover, for
$\boldsymbol\theta=(\log\mathbf x^{(t)},\log\mathbf y^{(t)},\log\mathbf g^{(t+1)},\log\mathbf h^{(t+1)})$,
\begin{equation}\label{eq: candidate decrease gradient bound}
\max\{\Delta_{\mathrm{row}}^{(t+1)},\Delta_{\mathrm{col}}^{(t+1)}\}
\ge\frac14\|\nabla\Phi(\boldsymbol\theta)\|_2^2.
\end{equation}
\end{lem}

\begin{proof}
Multiply the row and column equations in~\eqref{eq: Dyson equation} by $g_i^{(t+1)}$ and $h_j^{(t+1)}$, respectively, to obtain
\begin{equation}\label{eq:thm2_dyson_relations}
g_i^{(t+1)}\sum_j\widetilde S_{ij}^{(t)}h_j^{(t+1)}=1-\eta g_i^{(t+1)},
\qquad
h_j^{(t+1)}\sum_i\widetilde S_{ij}^{(t)}g_i^{(t+1)}=1-\eta h_j^{(t+1)},
\end{equation}
where $\widetilde S^{(t)}=D\{\mathbf x^{(t)}\}S D\{\mathbf y^{(t)}\}$, as in~\eqref{eq:intro_normalized_variance}. Substituting~\eqref{eq:thm2_candidate_vectors} into~\eqref{eq:thm2_decrease_definitions} and using~\eqref{eq:thm2_dyson_relations} gives~\eqref{eq:thm2_candidate_decreases}. By~\eqref{eq:thm2_scalar_decrease} of
Lemma~\ref{lem: scalar candidate decrease estimates},
$\varphi(u)\ge0$ for $u>0$, with $\varphi(u)=0$ if and only if $u=1$.

We evaluate the gradient of $\Phi$ at
\[
\boldsymbol\theta
=
(\log\mathbf x^{(t)},\log\mathbf y^{(t)},
 \log\mathbf g^{(t+1)},\log\mathbf h^{(t+1)}).
\]
The four logarithmic variable blocks are treated as independent
when differentiating. By the chain rule,
\[
\frac{\partial\Phi}{\partial\log x_i}
=x_i\frac{\partial J}{\partial x_i},
\]
with analogous identities for $y_j$, $g_i$, and $h_j$.
Using
$\widetilde S_{ij}^{(t)}=x_i^{(t)}S_{ij}y_j^{(t)}$,
we obtain the following derivatives from~\eqref{eq: potential fun}.
The last equality in each line follows from the Dyson
relations~\eqref{eq:thm2_dyson_relations}:
\begin{equation}\label{eq: gradient at Dyson point}
\begin{aligned}
\frac{\partial\Phi}{\partial\log x_i}(\boldsymbol\theta)
&=
g_i^{(t+1)}
\sum_j\widetilde S_{ij}^{(t)}h_j^{(t+1)}-\alpha
=
1-\eta g_i^{(t+1)}-\alpha,\\
\frac{\partial\Phi}{\partial\log y_j}(\boldsymbol\theta)
&=
h_j^{(t+1)}
\sum_i\widetilde S_{ij}^{(t)}g_i^{(t+1)}
-\frac{\alpha m}{n}
=
1-\eta h_j^{(t+1)}-\frac{\alpha m}{n},\\
\frac{\partial\Phi}{\partial\log g_i}(\boldsymbol\theta)
&=
g_i^{(t+1)}
\sum_j\widetilde S_{ij}^{(t)}h_j^{(t+1)}
+\eta g_i^{(t+1)}-1
=0,\\
\frac{\partial\Phi}{\partial\log h_j}(\boldsymbol\theta)
&=
h_j^{(t+1)}
\sum_i\widetilde S_{ij}^{(t)}g_i^{(t+1)}
+\eta h_j^{(t+1)}-1
=0.
\end{aligned}
\end{equation} Equations~\eqref{eq:thm2_candidate_decreases},
\eqref{eq:thm2_scalar_decrease}, and
\eqref{eq: gradient at Dyson point} show that
$\Delta_{\mathrm{row}}^{(t+1)}
=\Delta_{\mathrm{col}}^{(t+1)}=0$
if and only if $\nabla\Phi(\boldsymbol\theta)=\mathbf0$.
By the chain rule and positivity of the coordinates,
$\nabla\Phi(\boldsymbol\theta)=\mathbf0$ if and only if
\[
\nabla J(\mathbf x^{(t)},\mathbf y^{(t)},
         \mathbf g^{(t+1)},\mathbf h^{(t+1)})=\mathbf0.
\]
The characterization of stationary points in
Lemma~\ref{lem: properties of potential function},
together with uniqueness of the Dyson solution, shows that
this is equivalent to $(\mathbf x^{(t)},\mathbf y^{(t)})$
satisfying~\eqref{eq: scaling equations}.

We now prove~\eqref{eq: candidate decrease gradient bound}
using~\eqref{eq: quadratic candidate decrease} of
Lemma~\ref{lem: scalar candidate decrease estimates}.

By~\eqref{eq:thm2_dyson_relations} and~\eqref{eq: alpha def},
the values $1-\eta g_i^{(t+1)}$, $1-\eta h_j^{(t+1)}$,
$\alpha$, and $\alpha m/n$ all belong to $(0,1)$.
For the substitutions below, $\max\{u,v\}\le1$ in
\eqref{eq: quadratic candidate decrease}. Applying that inequality
to each summand in~\eqref{eq:thm2_candidate_decreases} therefore gives
the bounds below.
For the row decrease, take
$u=1-\eta g_i^{(t+1)}$ and $v=\alpha$.
Using~\eqref{eq: gradient at Dyson point}, we can write
\[
\begin{aligned}
\Delta_{\mathrm{row}}^{(t+1)}
&=
\alpha\sum_i
\varphi\left(\frac{1-\eta g_i^{(t+1)}}{\alpha}\right)\\
&\ge
\frac12\sum_i
\left(1-\eta g_i^{(t+1)}-\alpha\right)^2\\
&=
\frac12
\|\nabla_{\log\mathbf x}\Phi(\boldsymbol\theta)\|_2^2.
\end{aligned}
\]
For the column decrease, take
$u=1-\eta h_j^{(t+1)}$ and $v=\alpha m/n$.
The same argument gives
\[
\begin{aligned}
\Delta_{\mathrm{col}}^{(t+1)}
&=
\frac{\alpha m}{n}\sum_j
\varphi\left(
\frac{1-\eta h_j^{(t+1)}}{\alpha m/n}
\right)\\
&\ge
\frac12\sum_j
\left(1-\eta h_j^{(t+1)}-\frac{\alpha m}{n}\right)^2\\
&=
\frac12
\|\nabla_{\log\mathbf y}\Phi(\boldsymbol\theta)\|_2^2.
\end{aligned}
\]

The remaining gradient blocks satisfy
$\nabla_{\log\mathbf g}\Phi(\boldsymbol\theta)=\mathbf0_m$
and
$\nabla_{\log\mathbf h}\Phi(\boldsymbol\theta)=\mathbf0_n$
by~\eqref{eq: gradient at Dyson point}.
Combining the two bounds above and using the fact that
the maximum of two numbers is at least their average, we obtain
\[
\begin{aligned}
\max\{\Delta_{\mathrm{row}}^{(t+1)},
      \Delta_{\mathrm{col}}^{(t+1)}\}
&\ge
\frac12\bigl(
\Delta_{\mathrm{row}}^{(t+1)}
+\Delta_{\mathrm{col}}^{(t+1)}
\bigr)\\
&\ge
\frac14\bigl(
\|\nabla_{\log\mathbf x}\Phi(\boldsymbol\theta)\|_2^2
+\|\nabla_{\log\mathbf y}\Phi(\boldsymbol\theta)\|_2^2
\bigr)\\
&=
\frac14\|\nabla\Phi(\boldsymbol\theta)\|_2^2.
\end{aligned}
\]
\end{proof}

\subsection{Main proof}

\begin{proof}[Proof of Theorem~\ref{thm:population_convergence}]
For every positive pair $(\mathbf x^{(t)},\mathbf y^{(t)})$, Lemma~\ref{lem: properties of potential function} supplies a unique positive Dyson solution $(\mathbf g^{(t+1)},\mathbf h^{(t+1)})$, with $g_i^{(t+1)},h_j^{(t+1)}<1/\eta$. The updates~\eqref{eq: x_t update}--\eqref{eq: y_t update} therefore produce a positive pair $(\mathbf x^{(t+1)},\mathbf y^{(t+1)})$. Starting from positive $(\mathbf x^{(0)},\mathbf y^{(0)})$, the procedure is well-defined at every iteration.

\emph{Monotonicity of the potential.}
For $t\ge1$, $(\mathbf g^{(t+1)},\mathbf h^{(t+1)})$ minimizes $J(\mathbf x^{(t)},\mathbf y^{(t)},\mathbf g,\mathbf h)$ over positive $(\mathbf g,\mathbf h)$ by Lemma~\ref{lem: properties of potential function}. Comparing with $(\mathbf g^{(t)},\mathbf h^{(t)})$ gives
\begin{equation}\label{eq:thm2_dyson_descent}
J(\mathbf x^{(t)},\mathbf y^{(t)},\mathbf g^{(t+1)},\mathbf h^{(t+1)})
\le J(\mathbf x^{(t)},\mathbf y^{(t)},\mathbf g^{(t)},\mathbf h^{(t)}).
\end{equation}
Let $\Delta_{\mathrm{row}}^{(t+1)}$ and
$\Delta_{\mathrm{col}}^{(t+1)}$ denote the row and column
potential decreases defined
in~\eqref{eq:thm2_decrease_definitions} of
Lemma~\ref{lem: population candidate decreases},
with $(\mathbf g^{(t+1)},\mathbf h^{(t+1)})$ held fixed.
By~\eqref{eq: condition for updating row scaling vector},
the selected scaling update decreases $J$ by
$\max\{\Delta_{\mathrm{row}}^{(t+1)},
       \Delta_{\mathrm{col}}^{(t+1)}\}$. Thus,~\eqref{eq: normalized potential fun} gives
\begin{equation}\label{eq:thm2_descent_bound}
m(J_t-J_{t+1})\ge
\max\{\Delta_{\mathrm{row}}^{(t+1)},\Delta_{\mathrm{col}}^{(t+1)}\}\ge0.
\end{equation}
By Lemma~\ref{lem: population candidate decreases}, the maximum in~\eqref{eq:thm2_descent_bound} is positive whenever $(\mathbf x^{(t)},\mathbf y^{(t)})$ fails to satisfy~\eqref{eq: scaling equations}. This proves the monotonicity assertion in part~\ref{thm: population qualitative}.

\emph{Geometric convergence under Assumption~\ref{assump: population bounds}.}
Assume that $\mathbf x^{(0)}=\mathbf1_m$ and $\mathbf y^{(0)}=\mathbf1_n$. By Lemma~\ref{lem: finite-time bounds population}, $(\mathbf x^{(t)},\mathbf y^{(t)},\mathbf g^{(t+1)},\mathbf h^{(t+1)})$ satisfies~\eqref{eq: population curvature hypotheses} for every $t\ge0$. Combining~\eqref{eq: logarithmic gradient gap} with~\eqref{eq: candidate decrease gradient bound} gives
\begin{equation}\label{eq: quantitative greedy decrease}
\max\{\Delta_{\mathrm{row}}^{(t+1)},\Delta_{\mathrm{col}}^{(t+1)}\}
\ge\frac\mu2\bigl[
J(\mathbf x^{(t)},\mathbf y^{(t)},\mathbf g^{(t+1)},\mathbf h^{(t+1)})-\Phi_\star\bigr].
\end{equation}
Equation~\eqref{eq: quantitative greedy decrease} holds for any choices of row or column updates at iterations $1,\ldots,t$, because Lemma~\ref{lem: finite-time bounds population} covers every such sequence. For the greedy choice at iteration $t+1$, the definitions~\eqref{eq: normalized potential fun} and~\eqref{eq:thm2_decrease_definitions}, followed by~\eqref{eq: quantitative greedy decrease} and~\eqref{eq:thm2_dyson_descent}, give
\begin{equation}\label{eq: normalized potential contraction}
\begin{aligned}
J_{t+1}-J_\star
&=\frac1m\bigl[J(\mathbf x^{(t)},\mathbf y^{(t)},\mathbf g^{(t+1)},\mathbf h^{(t+1)})-\Phi_\star\bigr]\\
&\quad-\frac1m\max\{\Delta_{\mathrm{row}}^{(t+1)},\Delta_{\mathrm{col}}^{(t+1)}\}\\
&\le\frac{1-\mu/2}{m}\bigl[J(\mathbf x^{(t)},\mathbf y^{(t)},\mathbf g^{(t+1)},\mathbf h^{(t+1)})-\Phi_\star\bigr]\\
&\le(1-\mu/2)(J_t-J_\star),\qquad t\ge1.
\end{aligned}
\end{equation}
Lemma~\ref{lem: finite-time bounds population} also gives~\eqref{eq: population curvature hypotheses} for $(\mathbf x^{(t)},\mathbf y^{(t)},\mathbf g^{(t)},\mathbf h^{(t)})$, so~\eqref{eq: distance controlled by potential} yields
\begin{equation}\label{eq: population iterate error bound}
d((\mathbf x^{(t)},\mathbf y^{(t)}),(\mathbf x^\star,\mathbf y^\star))^2
\le\frac2\mu(J_t-J_\star),\qquad t\ge1.
\end{equation}

The Dyson minimization and scaling update at iteration $1$ give the first inequality below. The equality uses $g_i^\star h_j^\star=\alpha$, from~\eqref{eq: formulas for g_star and h_star} and~\eqref{eq: critical point analysis 3}, and the definition~\eqref{eq: logarithmic barrier function}:
\[
\begin{aligned}
J_1-J_\star
&\le\frac1m\bigl[J(\mathbf1_m,\mathbf1_n,\mathbf g^\star,\mathbf h^\star)-\Phi_\star\bigr]\\
&=\frac\alpha m\bigl[L(\mathbf1_m,\mathbf1_n)-L(\mathbf x^\star,\mathbf y^\star)\bigr]
\le K_0.
\end{aligned}
\]
The bound $K_0>0$ depends only on $\eta,c,C,\gamma_-,\gamma_+$: indeed, $\sum_{i,j}S_{ij}\le Cm$, $(\mathbf x^\star)^TS\mathbf y^\star=m$, and~\eqref{eq: bounded optimal scalings} bounds $\log x_i^\star$ and $\log y_j^\star$. Iterating~\eqref{eq: normalized potential contraction} and applying~\eqref{eq: population iterate error bound}, with $\rho=\sqrt{1-\mu/2}\in(0,1)$, gives
\begin{equation}\label{eq: population proof rates}
0\le J_t-J_\star\le K_0\rho^{2(t-1)},
\qquad
d((\mathbf x^{(t)},\mathbf y^{(t)}),(\mathbf x^\star,\mathbf y^\star))
\le\sqrt{2K_0/\mu}\,\rho^{t-1},\quad t\ge1.
\end{equation}
Choosing $K>0$ with $K\ge\max\{K_0/\rho^2,\sqrt{2K_0/\mu}/\rho\}$ gives~\eqref{eq: population geometric rates}, proving part~\ref{thm: population quantitative}.

\emph{Convergence for any positive profile and initialization.}
We first consider an arbitrary positive matrix $S$ with
$\mathbf x^{(0)}=\mathbf1_m$ and
$\mathbf y^{(0)}=\mathbf1_n$.
For fixed $m,n$ and $S$, the bounds used in the preceding
argument hold with
\[
c=n\min_{i,j}S_{ij}>0,
\qquad
C=n\max_{i,j}S_{ij}<\infty,
\qquad
\gamma_-=\gamma_+=m/n.
\]
Thus~\eqref{eq: population proof rates} applies, with constants
that may depend on $S,m,n$, and gives convergence in the
distance~\eqref{eq: averaged logarithmic distance}.
Uniform bounds over different matrices are therefore unnecessary
for convergence with $S,m,n$ fixed.

We next reduce an arbitrary positive initialization to the
all-ones initialization. Define
\[
S_0=D\{\mathbf x^{(0)}\}S D\{\mathbf y^{(0)}\},
\qquad
\mathbf u^{(t)}=\frac{\mathbf x^{(t)}}{\mathbf x^{(0)}},
\qquad
\mathbf v^{(t)}=\frac{\mathbf y^{(t)}}{\mathbf y^{(0)}},
\]
where the quotients are taken entrywise.
Then $\mathbf u^{(0)}=\mathbf1_m$,
$\mathbf v^{(0)}=\mathbf1_n$, and
\[
D\{\mathbf u^{(t)}\}S_0D\{\mathbf v^{(t)}\}
=
D\{\mathbf x^{(t)}\}SD\{\mathbf y^{(t)}\}.
\]
Since the scaled variance matrices agree, the Dyson
equations~\eqref{eq: Dyson equation} give the same vectors
$(\mathbf g^{(t+1)},\mathbf h^{(t+1)})$ in both formulations.
Dividing the updates~\eqref{eq: x_t update}
and~\eqref{eq: y_t update} by the fixed initial coordinates
gives the corresponding updates for $\mathbf u^{(t)}$
and $\mathbf v^{(t)}$.
The selection rule~\eqref{eq: condition for updating row scaling vector}
also makes the same decision.
Consequently, $(\mathbf u^{(t)},\mathbf v^{(t)})$ is precisely
the population sequence for $S_0$ initialized at
$(\mathbf1_m,\mathbf1_n)$.

Define
\[
\mathbf u^\star=\frac{\mathbf x^\star}{\mathbf x^{(0)}},
\qquad
\mathbf v^\star=\frac{\mathbf y^\star}{\mathbf y^{(0)}}.
\]
The identity
\[
D\{\mathbf u^\star\}S_0D\{\mathbf v^\star\}
=
D\{\mathbf x^\star\}SD\{\mathbf y^\star\}
\]
shows that $(\mathbf u^\star,\mathbf v^\star)$ solves the
scaling equations for $S_0$.
Since $S_0$ is positive, the convergence result for all-ones
initialization established above applies to this sequence.
Moreover,
$\mathbf u^{(t)}/\mathbf u^\star
=\mathbf x^{(t)}/\mathbf x^\star$ and
$\mathbf v^{(t)}/\mathbf v^\star
=\mathbf y^{(t)}/\mathbf y^\star$.
Hence, by~\eqref{eq: averaged logarithmic distance},
\[
\begin{aligned}
d((\mathbf x^{(t)},\mathbf y^{(t)}),
  (\mathbf x^\star,\mathbf y^\star))
=
d((\mathbf u^{(t)},\mathbf v^{(t)}),
  (\mathbf u^\star,\mathbf v^\star))
\longrightarrow0.
\end{aligned}
\]

To obtain convergence of the scaling vectors themselves,
choose $a_t>0$ attaining the infimum
in~\eqref{eq: averaged logarithmic distance}.
Such a minimizer exists because the expression being minimized
is a strictly convex quadratic in $\log a$.
By its definition,
\[
\begin{aligned}
&\frac1m
\left\|\log\frac{a_t\mathbf x^{(t)}}{\mathbf x^\star}\right\|_2^2
+
\frac1n
\left\|\log\frac{a_t^{-1}\mathbf y^{(t)}}{\mathbf y^\star}\right\|_2^2\\
&\qquad=
d((\mathbf x^{(t)},\mathbf y^{(t)}),
  (\mathbf x^\star,\mathbf y^\star))^2
\longrightarrow0.
\end{aligned}
\]
Because $m,n$ are fixed, every coordinate of the two logarithmic
error vectors tends to zero. Exponentiating gives
\[
(a_t\mathbf x^{(t)},a_t^{-1}\mathbf y^{(t)})
\longrightarrow
(\mathbf x^\star,\mathbf y^\star),
\]
which proves part~\ref{thm: population qualitative}.
\end{proof}

\section[Derivation of the formula for Jt]{Derivation of the formula for $J_t$ in~\eqref{eq: normalized potential fun} } \label{appendix: derivation of J_t formula}
According to the definition of the potential function $J$ in~\eqref{eq: potential fun}, we have
\begin{align}   J(\mathbf{x}^{(t)},\mathbf{y}^{(t)},\mathbf{g}^{(t)},\mathbf{h}^{(t)})
    &= \sum_{i=1}^m \sum_{j=1}^n \mathbf{g}_i^{(t)} \mathbf{x}_i^{(t)} S_{i,j} \mathbf{y}_j^{(t)} \mathbf{h}_j^{(t)} \nonumber \\
    &+ \sum_{i=1}^m \left( \eta \mathbf{g}^{(t)}_i - \log (\mathbf{g}^{(t)}_i) - \alpha \log (\mathbf{x}^{(t)}_i) \right) + \sum_{j=1}^n \left(\eta \mathbf{h}^{(t)}_j - \log (\mathbf{h}^{(t)}_j) - \alpha \frac{m}{n} \log (\mathbf{y}^{(t)}_j) \right).
\end{align}
If the update in iteration $t$ involved the row scaling vector $\mathbf{x}^{(t)}$ according to~\eqref{eq: x_t update}, we can write
\begin{align}
    \sum_{i=1}^m \sum_{j=1}^n \mathbf{g}_i^{(t)} \mathbf{x}_i^{(t)} S_{i,j} \mathbf{y}_j^{(t)} \mathbf{h}_j^{(t)}
    &= \sum_{i=1}^m \sum_{j=1}^n \mathbf{g}_i^{(t)} \mathbf{x}^{(t-1)}_i \left( \frac{\alpha }{1 - \eta {\mathbf{g}}^{(t)}_i} \right) S_{i,j} \mathbf{y}_j^{(t-1)} \mathbf{h}_j^{(t)} \nonumber \\
    &= \sum_{i=1}^m \sum_{j=1}^n \mathbf{g}_i^{(t)} \mathbf{x}^{(t-1)}_i \left( \frac{\alpha }{\sum_{k=1}^n \mathbf{g}_i^{(t)} \mathbf{x}_i^{(t-1)} S_{ik} \mathbf{y}_k^{(t-1)} \mathbf{h}_k^{(t)}} \right) S_{i,j} \mathbf{y}_j^{(t-1)} \mathbf{h}_j^{(t)} \nonumber \\
    &= \alpha m,
\end{align}
where in the second equality we used the fact that according to our proposed procedure, the pair of vectors $({\mathbf{g}^{(t)}},{\mathbf{h}^{(t)}})$ is the pair of positive vectors $(\overline{\mathbf{g}},\overline{\mathbf{h}})$ that solves~\eqref{eq: Dyson equation} for $\overline{\mathbf{x}} = \mathbf{x}^{(t-1)}$ and $\overline{\mathbf{y}} = \mathbf{y}^{(t-1)}$, hence $1 - \eta {\mathbf{g}}^{(t)}_i = \sum_{j=1}^n \mathbf{g}_i^{(t)} \mathbf{x}_i^{(t-1)} S_{i,j} \mathbf{y}_j^{(t-1)} \mathbf{h}_j^{(t)}$ for all $i\in [m]$.

Similarly, if the update in iteration $t$ involved the column scaling vector $\mathbf{y}^{(t)}$ according to~\eqref{eq: y_t update}, we obtain
\begin{align}
    \sum_{i=1}^m \sum_{j=1}^n \mathbf{g}_i^{(t)} \mathbf{x}_i^{(t)} S_{i,j} \mathbf{y}_j^{(t)} \mathbf{h}_j^{(t)}
    &= \sum_{i=1}^m \sum_{j=1}^n \mathbf{g}_i^{(t)} \mathbf{x}^{(t-1)}_i \left( \frac{\alpha \frac{m}{n} }{1 - \eta {\mathbf{h}}^{(t)}_j} \right) S_{i,j} \mathbf{y}_j^{(t-1)} \mathbf{h}_j^{(t)} \nonumber \\
    &= \sum_{i=1}^m \sum_{j=1}^n \mathbf{g}_i^{(t)} \mathbf{x}^{(t-1)}_i \left( \frac{\alpha \frac{m}{n}}{\sum_{\ell=1}^m \mathbf{g}_\ell^{(t)} \mathbf{x}_\ell^{(t-1)} S_{\ell j} \mathbf{y}_j^{(t-1)} \mathbf{h}_j^{(t)}} \right) S_{i,j} \mathbf{y}_j^{(t-1)} \mathbf{h}_j^{(t)} \nonumber \\
    &= \alpha m.
\end{align}
Finally, combining all of the above, we get
\begin{align}
    J_t &= \frac{J(\mathbf{x}^{(t)},\mathbf{y}^{(t)},\mathbf{g}^{(t)},\mathbf{h}^{(t)})}{m} - \alpha \nonumber \\
    &= \frac{1}{m}\sum_{i=1}^m \left( \eta \mathbf{g}^{(t)}_i - \log (\mathbf{g}^{(t)}_i) - \alpha \log (\mathbf{x}^{(t)}_i) \right) + \frac{1}{m} \sum_{j=1}^n \left(\eta \mathbf{h}^{(t)}_j - \log (\mathbf{h}^{(t)}_j) - \alpha \frac{m}{n} \log (\mathbf{y}^{(t)}_j) \right).
\end{align}

\section{Proof of Theorem~\ref{thm: main theoretical guarantees}}
\label{appendix:proof_thm_main}

\subsection{Supporting lemmas}

For real symmetric matrices $A$ and $B$ of the same size,
we write $A\preceq B$ (equivalently, $B\succeq A$)
if $B-A$ is positive semidefinite.

\subsubsection{Entrywise concentration of the data resolvent} \label{appendix: lem:resolvent_supnorm_concentration}
\begin{lem} [\cite{landa2025dyson}]
\label{lem:resolvent_supnorm_concentration}
Let $\mathbf{x}$ and $\mathbf{y}$ be positive deterministic vectors satisfying
\[
0<c_x \le \mathbf x_i \le C_x,\qquad 0<c_y \le \mathbf y_j \le C_y \qquad \forall i\in[m],\,j\in[n],
\]
for some global constants $c_x,C_x,c_y,C_y > 0$. Let $\widetilde{Y}$, $\widetilde{X}$, $\widetilde{E}$ be the diagonally scaled data, signal, and noise matrices in~\eqref{eq: Y_tilde def}, and let $\widetilde{S}$ be the corresponding variance profile in~\eqref{eq:intro_normalized_variance}.
Let $(\mathbf g,\mathbf h)$ be the positive Dyson solution~\eqref{eq: real-valued Dyson} for $\widetilde S$, and let $(\hat{\mathbf g},\hat{\mathbf h})$ be its row and column resolvent estimates computed from $\widetilde Y$ according to~\eqref{eq: right resolvent formula} and~\eqref{eq: left resolvent formula}. Then, under Assumption~\ref{assump: population bounds} and the moment and rank conditions of Assumption~\ref{assump: main assumption}, for any $\varepsilon\in(0,\delta_0)$ and any $\tau>0$, there exists a constant $K>0$
(depending only on $\eta,\varepsilon,\tau$, $c_x,C_x,c_y,C_y$, and the global constants in these conditions)
such that, with probability at least $1-Kn^{-\tau}$,
\[
\max \left\{\|\hat{\mathbf g}-\mathbf g\|_\infty , \; \|\hat{\mathbf h}-\mathbf h\|_\infty \right\}
\;\le\;
K\bigl(n^{\varepsilon-1/2}+\kappa_n\bigr).
\]
\end{lem}

\begin{proof}
If \(X\) is random, we condition on it, hence the noise distribution is unchanged due to the independence of $X$ and $E$. Therefore, the bounds below are uniform over signal realizations satisfying the stated assumptions. We apply Theorem~6 of~\cite{landa2025dyson} to the scaled model
$\widetilde Y=\widetilde X+\widetilde E$.
We verify the noise moment bounds and signal rank condition
required by that theorem, and then bound the squared row norms
of the scaled signal's singular-vector matrices, which enter
its error estimate.

The diagonal scaling of the data in~\eqref{eq: Y_tilde def} multiplies $E_{ij}$ by $\sqrt{x_i y_j}$, while the scaled variance profile in~\eqref{eq:intro_normalized_variance} satisfies $\widetilde S_{ij}=x_i y_j S_{ij}$. Since the scaling is deterministic, independence and zero means are preserved.
The moment and variance bounds become
\[
\begin{aligned}
\mathbb E|\sqrt n\,\widetilde E_{ij}|^q
&=(x_i y_j)^{q/2}\mathbb E|\sqrt n\,E_{ij}|^q
\le(C_xC_y)^{q/2}\mu_q \qquad(q\in\mathbb N),\\
\frac{c_xc_yc}{n}&\le\widetilde S_{ij}
\le\frac{C_xC_yC}{n}.
\end{aligned}
\]
Since both diagonal scaling matrices are invertible,
\[
\operatorname{rank}(\widetilde X)
=\operatorname{rank}(X)=r
\le\widetilde C_0n^{1/2-\delta_0}.
\]
Thus the moment constants change only by factors depending
on the scaling bounds, while the rank condition is unchanged.

Let $\widetilde U,\widetilde V$ be the singular-vector matrices
of $X$ in~\eqref{eq: SVD of X}, and let $U_{\mathrm{sc}},V_{\mathrm{sc}}$
be those of $\widetilde X$. Their columns are orthonormal:
\[
U_{\mathrm{sc}}^TU_{\mathrm{sc}}
=V_{\mathrm{sc}}^TV_{\mathrm{sc}}=I_r.
\]
Right multiplication by an invertible diagonal matrix preserves
the column span. Applying this fact to $\widetilde X$ and
$\widetilde X^T$ gives
\[
\begin{aligned}
\operatorname{col}(U_{\mathrm{sc}})
&=\operatorname{col}(\widetilde X)
=\operatorname{col}\bigl(D\{\mathbf x\}^{1/2}\widetilde U\bigr),\\
\operatorname{col}(V_{\mathrm{sc}})
&=\operatorname{col}(\widetilde X^T)
=\operatorname{col}\bigl(D\{\mathbf y\}^{1/2}\widetilde V\bigr).
\end{aligned}
\]
The orthogonal projectors onto these two subspaces are therefore
\[
\begin{aligned}
P_x:=U_{\mathrm{sc}}U_{\mathrm{sc}}^T
&=D\{\mathbf x\}^{1/2}\widetilde U
(\widetilde U^TD\{\mathbf x\}\widetilde U)^{-1}
\widetilde U^TD\{\mathbf x\}^{1/2},\\
P_y:=V_{\mathrm{sc}}V_{\mathrm{sc}}^T
&=D\{\mathbf y\}^{1/2}\widetilde V
(\widetilde V^TD\{\mathbf y\}\widetilde V)^{-1}
\widetilde V^TD\{\mathbf y\}^{1/2}.
\end{aligned}
\]
The original singular-vector matrices have orthonormal columns,
so
\[
\widetilde U^TD\{\mathbf x\}\widetilde U\succeq c_xI_r,
\qquad
\widetilde V^TD\{\mathbf y\}\widetilde V\succeq c_yI_r.
\]
Inverting these inequalities and taking projector diagonal
entries yields
\[
\begin{aligned}
\sum_{k=1}^r(U_{\mathrm{sc}})_{ik}^2
=(P_x)_{ii}
&\le\frac{x_i}{c_x}\sum_{k=1}^r\widetilde U_{ik}^2
\le\frac{C_x}{c_x}\sum_{k=1}^r\widetilde U_{ik}^2,\\
\sum_{k=1}^r(V_{\mathrm{sc}})_{jk}^2
=(P_y)_{jj}
&\le\frac{y_j}{c_y}\sum_{k=1}^r\widetilde V_{jk}^2
\le\frac{C_y}{c_y}\sum_{k=1}^r\widetilde V_{jk}^2.
\end{aligned}
\]
Consequently, the localization parameter of $\widetilde X$
satisfies
\[
\widetilde\kappa_n
:=\max\left\{\max_i(P_x)_{ii},\max_j(P_y)_{jj}\right\}
\le\max\{C_x/c_x,C_y/c_y\}\kappa_n.
\]

The unchanged rank condition permits every
$\varepsilon\in(0,\delta_0)$ in Theorem~6
of~\cite{landa2025dyson}. Apply that theorem to $\widetilde Y$
with confidence exponent $\tau+1$, and use the preceding
bound on $\widetilde\kappa_n$. There is a constant $K_{\mathrm D}>0$
with the stated parameter dependence such that
\[
\begin{aligned}
\mathbb P\left(
|\hat g_i-g_i|>K_{\mathrm D}(n^{\varepsilon-1/2}+\kappa_n)
\right)&\le K_{\mathrm D}n^{-(\tau+1)},\\
\mathbb P\left(
|\hat h_j-h_j|>K_{\mathrm D}(n^{\varepsilon-1/2}+\kappa_n)
\right)&\le K_{\mathrm D}n^{-(\tau+1)}.
\end{aligned}
\]
These estimates are uniform in the signal singular values.
A union bound over the $m+n\le2n$ coordinates gives
\[
\begin{aligned}
\mathbb P\left(
\max\{\|\hat{\mathbf g}-\mathbf g\|_\infty,
      \|\hat{\mathbf h}-\mathbf h\|_\infty\}
>K_{\mathrm D}(n^{\varepsilon-1/2}+\kappa_n)
\right)
&\le(m+n)K_{\mathrm D}n^{-(\tau+1)}\\
&\le2K_{\mathrm D}n^{-\tau}.
\end{aligned}
\]
Taking $K\ge2K_{\mathrm D}$ proves the lemma.
\end{proof}

\subsubsection{Nonexpansiveness of the empirical updates}
\begin{lem}\label{lem: empirical update stability}
Suppose that $Y$ has no zero rows or columns. For positive scaling pairs
$(\mathbf x,\mathbf y)$ and $(\mathbf u,\mathbf v)$, let
$(\hat{\mathbf g},\hat{\mathbf h})$ and
$(\widetilde{\mathbf g},\widetilde{\mathbf h})$ be the row and column resolvent estimates
\eqref{eq: right resolvent formula}--\eqref{eq: left resolvent formula}
at these respective pairs, using the diagonal scaling of the data
in~\eqref{eq: Y_tilde def}.
Let $(\mathbf x',\mathbf y')$ and $(\mathbf u',\mathbf v')$ be the
outputs of the same prescribed row or column update
\eqref{eq: x_t update} or~\eqref{eq: y_t update}, using the corresponding
resolvent estimates in the update multipliers.
For
\[
e:=\max\left\{
\left\|\log\frac{\mathbf x}{\mathbf u}\right\|_\infty,
\left\|\log\frac{\mathbf y}{\mathbf v}\right\|_\infty\right\},
\]
both updates are well-defined and satisfy
\begin{equation}\label{eq: nonexpansive empirical updates}
\max\left\{
\left\|\log\frac{\mathbf x'}{\mathbf u'}\right\|_\infty,
\left\|\log\frac{\mathbf y'}{\mathbf v'}\right\|_\infty\right\}\le e.
\end{equation}
Their resolvent denominators satisfy
\begin{equation}\label{eq: empirical denominator stability}
\max\left\{
\left\|\log\frac{\mathbf1_m-\eta\hat{\mathbf g}}
                 {\mathbf1_m-\eta\widetilde{\mathbf g}}\right\|_\infty,
\left\|\log\frac{\mathbf1_n-\eta\hat{\mathbf h}}
                 {\mathbf1_n-\eta\widetilde{\mathbf h}}\right\|_\infty
\right\}\le2e.
\end{equation}
\end{lem}

\begin{proof}
We first consider the case where both scaling pairs undergo
the row update~\eqref{eq: x_t update}. Then
$\mathbf y'=\mathbf y$ and $\mathbf v'=\mathbf v$.
Let $\mathbf a_i$ be row $i$ of $Y$ and define
\[
M:=\eta^2D\{\mathbf y^{-1}\}+Y^TD\{\mathbf x\}Y,
\qquad
N:=\eta^2D\{\mathbf v^{-1}\}+Y^TD\{\mathbf u\}Y.
\]
Both matrices are positive definite. By the definition of the row
resolvent estimate $\hat g_i$ in~\eqref{eq: right resolvent formula},
\[
\begin{aligned}
1-\eta\hat g_i
&=1-\eta^2\left[(\widetilde Y\widetilde Y^T+\eta^2I_m)^{-1}\right]_{ii}\\
&=\left[\widetilde Y(\widetilde Y^T\widetilde Y+\eta^2I_n)^{-1}
\widetilde Y^T\right]_{ii},
\end{aligned}
\]
The diagonal scaling of the data in~\eqref{eq: Y_tilde def} gives
\[
\begin{aligned}
\widetilde Y^T\widetilde Y+\eta^2I_n
&=D\{\mathbf y\}^{1/2}M D\{\mathbf y\}^{1/2},\\
(\widetilde Y^T\widetilde Y+\eta^2I_n)^{-1}
&=D\{\mathbf y\}^{-1/2}M^{-1}D\{\mathbf y\}^{-1/2}.
\end{aligned}
\]
Since row $i$ of $\widetilde Y$ is
$\sqrt{x_i}\,\mathbf a_iD\{\mathbf y\}^{1/2}$, substitution into the
diagonal expression above cancels the $D\{\mathbf y\}^{1/2}$ factors:
\begin{equation}\label{eq: empirical row update inverse formula}
\begin{aligned}
1-\eta\hat g_i
&=x_i\mathbf a_iD\{\mathbf y\}^{1/2}
(\widetilde Y^T\widetilde Y+\eta^2I_n)^{-1}
D\{\mathbf y\}^{1/2}\mathbf a_i^T\\
&=x_i\mathbf a_iM^{-1}\mathbf a_i^T.
\end{aligned}
\end{equation}
The row update~\eqref{eq: x_t update} multiplies $x_i$ by
$\alpha/(1-\eta\hat g_i)$. Cancelling $x_i$, and making the same
substitution at $(\mathbf u,\mathbf v)$, therefore gives
\[
x_i'=\frac{\alpha}{\mathbf a_iM^{-1}\mathbf a_i^T},
\qquad
u_i'=\frac{\alpha}{\mathbf a_iN^{-1}\mathbf a_i^T}.
\]
These quadratic forms are positive because $\mathbf a_i\ne0$.

The definition of $e$ gives
\[
e^{-e}u_i\le x_i\le e^eu_i,
\qquad e^{-e}v_j^{-1}\le y_j^{-1}\le e^ev_j^{-1}.
\]
These coordinate inequalities imply
\[
\begin{aligned}
e^{-e}\eta^2D\{\mathbf v^{-1}\}
&\preceq\eta^2D\{\mathbf y^{-1}\}
\preceq e^e\eta^2D\{\mathbf v^{-1}\},\\
e^{-e}Y^TD\{\mathbf u\}Y
&\preceq Y^TD\{\mathbf x\}Y
\preceq e^eY^TD\{\mathbf u\}Y.
\end{aligned}
\]
Adding these two displayed bounds gives
\[
e^{-e}N\preceq M\preceq e^eN.
\]
Since $M,N\succ0$, inverting each bound separately yields
\[
\begin{aligned}
M\preceq e^eN
&\quad\Longrightarrow\quad e^{-e}N^{-1}\preceq M^{-1},\\
e^{-e}N\preceq M
&\quad\Longrightarrow\quad M^{-1}\preceq e^eN^{-1}.
\end{aligned}
\]
Applying the inverse bounds to $\mathbf a_i^T$ yields
\[
e^{-e}\le\frac{x_i'}{u_i'}
=\frac{\mathbf a_iN^{-1}\mathbf a_i^T}
      {\mathbf a_iM^{-1}\mathbf a_i^T}\le e^e.
\]
For the unchanged column coordinates, $y_j^{\prime}/v_j^{\prime}=y_j/v_j$,
so the same bounds hold.
Multiplying the quadratic-form ratio by $x_i/u_i$ also gives
\begin{equation}\label{eq: empirical denominator quadratic form ratio}
e^{-2e}\le \frac{x_i\mathbf a_iM^{-1}\mathbf a_i^T}
      {u_i\mathbf a_iN^{-1}\mathbf a_i^T} =
\frac{1-\eta\hat g_i}{1-\eta\widetilde g_i}
\le e^{2e}.
\end{equation}
The column resolvent formula~\eqref{eq: left resolvent formula}
and the column update~\eqref{eq: y_t update} have the same form
after replacing
$Y,\mathbf x,\mathbf y,\mathbf u,\mathbf v,\alpha$ with
$Y^T,\mathbf y,\mathbf x,\mathbf v,\mathbf u,\alpha m/n$, respectively.
This proves nonexpansiveness~\eqref{eq: nonexpansive empirical updates}
and the denominator bound~\eqref{eq: empirical denominator stability}.
\end{proof}

\subsubsection{Comparison of analytical and empirical updates}

Suppose that $m\le n$ and that $Y\in\mathbb R^{m\times n}$
has no zero rows or columns. Fix $\eta>0$ and
$(\mathbf x,\mathbf y)\in(0,\infty)^m\times(0,\infty)^n$.
Let $\widetilde Y$ be the diagonally scaled data matrix
in~\eqref{eq: Y_tilde def} for this pair, and define
$\hat{\mathbf g}$ and $\hat{\mathbf h}$ by its row and column
resolvent formulas~\eqref{eq: right resolvent formula}
and~\eqref{eq: left resolvent formula}, respectively.
Let $\alpha\in(0,1)$ be as in~\eqref{eq: alpha def}.

For arbitrary
$(\mathbf g,\mathbf h)\in(0,1/\eta)^m\times(0,1/\eta)^n$,
define the quantities below using $\varphi$
from~\eqref{eq: scalar decrease definition} and the candidate
decrease formulas~\eqref{eq:thm2_candidate_decreases}:
\begin{align}
\Delta_{\mathrm{row}}
&:=\alpha\sum_{i=1}^m
\varphi\!\left(\frac{1-\eta g_i}{\alpha}\right),
&
\hat\Delta_{\mathrm{row}}
&:=\alpha\sum_{i=1}^m
\varphi\!\left(\frac{1-\eta\hat g_i}{\alpha}\right),
\label{eq: row candidate definitions}\\
\Delta_{\mathrm{col}}
&:=\frac{\alpha m}{n}\sum_{j=1}^n
\varphi\!\left(\frac{1-\eta h_j}{\alpha m/n}\right),
&
\hat\Delta_{\mathrm{col}}
&:=\frac{\alpha m}{n}\sum_{j=1}^n
\varphi\!\left(\frac{1-\eta\hat h_j}{\alpha m/n}\right).
\label{eq: column candidate definitions}
\end{align}
Define the maximum logarithmic discrepancy between the denominators by
\begin{equation}\label{eq: candidate logarithmic denominator error}
e:=\max\left\{
\left\|\log\frac{\mathbf1_m-\eta\hat{\mathbf g}}
                 {\mathbf1_m-\eta\mathbf g}\right\|_\infty,
\left\|\log\frac{\mathbf1_n-\eta\hat{\mathbf h}}
                 {\mathbf1_n-\eta\mathbf h}\right\|_\infty
\right\}.
\end{equation}
Write the empirical selection rule
\eqref{eq: condition for updating row scaling vector} explicitly as
\begin{equation}\label{eq: empirical candidate selection}
b:=
\begin{cases}
\mathrm{row},
&\displaystyle
\frac1m\sum_{i=1}^m\log\frac{\alpha}{1-\eta\hat g_i}
>
\frac1n\sum_{j=1}^n\log\frac{\alpha m/n}{1-\eta\hat h_j},
\\[6pt]
\mathrm{col},&\text{otherwise}.
\end{cases}
\end{equation}

\begin{lem}\label{lem: candidate decrease comparison}
With $e$ and $b$ defined in
\eqref{eq: candidate logarithmic denominator error} and
\eqref{eq: empirical candidate selection}, respectively,
\begin{equation}\label{eq: perturbed greedy comparison}
\Delta_b
\ge\frac14\max\{\Delta_{\mathrm{row}},\Delta_{\mathrm{col}}\}
-\frac34ne^2.
\end{equation}
\end{lem}

\begin{proof}
By Lemma~\ref{lem: empirical update stability}, the positive scaling
vectors and the absence of zero rows or columns give
\begin{equation}\label{eq: empirical candidate denominator positivity}
1-\eta\hat g_i>0,\qquad 1-\eta\hat h_j>0.
\end{equation}
By the definitions of the empirical resolvent diagonals in
\eqref{eq: right resolvent formula} and
\eqref{eq: left resolvent formula},
\begin{equation}\label{eq: empirical candidate resolvent positivity}
\begin{aligned}
\hat g_i
&=\eta\left[
(\widetilde Y\widetilde Y^T+\eta^2I_m)^{-1}\right]_{ii}>0,\\
\hat h_j
&=\eta\left[
(\widetilde Y^T\widetilde Y+\eta^2I_n)^{-1}\right]_{jj}>0.
\end{aligned}
\end{equation}
The inequalities in~\eqref{eq: empirical candidate resolvent positivity}
follow because both matrices inside the inverses are positive
definite and $\eta>0$.
Combining~\eqref{eq: empirical candidate denominator positivity}
and~\eqref{eq: empirical candidate resolvent positivity} with the
assumed ranges of $\mathbf g,\mathbf h$ gives
\begin{equation}\label{eq: candidate denominator ranges}
\begin{aligned}
0<1-\eta\hat g_i<1,\qquad &0<1-\eta g_i<1,\\
0<1-\eta\hat h_j<1,\qquad &0<1-\eta h_j<1.
\end{aligned}
\end{equation}
The assumption $m\le n$ and the coefficient $\alpha$ defined in
\eqref{eq: alpha def} also give
\begin{equation}\label{eq: candidate coefficient bounds}
1\le m\le n,\qquad 0<\alpha m/n\le\alpha<1.
\end{equation}

The ranges~\eqref{eq: candidate denominator ranges} and the coefficient
bounds~\eqref{eq: candidate coefficient bounds} allow us to apply
\eqref{eq: quadratic decrease comparison} with
$u=1-\eta g_i$, $\widetilde u=1-\eta\hat g_i$, and $v=\alpha$.
Together with~\eqref{eq: candidate logarithmic denominator error},
this gives, for each $i\in[m]$,
\begin{equation}\label{eq: row scalar candidate comparison}
\begin{aligned}
\alpha\varphi\!\left(\frac{1-\eta g_i}{\alpha}\right)
&\le
2\alpha\varphi\!\left(\frac{1-\eta\hat g_i}{\alpha}\right)
+\left[\log\frac{1-\eta g_i}{1-\eta\hat g_i}\right]^2\\
&\le
2\alpha\varphi\!\left(\frac{1-\eta\hat g_i}{\alpha}\right)+e^2.
\end{aligned}
\end{equation}
Summing~\eqref{eq: row scalar candidate comparison} and then
exchanging $g_i$ and $\hat g_i$ in that inequality proves, by
\eqref{eq: row candidate definitions},
\begin{equation}\label{eq: row candidate comparison}
\Delta_{\mathrm{row}}\le2\hat\Delta_{\mathrm{row}}+me^2,
\qquad
\hat\Delta_{\mathrm{row}}\le2\Delta_{\mathrm{row}}+me^2.
\end{equation}
Applying~\eqref{eq: quadratic decrease comparison} with
$u=1-\eta h_j$, $\widetilde u=1-\eta\hat h_j$, and $v=\alpha m/n$,
and using~\eqref{eq: candidate logarithmic denominator error}, gives
\begin{equation}\label{eq: column scalar candidate comparison}
\begin{aligned}
\frac{\alpha m}{n}
\varphi\!\left(\frac{1-\eta h_j}{\alpha m/n}\right)
&\le
2\frac{\alpha m}{n}
\varphi\!\left(\frac{1-\eta\hat h_j}{\alpha m/n}\right)
+\left[\log\frac{1-\eta h_j}{1-\eta\hat h_j}\right]^2\\
&\le
2\frac{\alpha m}{n}
\varphi\!\left(\frac{1-\eta\hat h_j}{\alpha m/n}\right)+e^2.
\end{aligned}
\end{equation}
Summing~\eqref{eq: column scalar candidate comparison} and then
exchanging $h_j$ and $\hat h_j$ proves, by
\eqref{eq: column candidate definitions},
\begin{equation}\label{eq: column candidate comparison}
\Delta_{\mathrm{col}}\le2\hat\Delta_{\mathrm{col}}+ne^2,
\qquad
\hat\Delta_{\mathrm{col}}\le2\Delta_{\mathrm{col}}+ne^2.
\end{equation}
Since $m\le n$ by~\eqref{eq: candidate coefficient bounds},
\eqref{eq: row candidate comparison} and
\eqref{eq: column candidate comparison} imply
\begin{equation}\label{eq: two sided candidate comparison}
\Delta_{b'}\le2\hat\Delta_{b'}+ne^2,
\qquad
\hat\Delta_{b'}\le2\Delta_{b'}+ne^2,
\qquad b'\in\{\mathrm{row},\mathrm{col}\}.
\end{equation}

We next prove that $b$ in~\eqref{eq: empirical candidate selection}
satisfies $\hat\Delta_b=\max\{\hat\Delta_{\mathrm{row}},
\hat\Delta_{\mathrm{col}}\}$.
The diagonal formula for $1-\eta\hat g_i$ derived in
Lemma~\ref{lem: empirical update stability}
(see~\eqref{eq: empirical row update inverse formula}) and the
definition of $\hat h_j$ in~\eqref{eq: left resolvent formula} give
\begin{equation}\label{eq: empirical denominator sum identity}
\begin{aligned}
\sum_{i=1}^m(1-\eta\hat g_i)
&=\operatorname{tr}\!\left[
\widetilde Y(\widetilde Y^T\widetilde Y+\eta^2I_n)^{-1}
\widetilde Y^T\right]\\
&=\operatorname{tr}\!\left[
\widetilde Y^T\widetilde Y
(\widetilde Y^T\widetilde Y+\eta^2I_n)^{-1}\right]\\
&=\operatorname{tr}\!\left[
I_n-\eta^2(\widetilde Y^T\widetilde Y+\eta^2I_n)^{-1}\right]\\
&=\sum_{j=1}^n(1-\eta\hat h_j).
\end{aligned}
\end{equation}
The second equality in~\eqref{eq: empirical denominator sum identity}
uses cyclicity of the trace; the third follows by writing
$\widetilde Y^T\widetilde Y$ as
$(\widetilde Y^T\widetilde Y+\eta^2I_n)-\eta^2I_n$.

Expanding $\varphi$ in~\eqref{eq: row candidate definitions} and
\eqref{eq: column candidate definitions} according to
\eqref{eq: scalar decrease definition} gives
\begin{align}
\hat\Delta_{\mathrm{row}}
&=\sum_{i=1}^m(1-\eta\hat g_i)-m\alpha
+\alpha\sum_{i=1}^m\log\frac{\alpha}{1-\eta\hat g_i},
\label{eq: empirical row candidate expansion}\\
\hat\Delta_{\mathrm{col}}
&=\sum_{j=1}^n(1-\eta\hat h_j)-n\frac{\alpha m}{n}
+\frac{\alpha m}{n}
\sum_{j=1}^n\log\frac{\alpha m/n}{1-\eta\hat h_j}.
\label{eq: empirical column candidate expansion}
\end{align}
Subtracting~\eqref{eq: empirical column candidate expansion}
from~\eqref{eq: empirical row candidate expansion} and using
\eqref{eq: empirical denominator sum identity} yields
\begin{equation}\label{eq: empirical candidate difference}
\begin{aligned}
\hat\Delta_{\mathrm{row}}-\hat\Delta_{\mathrm{col}}
&=\underbrace{
\sum_{i=1}^m(1-\eta\hat g_i)
-\sum_{j=1}^n(1-\eta\hat h_j)}_{=\,0}
+\underbrace{n\frac{\alpha m}{n}-m\alpha}_{=\,0}\\
&\quad+\alpha\sum_{i=1}^m\log\frac{\alpha}{1-\eta\hat g_i}
-\frac{\alpha m}{n}
\sum_{j=1}^n\log\frac{\alpha m/n}{1-\eta\hat h_j}\\
&=m\alpha\left[
\frac1m\sum_{i=1}^m\log\frac{\alpha}{1-\eta\hat g_i}
-\frac1n\sum_{j=1}^n\log\frac{\alpha m/n}{1-\eta\hat h_j}
\right].
\end{aligned}
\end{equation}
Since $m\alpha>0$ by~\eqref{eq: candidate coefficient bounds},
\eqref{eq: empirical candidate difference} and
\eqref{eq: empirical candidate selection} imply
\begin{equation}\label{eq: selected empirical candidate maximum}
\hat\Delta_b=
\begin{cases}
\hat\Delta_{\mathrm{row}},
&\hat\Delta_{\mathrm{row}}>\hat\Delta_{\mathrm{col}},\\
\hat\Delta_{\mathrm{col}},
&\hat\Delta_{\mathrm{row}}\le\hat\Delta_{\mathrm{col}}
\end{cases}
=\max\{\hat\Delta_{\mathrm{row}},\hat\Delta_{\mathrm{col}}\}.
\end{equation}
Finally, applying the first inequality in
\eqref{eq: two sided candidate comparison} to both
$\Delta_{\mathrm{row}}$ and $\Delta_{\mathrm{col}}$, then using
\eqref{eq: selected empirical candidate maximum}, and then applying
the second inequality in~\eqref{eq: two sided candidate comparison}
to $\hat\Delta_b$, gives
\begin{equation}\label{eq: candidate comparison conclusion}
\begin{aligned}
\max\{\Delta_{\mathrm{row}},\Delta_{\mathrm{col}}\}
&\le2\max\{\hat\Delta_{\mathrm{row}},\hat\Delta_{\mathrm{col}}\}+ne^2\\
&=2\hat\Delta_b+ne^2\\
&\le2(2\Delta_b+ne^2)+ne^2
=4\Delta_b+3ne^2.
\end{aligned}
\end{equation}
Rearranging~\eqref{eq: candidate comparison conclusion} proves
\eqref{eq: perturbed greedy comparison}.
\end{proof}

\subsubsection{Uniform comparison along the empirical choices}

Fix $A>0$, $\varepsilon\in(0,\delta_0)$, and $\tau>0$ as in
Theorem~\ref{thm: main theoretical guarantees}.
Let $T_n$ and $\omega_n$ be defined
by~\eqref{eq: statistical error and iteration horizon}.
The vectors $(\hat{\mathbf x}^{(t)},\hat{\mathbf y}^{(t)})$
are the scaling factors produced by the empirical procedure.
At iteration $t+1$, the vectors
$(\hat{\mathbf g}^{(t+1)},\hat{\mathbf h}^{(t+1)})$
are computed from $Y$ at
$(\hat{\mathbf x}^{(t)},\hat{\mathbf y}^{(t)})$
by the data resolvent formulas
\eqref{eq: g_hat formula} and~\eqref{eq: h_hat formula}.
The row or column choice
in~\eqref{eq: empirical candidate selection} is therefore
\begin{equation}\label{eq: comparison empirical branch selection}
b_{t+1}:=
\begin{cases}
\mathrm{row},
&\displaystyle
\frac1m\sum_{i=1}^m
\log\frac{\alpha}{1-\eta\hat g_i^{(t+1)}}
>
\frac1n\sum_{j=1}^n
\log\frac{\alpha m/n}{1-\eta\hat h_j^{(t+1)}},
\\[8pt]
\mathrm{col},
&\displaystyle
\frac1m\sum_{i=1}^m
\log\frac{\alpha}{1-\eta\hat g_i^{(t+1)}}
\le
\frac1n\sum_{j=1}^n
\log\frac{\alpha m/n}{1-\eta\hat h_j^{(t+1)}}.
\end{cases}
\end{equation}

For the proof, we introduce scaling vectors
$(\bar{\mathbf x}^{(t)},\bar{\mathbf y}^{(t)})$.
Their updates use the choices $b_{t+1}$
in~\eqref{eq: comparison empirical branch selection}
and the exact population Dyson solution at
$(\bar{\mathbf x}^{(t)},\bar{\mathbf y}^{(t)})$.
Set
\begin{equation}\label{eq: comparison population initialization}
(\bar{\mathbf x}^{(0)},\bar{\mathbf y}^{(0)})
=(\mathbf1_m,\mathbf1_n).
\end{equation}
For $0\le t<T_n$, let
$(\bar{\mathbf g}^{(t+1)},\bar{\mathbf h}^{(t+1)})$
be the positive solution of the population Dyson
equations~\eqref{eq: Dyson equation} at
$(\mathbf x,\mathbf y)
=(\bar{\mathbf x}^{(t)},\bar{\mathbf y}^{(t)})$.
The row and column scaling updates
\eqref{eq: x_t update}--\eqref{eq: y_t update} then give
\begin{equation}\label{eq: comparison population recursion}
(\bar{\mathbf x}^{(t+1)},\bar{\mathbf y}^{(t+1)})
=
\begin{cases}
\displaystyle
\left(
\frac{\alpha\bar{\mathbf x}^{(t)}}
     {\mathbf1_m-\eta\bar{\mathbf g}^{(t+1)}},
\bar{\mathbf y}^{(t)}
\right),
& b_{t+1}=\mathrm{row},
\\[8pt]
\displaystyle
\left(
\bar{\mathbf x}^{(t)},
\frac{(\alpha m/n)\bar{\mathbf y}^{(t)}}
     {\mathbf1_n-\eta\bar{\mathbf h}^{(t+1)}}
\right),
& b_{t+1}=\mathrm{col}.
\end{cases}
\end{equation}
All vector quotients in~\eqref{eq: comparison population recursion}
are entrywise.
For fixed $S$, the vectors
$(\bar{\mathbf x}^{(t)},\bar{\mathbf y}^{(t)})$
depend on $Y$ only through $b_1,\ldots,b_t$.

Using $\varphi$ from~\eqref{eq: scalar decrease definition},
define
\begin{equation}\label{eq: comparison population candidate decreases}
\begin{aligned}
\bar\Delta_{\mathrm{row}}^{(t+1)}
&:=\alpha\sum_{i=1}^m
\varphi\!\left(
\frac{1-\eta\bar g_i^{(t+1)}}{\alpha}
\right),\\
\bar\Delta_{\mathrm{col}}^{(t+1)}
&:=\frac{\alpha m}{n}\sum_{j=1}^n
\varphi\!\left(
\frac{1-\eta\bar h_j^{(t+1)}}{\alpha m/n}
\right),
\end{aligned}
\end{equation}
By the candidate decrease formulas
\eqref{eq:thm2_candidate_decreases},
$\bar\Delta_{\mathrm{row}}^{(t+1)}$ is the decrease of $J$
under the row update~\eqref{eq: x_t update}, and
$\bar\Delta_{\mathrm{col}}^{(t+1)}$ is the decrease under
the column update~\eqref{eq: y_t update}, both starting from
$(\bar{\mathbf x}^{(t)},\bar{\mathbf y}^{(t)})$ and holding
$(\bar{\mathbf g}^{(t+1)},\bar{\mathbf h}^{(t+1)})$ fixed,
as specified in~\eqref{eq:thm2_decrease_definitions}.
Define
\begin{equation}\label{eq: comparison scaling discrepancy}
e_t:=
\max\left\{
\left\|\log\frac{\hat{\mathbf x}^{(t)}}{\bar{\mathbf x}^{(t)}}\right\|_\infty,
\left\|\log\frac{\hat{\mathbf y}^{(t)}}{\bar{\mathbf y}^{(t)}}\right\|_\infty
\right\}.
\end{equation}

\begin{lem}\label{lem: uniform empirical population comparison}
Under the assumptions of
Theorem~\ref{thm: main theoretical guarantees},
for all sufficiently large dimensions there is an event
$\mathcal E_n$ with $\mathbb P(\mathcal E_n^c)\le Kn^{-\tau}$
on which all empirical updates through iteration $T_n$ are
well-defined and, simultaneously,
\begin{equation}\label{eq: comparison tracking bound}
e_t\le Kt\omega_n,
\qquad 0\le t\le T_n,
\end{equation}
and
\begin{equation}\label{eq: approximate greedy population decrease}
\bar\Delta_{b_{t+1}}^{(t+1)}
\ge
\frac14\max\left\{
\bar\Delta_{\mathrm{row}}^{(t+1)},
\bar\Delta_{\mathrm{col}}^{(t+1)}
\right\}
-Km(t+1)^2\omega_n^2,
\qquad 0\le t<T_n.
\end{equation}
The constant $K$ and the required lower bound on the dimensions
depend only on $\eta,c,C,\gamma_-,\gamma_+,A,\varepsilon,\tau$
and the global constants in
Assumption~\ref{assump: main assumption}.
\end{lem}

By~\eqref{eq: comparison scaling discrepancy},
the estimate~\eqref{eq: comparison tracking bound} bounds both
$\|\log(\hat{\mathbf x}^{(t)}/\bar{\mathbf x}^{(t)})\|_\infty$
and $\|\log(\hat{\mathbf y}^{(t)}/\bar{\mathbf y}^{(t)})\|_\infty$
by $Kt\omega_n$.
The main proof uses~\eqref{eq: approximate greedy population decrease}
to derive the potential contraction
inequality~\eqref{eq: approximate potential contraction}.

\begin{proof}
Constants denoted by $K$ depend only on the parameters listed in the lemma and may differ across the proof.
In particular, they do not depend on $m,n,t$.
We take the dimensions large enough that $T_n\ge1$.
Let $w=(w_1,\ldots,w_s)\in\{\mathrm{row},\mathrm{col}\}^s$
be a fixed sequence, and write $|w|=s$.
Denote the empty sequence by $\varnothing$.
The notation $(w,\mathrm{row})$ means
$(w_1,\ldots,w_s,\mathrm{row})$, and $(w,\mathrm{col})$ means
$(w_1,\ldots,w_s,\mathrm{col})$.
Define the population scaling vectors recursively by
\begin{equation}\label{eq: comparison deterministic histories}
\begin{aligned}
(\mathbf x_\varnothing,\mathbf y_\varnothing)
&=(\mathbf1_m,\mathbf1_n),\\
(\mathbf x_{(w,\mathrm{row})},\mathbf y_{(w,\mathrm{row})})
&=
\left(
\frac{\alpha\mathbf x_w}{\mathbf1_m-\eta\mathbf g_w},
\mathbf y_w
\right),\\
(\mathbf x_{(w,\mathrm{col})},\mathbf y_{(w,\mathrm{col})})
&=
\left(
\mathbf x_w,
\frac{(\alpha m/n)\mathbf y_w}{\mathbf1_n-\eta\mathbf h_w}
\right),
\end{aligned}
\end{equation}
where $(\mathbf g_w,\mathbf h_w)$ is the positive solution
of the population Dyson equations~\eqref{eq: Dyson equation}
at $(\mathbf x_w,\mathbf y_w)$.
Define $\widetilde{\mathbf g}_w$ and
$\widetilde{\mathbf h}_w$ by the data resolvent formulas
\eqref{eq: right resolvent formula} and
\eqref{eq: left resolvent formula}, respectively, with
$(\mathbf x,\mathbf y)=(\mathbf x_w,\mathbf y_w)$
in the data scaling formula~\eqref{eq: Y_tilde def}.
Thus $\widetilde{\mathbf g}_w$ and
$\widetilde{\mathbf h}_w$ depend on $Y$ even when $w$ is fixed.

For each fixed $w$, the vectors
$\mathbf x_w,\mathbf y_w,\mathbf g_w,\mathbf h_w$
in~\eqref{eq: comparison deterministic histories}
depend only on $S$ and $w$.
The uniform coordinate bounds
\eqref{eq: uniform population bounds} of
Lemma~\ref{lem: finite-time bounds population}
hold for every sequence of row and column choices.
They give
\begin{equation}\label{eq: comparison deterministic population bounds}
\begin{aligned}
c_0\le x_{w,i}\le C_0,\qquad
b_0\le g_{w,i}\le\frac{1-r_0}{\eta}
&\qquad(i\in[m]),\\
c_0\le y_{w,j}\le C_0,\qquad
b_0\le h_{w,j}\le\frac{1-r_0}{\eta}
&\qquad(j\in[n]),
\end{aligned}
\qquad |w|<T_n.
\end{equation}
The constants $c_0,C_0,b_0,r_0$
in~\eqref{eq: comparison deterministic population bounds}
are independent of $w$.
There are $2^s$ sequences of length $s$. Hence the definition
of $T_n$ in~\eqref{eq: statistical error and iteration horizon}
implies
\begin{equation}\label{eq: comparison number of histories}
\#\{w:|w|<T_n\}
=\sum_{s=0}^{T_n-1}2^s
=2^{T_n}-1
\le n^{A\log2}.
\end{equation}
Apply Lemma~\ref{lem:resolvent_supnorm_concentration}
at every $(\mathbf x_w,\mathbf y_w)$ with $|w|<T_n$,
with its parameter $\tau$ replaced by $\tau+A\log2+1$.
The bounds~\eqref{eq: comparison deterministic population bounds}
allow us to use $c_x=c_y=c_0$ and $C_x=C_y=C_0$
in that lemma.
We can therefore choose one concentration constant $K$
for all $|w|<T_n$.
Define
\begin{equation}\label{eq: all history concentration}
\mathcal E_n:=
\left\{
\max_{|w|<T_n}\max\left\{
\|\widetilde{\mathbf g}_w-\mathbf g_w\|_\infty,
\|\widetilde{\mathbf h}_w-\mathbf h_w\|_\infty
\right\}
\le K\omega_n
\right\}.
\end{equation}
The probability estimate
\eqref{eq: comparison concentration failure probability}
follows from the union bound, the probability estimate
in Lemma~\ref{lem:resolvent_supnorm_concentration},
and the count~\eqref{eq: comparison number of histories}:
\begin{equation}\label{eq: comparison concentration failure probability}
\begin{aligned}
\mathbb P(\mathcal E_n^c)
&\le(2^{T_n}-1)Kn^{-(\tau+A\log2+1)}
\le Kn^{-(\tau+1)}
\le Kn^{-\tau}.
\end{aligned}
\end{equation}

By the localization bound
$\kappa_n\le\widetilde C_1n^{-\delta_1}$ in
Assumption~\ref{assump: main assumption}.\ref{assump: singular vector localization}
and the definition of $\omega_n$
in~\eqref{eq: statistical error and iteration horizon},
\begin{equation}\label{eq: comparison vanishing statistical error}
0\le\omega_n
\le n^{\varepsilon-1/2}+\widetilde C_1n^{-\delta_1}
\longrightarrow0,
\end{equation}
since $\varepsilon<\delta_0\le1/2$ and $\delta_1>0$.
We henceforth work on $\mathcal E_n$.
By~\eqref{eq: comparison deterministic population bounds},
\eqref{eq: all history concentration},
and~\eqref{eq: comparison vanishing statistical error},
for sufficiently large dimensions,
\begin{equation}\label{eq: comparison history denominator positivity}
\begin{aligned}
1-\eta\widetilde g_{w,i}
&\ge1-\eta g_{w,i}
-\eta|\widetilde g_{w,i}-g_{w,i}|
\ge r_0-K\eta\omega_n\ge\frac{r_0}{2},\\
1-\eta\widetilde h_{w,j}
&\ge1-\eta h_{w,j}
-\eta|\widetilde h_{w,j}-h_{w,j}|
\ge r_0-K\eta\omega_n\ge\frac{r_0}{2},
\end{aligned}
\end{equation}
for every $|w|<T_n$, $i\in[m]$, and $j\in[n]$.
Equation~\eqref{eq: comparison deterministic population bounds}
also gives $1-\eta g_{w,i}\ge r_0$
and $1-\eta h_{w,j}\ge r_0$.
Consequently, all integration endpoints
in~\eqref{eq: all history logarithmic denominator error}
are at least $r_0/2$ by these two inequalities
and~\eqref{eq: comparison history denominator positivity}.
Using $1/s\le2/r_0$ on each integration interval and
the error bound in~\eqref{eq: all history concentration} gives
\begin{equation}\label{eq: all history logarithmic denominator error}
\begin{aligned}
\left|\log\frac{1-\eta\widetilde g_{w,i}}{1-\eta g_{w,i}}\right|
&=
\left|\int_{1-\eta g_{w,i}}^{1-\eta\widetilde g_{w,i}}
\frac{ds}{s}\right|
\le\frac{2\eta}{r_0}|\widetilde g_{w,i}-g_{w,i}|
\le K\omega_n,\\
\left|\log\frac{1-\eta\widetilde h_{w,j}}{1-\eta h_{w,j}}\right|
&=
\left|\int_{1-\eta h_{w,j}}^{1-\eta\widetilde h_{w,j}}
\frac{ds}{s}\right|
\le\frac{2\eta}{r_0}|\widetilde h_{w,j}-h_{w,j}|
\le K\omega_n.
\end{aligned}
\end{equation}
Equation~\eqref{eq: all history logarithmic denominator error}
holds for every $|w|<T_n$, $i\in[m]$, and $j\in[n]$.

At the initial pair
$(\mathbf x_\varnothing,\mathbf y_\varnothing)
=(\mathbf1_m,\mathbf1_n)$
from~\eqref{eq: comparison deterministic histories},
the resolvent formulas
\eqref{eq: right resolvent formula} and
\eqref{eq: left resolvent formula} imply
\begin{equation}\label{eq: comparison exclusion of zero rows and columns}
\begin{aligned}
Y_{i,:}=\mathbf0
&\ \Longrightarrow\
\widetilde g_{\varnothing,i}=1/\eta
\ \Longrightarrow\
1-\eta\widetilde g_{\varnothing,i}=0,\\
Y_{:,j}=\mathbf0
&\ \Longrightarrow\
\widetilde h_{\varnothing,j}=1/\eta
\ \Longrightarrow\
1-\eta\widetilde h_{\varnothing,j}=0.
\end{aligned}
\end{equation}
Since $T_n\ge1$, we can take $w=\varnothing$
in~\eqref{eq: comparison history denominator positivity}.
It gives $1-\eta\widetilde g_{\varnothing,i}\ge r_0/2>0$
and $1-\eta\widetilde h_{\varnothing,j}\ge r_0/2>0$.
Thus~\eqref{eq: comparison exclusion of zero rows and columns}
excludes $Y_{i,:}=\mathbf0$ for every $i\in[m]$
and $Y_{:,j}=\mathbf0$ for every $j\in[n]$.
The empirical initial vectors satisfy
$\hat{\mathbf x}^{(0)}=\mathbf1_m$ and
$\hat{\mathbf y}^{(0)}=\mathbf1_n$.
Lemma~\ref{lem: empirical update stability} shows that, when $Y$
has no zero rows or columns, either empirical update maps
positive scaling vectors to positive scaling vectors.
Induction on the number of updates therefore shows that
$(\hat{\mathbf x}^{(t)},\hat{\mathbf y}^{(t)})$ is positive and
well-defined for every $0\le t\le T_n$ on $\mathcal E_n$.

For $0\le t<T_n$, set $w=(b_1,\ldots,b_t)$
using the choices made by the empirical procedure.
At $t=0$, equations
\eqref{eq: comparison population initialization}
and~\eqref{eq: comparison deterministic histories}
give $(\bar{\mathbf x}^{(0)},\bar{\mathbf y}^{(0)})
=(\mathbf x_\varnothing,\mathbf y_\varnothing)$.
If $(\bar{\mathbf x}^{(t)},\bar{\mathbf y}^{(t)})
=(\mathbf x_w,\mathbf y_w)$, uniqueness of the population
Dyson solution gives
$(\bar{\mathbf g}^{(t+1)},\bar{\mathbf h}^{(t+1)})
=(\mathbf g_w,\mathbf h_w)$.
Equations~\eqref{eq: comparison population recursion}
and~\eqref{eq: comparison deterministic histories}
then give
$(\bar{\mathbf x}^{(t+1)},\bar{\mathbf y}^{(t+1)})
=(\mathbf x_{(w,b_{t+1})},\mathbf y_{(w,b_{t+1})})$.
Induction gives
\begin{equation}\label{eq: comparison selected history}
(\bar{\mathbf x}^{(t)},\bar{\mathbf y}^{(t)})
=(\mathbf x_w,\mathbf y_w),
\qquad
(\bar{\mathbf g}^{(t+1)},\bar{\mathbf h}^{(t+1)})
=(\mathbf g_w,\mathbf h_w).
\end{equation}
The event $\mathcal E_n$ in~\eqref{eq: all history concentration}
requires the concentration bound at every fixed sequence
of length less than $T_n$.
Hence that bound and
\eqref{eq: all history logarithmic denominator error}
hold also at $w=(b_1,\ldots,b_t)$, regardless of its dependence
on $Y$.

The data resolvent estimates at
$(\bar{\mathbf x}^{(t)},\bar{\mathbf y}^{(t)})$ are
$(\widetilde{\mathbf g}_w,\widetilde{\mathbf h}_w)$
by~\eqref{eq: comparison selected history}.
Substituting $(\widetilde{\mathbf g}_w,\widetilde{\mathbf h}_w)$
into the row and column scaling formulas
\eqref{eq: x_t update}--\eqref{eq: y_t update}, define
\begin{equation}\label{eq: comparison auxiliary empirical update}
(\mathbf x',\mathbf y')
=
\begin{cases}
\displaystyle
\left(
\frac{\alpha\bar{\mathbf x}^{(t)}}
     {\mathbf1_m-\eta\widetilde{\mathbf g}_w},
\bar{\mathbf y}^{(t)}
\right),
&b_{t+1}=\mathrm{row},
\\[8pt]
\displaystyle
\left(
\bar{\mathbf x}^{(t)},
\frac{(\alpha m/n)\bar{\mathbf y}^{(t)}}
     {\mathbf1_n-\eta\widetilde{\mathbf h}_w}
\right),
&b_{t+1}=\mathrm{col}.
\end{cases}
\end{equation}
The empirical update specified by $b_{t+1}$
in~\eqref{eq: comparison empirical branch selection}
maps $(\bar{\mathbf x}^{(t)},\bar{\mathbf y}^{(t)})$
to $(\mathbf x',\mathbf y')$
by~\eqref{eq: comparison auxiliary empirical update}.
The actual empirical procedure uses the same $b_{t+1}$ to update
$(\hat{\mathbf x}^{(t)},\hat{\mathbf y}^{(t)})$
and obtains $(\hat{\mathbf x}^{(t+1)},\hat{\mathbf y}^{(t+1)})$.
Apply the nonexpansiveness inequality
\eqref{eq: nonexpansive empirical updates} with input pairs
$(\mathbf x,\mathbf y)
=(\hat{\mathbf x}^{(t)},\hat{\mathbf y}^{(t)})$
and $(\mathbf u,\mathbf v)
=(\bar{\mathbf x}^{(t)},\bar{\mathbf y}^{(t)})$.
Their input discrepancy is $e_t$
by~\eqref{eq: comparison scaling discrepancy}, so
\begin{equation}\label{eq: comparison same choice nonexpansiveness}
\max\left\{
\left\|\log\frac{\hat{\mathbf x}^{(t+1)}}{\mathbf x'}\right\|_\infty,
\left\|\log\frac{\hat{\mathbf y}^{(t+1)}}{\mathbf y'}\right\|_\infty
\right\}
\le e_t.
\end{equation}

To compare $(\mathbf x',\mathbf y')$ with
$(\bar{\mathbf x}^{(t+1)},\bar{\mathbf y}^{(t+1)})$,
divide $x_i'$ by $\bar x_i^{(t+1)}$ and $y_j'$ by
$\bar y_j^{(t+1)}$ using
\eqref{eq: comparison auxiliary empirical update}
and~\eqref{eq: comparison population recursion}.
This gives
\begin{equation}\label{eq: comparison auxiliary update ratios}
\begin{array}{lll}
\displaystyle
\frac{x_i'}{\bar x_i^{(t+1)}}
=\frac{1-\eta\bar g_i^{(t+1)}}{1-\eta\widetilde g_{w,i}},
&
\displaystyle
\frac{y_j'}{\bar y_j^{(t+1)}}=1,
& b_{t+1}=\mathrm{row},
\\[8pt]
\displaystyle
\frac{x_i'}{\bar x_i^{(t+1)}}=1,
&
\displaystyle
\frac{y_j'}{\bar y_j^{(t+1)}}
=\frac{1-\eta\bar h_j^{(t+1)}}{1-\eta\widetilde h_{w,j}},
& b_{t+1}=\mathrm{col}.
\end{array}
\end{equation}
Take logarithms in~\eqref{eq: comparison auxiliary update ratios}.
For $b_{t+1}=\mathrm{row}$, use
$\bar g_i^{(t+1)}=g_{w,i}$
from~\eqref{eq: comparison selected history}
and the bound involving $\widetilde g_{w,i}$ and $g_{w,i}$
in~\eqref{eq: all history logarithmic denominator error}.
For $b_{t+1}=\mathrm{col}$, use
$\bar h_j^{(t+1)}=h_{w,j}$ and the bound involving
$\widetilde h_{w,j}$ and $h_{w,j}$
in~\eqref{eq: all history logarithmic denominator error}.
The logarithms of the ratios equal to $1$
in~\eqref{eq: comparison auxiliary update ratios} are zero.
Thus
\begin{equation}\label{eq: comparison one step estimation error}
\max\left\{
\left\|\log\frac{\mathbf x'}{\bar{\mathbf x}^{(t+1)}}\right\|_\infty,
\left\|\log\frac{\mathbf y'}{\bar{\mathbf y}^{(t+1)}}\right\|_\infty
\right\}
\le K\omega_n.
\end{equation}
For the positive vectors $(\mathbf x',\mathbf y')$
in~\eqref{eq: comparison auxiliary empirical update},
the following identities hold entrywise:
\begin{equation}\label{eq: comparison tracking decomposition}
\begin{aligned}
\log\frac{\hat{\mathbf x}^{(t+1)}}{\bar{\mathbf x}^{(t+1)}}
&=
\log\frac{\hat{\mathbf x}^{(t+1)}}{\mathbf x'}
+\log\frac{\mathbf x'}{\bar{\mathbf x}^{(t+1)}},\\
\log\frac{\hat{\mathbf y}^{(t+1)}}{\bar{\mathbf y}^{(t+1)}}
&=
\log\frac{\hat{\mathbf y}^{(t+1)}}{\mathbf y'}
+\log\frac{\mathbf y'}{\bar{\mathbf y}^{(t+1)}}.
\end{aligned}
\end{equation}
Apply the triangle inequality for $\|\cdot\|_\infty$
to each line of~\eqref{eq: comparison tracking decomposition}.
Then use
\eqref{eq: comparison same choice nonexpansiveness}
and~\eqref{eq: comparison one step estimation error},
together with the definition of $e_{t+1}$
in~\eqref{eq: comparison scaling discrepancy}, to obtain
\begin{equation}\label{eq: comparison tracking recursion}
e_{t+1}\le e_t+K\omega_n,\qquad e_0=0.
\end{equation}
Here $e_0=0$ follows from the empirical initialization
and~\eqref{eq: comparison population initialization}.
Summing $e_{\ell+1}-e_\ell\le K\omega_n$ from
\eqref{eq: comparison tracking recursion}
over $\ell=0,\ldots,t-1$ gives
\begin{equation}\label{eq: comparison accumulated tracking error}
e_t\le e_0+\sum_{\ell=0}^{t-1}K\omega_n
=Kt\omega_n,
\qquad 1\le t\le T_n.
\end{equation}
Equation~\eqref{eq: comparison accumulated tracking error}
proves~\eqref{eq: comparison tracking bound} for $1\le t\le T_n$;
the case $t=0$ follows from $e_0=0$
in~\eqref{eq: comparison tracking recursion}.

Apply the resolvent denominator inequality
\eqref{eq: empirical denominator stability} at
$(\mathbf x,\mathbf y)
=(\hat{\mathbf x}^{(t)},\hat{\mathbf y}^{(t)})$
and $(\mathbf u,\mathbf v)
=(\bar{\mathbf x}^{(t)},\bar{\mathbf y}^{(t)})$.
The data resolvent vectors at the first and second pairs are
$(\hat{\mathbf g}^{(t+1)},\hat{\mathbf h}^{(t+1)})$
and $(\widetilde{\mathbf g}_w,\widetilde{\mathbf h}_w)$,
respectively, and the input discrepancy is $e_t$
by~\eqref{eq: comparison scaling discrepancy}.
Therefore
\begin{equation}\label{eq: comparison empirical denominator perturbation}
\max\left\{
\left\|\log\frac{\mathbf1_m-\eta\hat{\mathbf g}^{(t+1)}}
                 {\mathbf1_m-\eta\widetilde{\mathbf g}_w}\right\|_\infty,
\left\|\log\frac{\mathbf1_n-\eta\hat{\mathbf h}^{(t+1)}}
                 {\mathbf1_n-\eta\widetilde{\mathbf h}_w}\right\|_\infty
\right\}\le2e_t.
\end{equation}
Substitute $\bar{\mathbf g}^{(t+1)}=\mathbf g_w$ and
$\bar{\mathbf h}^{(t+1)}=\mathbf h_w$
from~\eqref{eq: comparison selected history}.
The logarithm of a product is the sum of the logarithms,
so the triangle inequality for $\|\cdot\|_\infty$,
the bound~\eqref{eq: comparison empirical denominator perturbation},
and~\eqref{eq: all history logarithmic denominator error} give
\begin{equation}\label{eq: comparison denominator decomposition}
\begin{aligned}
\left\|\log\frac{\mathbf1_m-\eta\hat{\mathbf g}^{(t+1)}}
              {\mathbf1_m-\eta\bar{\mathbf g}^{(t+1)}}\right\|_\infty
&=
\left\|
\log\frac{\mathbf1_m-\eta\hat{\mathbf g}^{(t+1)}}
              {\mathbf1_m-\eta\widetilde{\mathbf g}_w}
+
\log\frac{\mathbf1_m-\eta\widetilde{\mathbf g}_w}
              {\mathbf1_m-\eta\mathbf g_w}
\right\|_\infty\\
&\le
\left\|\log\frac{\mathbf1_m-\eta\hat{\mathbf g}^{(t+1)}}
              {\mathbf1_m-\eta\widetilde{\mathbf g}_w}\right\|_\infty
+
\left\|\log\frac{\mathbf1_m-\eta\widetilde{\mathbf g}_w}
              {\mathbf1_m-\eta\mathbf g_w}\right\|_\infty\\
&\le2e_t+K\omega_n,\\[4pt]
\left\|\log\frac{\mathbf1_n-\eta\hat{\mathbf h}^{(t+1)}}
              {\mathbf1_n-\eta\bar{\mathbf h}^{(t+1)}}\right\|_\infty
&=
\left\|
\log\frac{\mathbf1_n-\eta\hat{\mathbf h}^{(t+1)}}
              {\mathbf1_n-\eta\widetilde{\mathbf h}_w}
+
\log\frac{\mathbf1_n-\eta\widetilde{\mathbf h}_w}
              {\mathbf1_n-\eta\mathbf h_w}
\right\|_\infty\\
&\le
\left\|\log\frac{\mathbf1_n-\eta\hat{\mathbf h}^{(t+1)}}
              {\mathbf1_n-\eta\widetilde{\mathbf h}_w}\right\|_\infty
+
\left\|\log\frac{\mathbf1_n-\eta\widetilde{\mathbf h}_w}
              {\mathbf1_n-\eta\mathbf h_w}\right\|_\infty\\
&\le2e_t+K\omega_n.
\end{aligned}
\end{equation}
Taking the maximum of the two norms on the left
in~\eqref{eq: comparison denominator decomposition}
and substituting
\eqref{eq: comparison accumulated tracking error} gives
\begin{equation}\label{eq: comparison logarithmic denominator error}
\max\left\{
\left\|\log\frac{\mathbf1_m-\eta\hat{\mathbf g}^{(t+1)}}
                 {\mathbf1_m-\eta\bar{\mathbf g}^{(t+1)}}\right\|_\infty,
\left\|\log\frac{\mathbf1_n-\eta\hat{\mathbf h}^{(t+1)}}
                 {\mathbf1_n-\eta\bar{\mathbf h}^{(t+1)}}\right\|_\infty
\right\}
\le2e_t+K\omega_n
\le K(t+1)\omega_n.
\end{equation}

Finally, apply Lemma~\ref{lem: candidate decrease comparison}
with
$(\mathbf g,\mathbf h)
=(\bar{\mathbf g}^{(t+1)},\bar{\mathbf h}^{(t+1)})$,
$(\hat{\mathbf g},\hat{\mathbf h})
=(\hat{\mathbf g}^{(t+1)},\hat{\mathbf h}^{(t+1)})$,
and $b=b_{t+1}$.
Equation~\eqref{eq: comparison empirical branch selection}
is the rule~\eqref{eq: empirical candidate selection}
for this choice of
$(\hat{\mathbf g},\hat{\mathbf h})$.
Under the substitution for $(\mathbf g,\mathbf h)$,
the quantities $\Delta_{\mathrm{row}}$ and
$\Delta_{\mathrm{col}}$ in
Lemma~\ref{lem: candidate decrease comparison}
equal $\bar\Delta_{\mathrm{row}}^{(t+1)}$ and
$\bar\Delta_{\mathrm{col}}^{(t+1)}$
from~\eqref{eq: comparison population candidate decreases}.
The discrepancy $e$ defined
in~\eqref{eq: candidate logarithmic denominator error}
is at most $K(t+1)\omega_n$
by~\eqref{eq: comparison logarithmic denominator error}.
The perturbed greedy inequality
\eqref{eq: perturbed greedy comparison} gives the first inequality
below, using the bound $e\le2e_t+K\omega_n$
in~\eqref{eq: comparison logarithmic denominator error}.
The second inequality uses
\eqref{eq: comparison accumulated tracking error}.
\begin{equation}\label{eq: comparison selected decrease estimate}
\begin{aligned}
\bar\Delta_{b_{t+1}}^{(t+1)}
&\ge
\frac14\max\left\{
\bar\Delta_{\mathrm{row}}^{(t+1)},
\bar\Delta_{\mathrm{col}}^{(t+1)}
\right\}
-\frac34n(2e_t+K\omega_n)^2\\
&\ge
\frac14\max\left\{
\bar\Delta_{\mathrm{row}}^{(t+1)},
\bar\Delta_{\mathrm{col}}^{(t+1)}
\right\}
-Kn(t+1)^2\omega_n^2.
\end{aligned}
\end{equation}
The aspect-ratio assumption gives
\begin{equation}\label{eq: comparison dimension conversion}
\frac mn\ge\gamma_-
\quad\Longrightarrow\quad
n\le\frac{m}{\gamma_-}.
\end{equation}
By~\eqref{eq: comparison dimension conversion},
the error term $Kn(t+1)^2\omega_n^2$
in~\eqref{eq: comparison selected decrease estimate}
is at most $(K/\gamma_-)m(t+1)^2\omega_n^2$.
Since the constant in the lemma may depend on $\gamma_-$,
replacing $K$ by a larger constant proves
\eqref{eq: approximate greedy population decrease}.
\end{proof}

\subsection{Main proof}

\begin{proof}[Proof of Theorem~\ref{thm: main theoretical guarantees}]
We work under the event $\mathcal E_n$ of Lemma~\ref{lem: uniform empirical population comparison}, and use its
comparison vectors and candidate decreases. Constants denoted by $K$
may increase between estimates and have the dependencies allowed
for $K_2$ in the theorem.
Let $\bar J_t$ be the potential normalized by $m$ and shifted by
$-\alpha$, as in~\eqref{eq: normalized potential fun}, evaluated at
$(\bar{\mathbf x}^{(t)},\bar{\mathbf y}^{(t)},
\bar{\mathbf g}^{(t)},\bar{\mathbf h}^{(t)})$, for $t\ge1$.
The uniform population bounds~\eqref{eq: uniform population bounds}
keep these scaling and Dyson coordinates in the domain of the
population curvature estimates; write $\mu\in(0,1)$ for their constant.

With the current Dyson vectors held fixed, the chosen scaling update
decreases $J$ by exactly $\bar\Delta_{b_{t+1}}^{(t+1)}$, by the
candidate-decrease definition~\eqref{eq:thm2_decrease_definitions}.
Hence
\begin{equation}\label{eq: approximate potential contraction}
\begin{aligned}
\bar J_{t+1}-J_\star
&=\frac{J(\bar{\mathbf x}^{(t)},\bar{\mathbf y}^{(t)},
        \bar{\mathbf g}^{(t+1)},\bar{\mathbf h}^{(t+1)})}{m}
-\alpha-J_\star-\frac{\bar\Delta_{b_{t+1}}^{(t+1)}}{m}\\
&\le(1-\mu/8)\left[
\frac{J(\bar{\mathbf x}^{(t)},\bar{\mathbf y}^{(t)},
        \bar{\mathbf g}^{(t+1)},\bar{\mathbf h}^{(t+1)})}{m}
-\alpha-J_\star\right]+K(t+1)^2\omega_n^2\\
&\le(1-\mu/8)(\bar J_t-J_\star)+K(t+1)^2\omega_n^2,
\qquad 1\le t<T_n.
\end{aligned}
\end{equation}
The first inequality combines the lower bound on the chosen decrease
\eqref{eq: approximate greedy population decrease} with the population
decrease estimate~\eqref{eq: quantitative greedy decrease}, which bounds
the best candidate below by $\mu/2$ times the current unnormalized
potential gap. For the last inequality, Dyson minimization
\eqref{eq:thm2_dyson_descent} says that replacing
$(\bar{\mathbf g}^{(t)},\bar{\mathbf h}^{(t)})$ by
$(\bar{\mathbf g}^{(t+1)},\bar{\mathbf h}^{(t+1)})$ cannot increase $J$
at the fixed scalings $(\bar{\mathbf x}^{(t)},\bar{\mathbf y}^{(t)})$.
The population curvature bound controlling distance by the potential
gap~\eqref{eq: distance controlled by potential} also gives
\begin{equation}\label{eq: comparison curvature estimates}
d\bigl((\bar{\mathbf x}^{(t)},\bar{\mathbf y}^{(t)}),
       (\mathbf x^\star,\mathbf y^\star)\bigr)^2
\le\frac2\mu(\bar J_t-J_\star),\qquad 1\le t\le T_n.
\end{equation}
It applies to this bounded four-tuple even though its Dyson vectors
were computed before the scaling update.

Choose the optimal representative with $\sum_i x_i^\star=m$,
whose coordinate bounds are given in~\eqref{eq: bounded optimal scalings}.
Dyson minimization at the initial scalings and nonnegativity of the
first chosen scaling decrease give
\begin{equation}\label{eq: comparison initialization bound}
\begin{aligned}
0\le\bar J_1-J_\star
&\le\frac{J(\mathbf1_m,\mathbf1_n,\mathbf g^\star,\mathbf h^\star)}m
-\alpha-J_\star\\
&=\frac\alpha m\bigl[L(\mathbf1_m,\mathbf1_n)
-L(\mathbf x^\star,\mathbf y^\star)\bigr]\le K_0.
\end{aligned}
\end{equation}
Here $L$ is the logarithmic barrier objective
\eqref{eq: logarithmic barrier function}. The equality uses
$g_i^\star h_j^\star=\alpha$, obtained from the optimal Dyson formulas
\eqref{eq: formulas for g_star and h_star} and the scalar identity
\eqref{eq: critical point analysis 3}.
The constants $K_0,\mu$ depend only on $\eta,c,C,\gamma_-,\gamma_+$.
Set $\widetilde\rho:=\sqrt{1-\mu/8}$.
Iterating the potential recurrence~\eqref{eq: approximate potential contraction}
from~\eqref{eq: comparison initialization bound} gives, for $1\le t\le T_n$,
\[
\begin{aligned}
\bar J_t-J_\star
&\le K_0\widetilde\rho^{2(t-1)}
+K\omega_n^2\sum_{s=2}^t\widetilde\rho^{2(t-s)}s^2\\
&\le K_0\widetilde\rho^{2(t-1)}+Kt^2\omega_n^2,
\end{aligned}
\]
using $s^2\le t^2$ and
$\sum_{s=2}^t\widetilde\rho^{2(t-s)}\le(1-\widetilde\rho^2)^{-1}$.

Choose $a=1$ in the definition of the averaged logarithmic distance
\eqref{eq: averaged logarithmic distance}.
Each of its two averages is at most $(Kt\omega_n)^2$ by the
coordinate tracking bound~\eqref{eq: comparison tracking bound}.
Thus the distance between the empirical and comparison scalings is
at most $\sqrt2Kt\omega_n$, with $\sqrt2$ absorbed into $K$ below.
The triangle inequality and the bound of distance by the potential gap
\eqref{eq: comparison curvature estimates} yield
\[
\begin{aligned}
d\bigl((\hat{\mathbf x}^{(t)},\hat{\mathbf y}^{(t)}),
       (\mathbf x^\star,\mathbf y^\star)\bigr)
&\le Kt\omega_n+\sqrt{\frac2\mu(\bar J_t-J_\star)}\\
&\le\sqrt{2K_0/\mu}\,\widetilde\rho^{t-1}+Kt\omega_n.
\end{aligned}
\]
At $t=0$, apply the same population bound of distance by the potential gap to
$(\mathbf1_m,\mathbf1_n,\mathbf g^\star,\mathbf h^\star)$.
Its normalized potential gap is bounded by $K_0$ in
\eqref{eq: comparison initialization bound}, so the initial distance
is at most $\sqrt{2K_0/\mu}$.
Take $K_1:=\sqrt{2K_0/\mu}/\widetilde\rho$ and choose $K_2$ large enough
to cover the final error coefficient and the failure probability in
Lemma~\ref{lem: uniform empirical population comparison}.
This proves the claimed bound simultaneously for $0\le t\le T_n$,
with the required dependencies of the constants and dimension threshold.
\end{proof}

\section{Proof of Theorem~\ref{thm: generic singular value gain}} \label{appendix: proof of theorem on singular value gain}

\subsection{Supporting lemmas}
\begin{lem}\label{lem: lower bound on average scaling factor}
Suppose that~\eqref{eq: average noise variance assumption} holds, let
$(\mathbf{x}^\star,\mathbf{y}^\star)$ be a solution to the scaling equations~\eqref{eq: scaling equations}, and let
$\Delta(S)$, $r_i$, and $c_j$ be as defined in Section~\ref{sec: signal enhancement}. Then
\begin{equation}
\frac{1}{mn}\sum_{i=1}^m\sum_{j=1}^n \mathbf{x}_i^\star \mathbf{y}_j^\star
=
\left(\frac{1}{m}\sum_{i=1}^m \mathbf{x}_i^\star\right)
\left(\frac{1}{n}\sum_{j=1}^n \mathbf{y}_j^\star\right)
\ge e^{\Delta(S)}
\ge
\max\left\{
\left(\prod_{i=1}^m r_i\right)^{-1/m},
\left(\prod_{j=1}^n \left(\frac{n}{m}c_j\right)\right)^{-1/n}
\right\}
\ge 1 .
\label{eq: Gamma lower bounds corrected}
\end{equation}
\end{lem}

\begin{proof}
Since~\eqref{eq: average noise variance assumption} holds, we have
\[
L(\mathbf{1}_m,\mathbf{1}_n)
=
\mathbf{1}_m^T S \mathbf{1}_n
=
\sum_{i=1}^m\sum_{j=1}^n S_{ij}
=
m.
\]
Also, because $(\mathbf{x}^\star,\mathbf{y}^\star)$ satisfies the scaling equations~\eqref{eq: scaling equations},
\[
(\mathbf{x}^\star)^T S \mathbf{y}^\star
=
\sum_{i=1}^m\sum_{j=1}^n \mathbf{x}_i^\star S_{ij}\mathbf{y}_j^\star
=
m.
\]
Therefore
\[
L(\mathbf{x}^\star,\mathbf{y}^\star)
=
m-\sum_{i=1}^m \log(\mathbf{x}_i^\star)
-\frac{m}{n}\sum_{j=1}^n \log(\mathbf{y}_j^\star),
\]
and from the definition of $\Delta(S)$ in~\eqref{eq: Delta S def},
\[
\Delta(S)
=
\frac{1}{m}\sum_{i=1}^m \log(\mathbf{x}_i^\star)
+
\frac{1}{n}\sum_{j=1}^n \log(\mathbf{y}_j^\star).
\]
Exponentiating gives
\[
e^{\Delta(S)}
=
\left(\prod_{i=1}^m \mathbf{x}_i^\star\right)^{1/m}
\left(\prod_{j=1}^n \mathbf{y}_j^\star\right)^{1/n}.
\]
Applying AM--GM to $\mathbf{x}^\star$ and $\mathbf{y}^\star$ yields
\[
\left(\prod_{i=1}^m \mathbf{x}_i^\star\right)^{1/m}
\le \frac{1}{m}\sum_{i=1}^m \mathbf{x}_i^\star,
\qquad
\left(\prod_{j=1}^n \mathbf{y}_j^\star\right)^{1/n}
\le \frac{1}{n}\sum_{j=1}^n \mathbf{y}_j^\star,
\]
so
\[
e^{\Delta(S)}
\le
\left(\frac{1}{m}\sum_{i=1}^m \mathbf{x}_i^\star\right)
\left(\frac{1}{n}\sum_{j=1}^n \mathbf{y}_j^\star\right)
=
\frac{1}{mn}\sum_{i=1}^m\sum_{j=1}^n \mathbf{x}_i^\star \mathbf{y}_j^\star .
\]

To obtain the explicit lower bounds, note that $(\mathbf{x}^\star,\mathbf{y}^\star)$ is a global minimizer of $L$, hence
\[
L(\mathbf{x}^\star,\mathbf{y}^\star)\le \inf_{\mathbf{x}\succ 0} L(\mathbf{x},\mathbf{1}_n).
\]
Therefore
\[
\Delta(S)
=
\frac{L(\mathbf{1}_m,\mathbf{1}_n)-L(\mathbf{x}^\star,\mathbf{y}^\star)}{m}
\ge
\frac{L(\mathbf{1}_m,\mathbf{1}_n)-\inf_{\mathbf{x}\succ 0} L(\mathbf{x},\mathbf{1}_n)}{m}.
\]
Since
\[
L(\mathbf{x},\mathbf{1}_n)
=
\sum_{i=1}^m \bigl(r_i \mathbf{x}_i-\log \mathbf{x}_i\bigr),
\qquad
r_i=\sum_{j=1}^n S_{ij},
\]
each summand is minimized at $\mathbf{x}_i=r_i^{-1}$, with minimum value $1+\log r_i$. Hence
\[
\inf_{\mathbf{x}\succ 0} L(\mathbf{x},\mathbf{1}_n)
=
\sum_{i=1}^m (1+\log r_i),
\]
and using $L(\mathbf{1}_m,\mathbf{1}_n)=m$ we get
\[
\Delta(S)\ge -\frac{1}{m}\sum_{i=1}^m \log r_i
=
\log\!\left( \left(\prod_{i=1}^m r_i\right)^{-1/m}\right).
\]
Thus
\[
e^{\Delta(S)}
\ge
\left(\prod_{i=1}^m r_i\right)^{-1/m}.
\]

The column bound is analogous. Since $(\mathbf{x}^\star,\mathbf{y}^\star)$ minimizes $L$,
\[
L(\mathbf{x}^\star,\mathbf{y}^\star)\le \inf_{\mathbf{y}\succ 0} L(\mathbf{1}_m,\mathbf{y}),
\]
and
\[
L(\mathbf{1}_m,\mathbf{y})
=
\sum_{j=1}^n \left(c_j \mathbf{y}_j-\frac{m}{n}\log \mathbf{y}_j\right),
\qquad
c_j=\sum_{i=1}^m S_{ij}.
\]
Each summand is minimized at $\mathbf{y}_j=(m/n)c_j^{-1}$, with minimum value
\[
\frac{m}{n}+\frac{m}{n}\log\!\left(\frac{n}{m}c_j\right).
\]
Hence
\[
\inf_{\mathbf{y}\succ 0} L(\mathbf{1}_m,\mathbf{y})
=
m+\frac{m}{n}\sum_{j=1}^n \log\!\left(\frac{n}{m}c_j\right),
\]
so
\[
\Delta(S)\ge -\frac{1}{n}\sum_{j=1}^n \log\!\left(\frac{n}{m}c_j\right)
=
\log\!\left(
\left(\prod_{j=1}^n \left(\frac{n}{m}c_j\right)\right)^{-1/n}
\right).
\]
Therefore
\[
e^{\Delta(S)}
\ge
\left(\prod_{j=1}^n \left(\frac{n}{m}c_j\right)\right)^{-1/n}.
\]

Finally, since $\frac{1}{m}\sum_{i=1}^m r_i=1$ and
$\frac{1}{n}\sum_{j=1}^n \frac{n}{m}c_j=1$, AM--GM gives
\[
\left(\prod_{i=1}^m r_i\right)^{1/m}\le 1,
\qquad
\left(\prod_{j=1}^n \left(\frac{n}{m}c_j\right)\right)^{1/n}\le 1,
\]
hence
\[
\left(\prod_{i=1}^m r_i\right)^{-1/m}\ge 1,
\qquad
\left(\prod_{j=1}^n \left(\frac{n}{m}c_j\right)\right)^{-1/n}\ge 1.
\]
This proves~\eqref{eq: Gamma lower bounds corrected}.
\end{proof}

\subsection{Main proof}

All quantities below depend on $(m,n)$, but we suppress this dependence in the notation.

Because of the scalar ambiguity
\[
(\mathbf{x}^\star,\mathbf{y}^\star)\mapsto (a\mathbf{x}^\star,a^{-1}\mathbf{y}^\star),
\qquad a>0,
\]
we may choose, for each pair $(m,n)$, the representative satisfying
\[
\frac{1}{m}\sum_{i=1}^m x_i^\star
=
\frac{1}{n}\sum_{j=1}^n y_j^\star
=
\Gamma(S)^{1/2},
\]
where
\[
\Gamma(S)
=
\left(\frac{1}{m}\sum_{i=1}^m x_i^\star\right)
\left(\frac{1}{n}\sum_{j=1}^n y_j^\star\right).
\]
Indeed, if
\[
\bar x := \frac{1}{m}\sum_{i=1}^m x_i^\star,
\qquad
\bar y := \frac{1}{n}\sum_{j=1}^n y_j^\star,
\]
then choosing \(a=(\bar y/\bar x)^{1/2}\) gives
\[
\frac{1}{m}\sum_{i=1}^m a x_i^\star
=
\frac{1}{n}\sum_{j=1}^n a^{-1} y_j^\star
=
(\bar x \bar y)^{1/2}
=
\Gamma(S)^{1/2}.
\]
This change does not affect \(\widetilde X\), \(\Gamma(S)\), or \(\Delta(S)\).

Under this normalization, the scaling equations and the bounds on \(S\) imply uniform bounds on the scaling factors.
Indeed, for each \(i\in[m]\),
\[
1
=
x_i^\star \sum_{j=1}^n S_{ij} y_j^\star,
\]
and since \(\sum_{j=1}^n y_j^\star = n\Gamma(S)^{1/2}\),
\[
c\,\Gamma(S)^{1/2}
\le
\sum_{j=1}^n S_{ij} y_j^\star
\le
C\,\Gamma(S)^{1/2}.
\]
Hence
\[
\frac{1}{C\Gamma(S)^{1/2}}
\le x_i^\star \le
\frac{1}{c\Gamma(S)^{1/2}}.
\]
Similarly, for each \(j\in[n]\),
\[
\frac{m}{n}
=
y_j^\star \sum_{i=1}^m x_i^\star S_{ij},
\]
and \(\sum_{i=1}^m x_i^\star = m\Gamma(S)^{1/2}\), so
\[
\frac{1}{C\Gamma(S)^{1/2}}
\le y_j^\star \le
\frac{1}{c\Gamma(S)^{1/2}}.
\]
Averaging the bounds for \(x_i^\star\) gives
\[
\frac{1}{C\Gamma(S)^{1/2}}
\le \Gamma(S)^{1/2}
\le \frac{1}{c\Gamma(S)^{1/2}},
\]
hence
\[
\frac{1}{C}\le \Gamma(S)\le \frac{1}{c}.
\]
Substituting this back yields uniform deterministic bounds
\[
\frac{\sqrt{c}}{C}\le x_i^\star,y_j^\star \le \frac{\sqrt{C}}{c}
\qquad \forall i\in[m],\ \forall j\in[n].
\]

Now define
\[
A_m := \widetilde U^T D\{\mathbf{x}^\star\}\widetilde U,
\qquad
B_n := \widetilde V^T D\{\mathbf{y}^\star\}\widetilde V.
\]
Let
\[
X=\widetilde U D\{s_1,\ldots,s_r\}\widetilde V^T
\]
be the SVD from~\eqref{eq: SVD of X}. By taking the polar decompositions of
\((D\{\mathbf{x}^\star\})^{1/2}\widetilde U\) and \((D\{\mathbf{y}^\star\})^{1/2}\widetilde V\),
there exist matrices \(Q_m\in\mathbb{R}^{m\times r}\) and \(P_n\in\mathbb{R}^{n\times r}\) with orthonormal columns such that
\[
(D\{\mathbf{x}^\star\})^{1/2}\widetilde U = Q_m A_m^{1/2},
\qquad
(D\{\mathbf{y}^\star\})^{1/2}\widetilde V = P_n B_n^{1/2}.
\]
Therefore
\[
\widetilde X
=
Q_m A_m^{1/2} D\{s_1,\ldots,s_r\} B_n^{1/2} P_n^T,
\]
and hence the nonzero singular values of \(\widetilde X\) are exactly those of
\(A_m^{1/2} D\{s_1,\ldots,s_r\} B_n^{1/2}\).

We next show that
\[
\left\|A_m-\Gamma(S)^{1/2}I_r\right\|_{\mathrm{op}}\to 0,
\qquad
\left\|B_n-\Gamma(S)^{1/2}I_r\right\|_{\mathrm{op}}\to 0
\qquad \text{almost surely as } m,n\to\infty.
\]
We prove the first convergence; the second is identical.

Using the standard Gaussian representation of the Haar measure on the Stiefel manifold (see, e.g.,~\cite{mardia1977uniform} or~\cite{chikuse2003statistics}), we may write
\[
\widetilde U = G_m (G_m^T G_m)^{-1/2},
\]
where \(G_m\in\mathbb{R}^{m\times r}\) has i.i.d.\ standard Gaussian entries.
Then
\[
A_m
=
\left(\frac{1}{m}G_m^T G_m\right)^{-1/2}
\left(\frac{1}{m}G_m^T D\{\mathbf{x}^\star\} G_m\right)
\left(\frac{1}{m}G_m^T G_m\right)^{-1/2}.
\]
Since \(r\) is fixed,
\[
\frac{1}{m}G_m^T G_m \to I_r
\qquad\text{almost surely}
\]
entrywise, hence also in operator norm.

It remains to analyze \(m^{-1}G_m^T D\{\mathbf{x}^\star\} G_m\).
Fix \(a,b\in[r]\). Write
\[
Z_{m,ab}
:=
\frac{1}{m}\bigl[G_m^T D\{\mathbf{x}^\star\} G_m\bigr]_{ab}
=
\frac{1}{m}\sum_{i=1}^m x_i^\star g_{ia}g_{ib}.
\]
If \(a\neq b\), then the summands are independent, mean zero, and uniformly bounded in fourth moment because the \(x_i^\star\) are uniformly bounded and Gaussian moments are finite. Hence
\[
\mathbb{E}\!\left[ |Z_{m,ab}|^4 \,\middle|\, S \right]
\le \frac{K}{m^2}
\]
for some constant \(K\) independent of \(m\). Markov's inequality and Borel--Cantelli, together with $m/n\geq \gamma_{-}$, give
\[
Z_{m,ab}\to 0
\qquad\text{almost surely.}
\]

If \(a=b\), then
\[
Z_{m,aa}
=
\frac{1}{m}\sum_{i=1}^m x_i^\star
+
\frac{1}{m}\sum_{i=1}^m x_i^\star(g_{ia}^2-1)
=
\Gamma(S)^{1/2}
+
\frac{1}{m}\sum_{i=1}^m x_i^\star(g_{ia}^2-1).
\]
Again the second term is an average of independent mean-zero random variables with uniformly bounded fourth moments, so
\[
\frac{1}{m}\sum_{i=1}^m x_i^\star(g_{ia}^2-1)\to 0
\qquad\text{almost surely.}
\]
Thus \(\vert Z_{m,aa} - \Gamma(S)^{1/2}\vert \to 0\) almost surely.

Since \(r\) is fixed, all entrywise differences converge to zero jointly, and therefore
\[
\left\Vert \frac{1}{m}G_m^T D\{\mathbf{x}^\star\} G_m
-
\Gamma(S)^{1/2} I_r \right\Vert_\mathrm{op} \to 0
\qquad\text{almost surely.}
\]
Combining this with \(m^{-1}G_m^T G_m\to I_r\) yields
\[
\left\Vert A_m - \Gamma(S)^{1/2} I_r \right\Vert_\mathrm{op} \to 0
\qquad\text{almost surely.}
\]
The same argument gives
\[
\left\Vert B_n - \Gamma(S)^{1/2} I_r \right\Vert_\mathrm{op} \to 0
\qquad\text{almost surely.}
\]

In particular,
\begin{equation} \label{eq: convergence of smallest eigenvalues of A_m and B_n}
    \vert \lambda_{\min}(A_m) - \Gamma(S)^{1/2} \vert \to 0,
\qquad
\vert \lambda_{\min}(B_n) -  \Gamma(S)^{1/2} \vert \to 0
\qquad\text{almost surely.}
\end{equation}

We now apply the standard singular-value inequality~\cite[Theorem~3.3.16(d), p.~178]{horn1994topics}
\[
\sigma_k(CMD)\ge \sigma_{\min}(C)\,\sigma_k(M)\,\sigma_{\min}(D)
\]
for square matrices \(C,M,D \in\mathbb{R}^{r\times r}\). Applying it with
\[
C=A_m^{1/2},\qquad M=D\{s_1,\ldots,s_r\},\qquad D=B_n^{1/2},
\]
we obtain
\[
\sigma_k(\widetilde X)
\ge
\lambda_{\min}(A_m)^{1/2}\, s_k\, \lambda_{\min}(B_n)^{1/2}.
\]
Therefore
\[
\frac{\sigma_k(\widetilde X)}{s_k e^{\Delta(S)/2}}
\ge
\frac{\lambda_{\min}(A_m)^{1/2}\lambda_{\min}(B_n)^{1/2}}{e^{\Delta(S)/2}}.
\]
Taking the liminf and using~\eqref{eq: convergence of smallest eigenvalues of A_m and B_n} gives
\[
\liminf_{m,n\to\infty}
\frac{\sigma_k(\widetilde X)}{s_k e^{\Delta(S)/2}}
\ge
\liminf_{m,n\to\infty} \frac{\Gamma(S)^{1/2}}{e^{\Delta(S)/2}}
\qquad\text{almost surely.}
\]
By Lemma~\ref{lem: lower bound on average scaling factor}, $\Gamma(S)\ge e^{\Delta(S)}$,
and hence
\[
\frac{\Gamma(S)^{1/2}}{e^{\Delta(S)/2}}\ge 1.
\]
This proves~\eqref{eq: liminf singular value lower bound Delta}. The explicit lower bound on \(e^{\Delta(S)/2}\) is exactly the one supplied by Lemma~\ref{lem: lower bound on average scaling factor}.

\section{Low-rank approximation of the signal} \label{sec: low-rank approximation}
This appendix provides details on the low-rank approximation of the signal used in Section~\ref{sec:experiments-scrna}.
Let $\hat{Y}$ denote the normalized matrix returned by Algorithm~\ref{alg:BRINE} applied to the library-normalized data, together with the estimated scaling factors $\hat{\mathbf{x}}$ and $\hat{\mathbf{y}}$. We have
\begin{equation}
\hat{Y} = \left( D\{\hat{\mathbf{x}}\}\right)^{1/2} (cY) \left(D\{\hat{\mathbf{y}}\}\right)^{1/2},
\label{eq: Y_hat final}
\end{equation}
where $c>0$ is the scalar normalization applied in the initialization step of Algorithm~\ref{alg:BRINE}. Writing the SVD of the normalized matrix as $\hat{Y} = \sum_{k=1}^m \hat{\sigma}_k \hat{\mathbf{u}}_k \hat{\mathbf{v}}_k^T$ with $\hat{\sigma}_1 \geq \ldots \geq \hat{\sigma}_m$, we estimate the rank $\hat{r}$ by counting the singular values that exceed the MP upper edge,
\begin{equation}
\hat{r} = \# \left\{ k \in [m] : \hat{\sigma}_k > 1 + \sqrt{m/n} \right\},
\label{eq: r_hat def}
\end{equation}
as described in Section~\ref{Sec: sinkhorn scaling the MP law}. We then truncate the normalized matrix at this rank,
\begin{equation}
\hat{Y}_{\hat{r}} = \sum_{k=1}^{\hat{r}} \hat{\sigma}_k \hat{\mathbf{u}}_k \hat{\mathbf{v}}_k^T,
\label{eq: Y_hat truncated}
\end{equation}
and map the result back to the scale of the observed data by inverting the row and column scalings from~\eqref{eq: Y_hat final},
\begin{equation}
\overline{X} = c^{-1} \left( D\{\hat{\mathbf{x}}\}\right)^{-1/2} \hat{Y}_{\hat{r}} \left(D\{\hat{\mathbf{y}}\}\right)^{-1/2}.
\label{eq: X_bar def}
\end{equation}

Two remarks are in order. First, since the diagonal scalings in~\eqref{eq: X_bar def} are invertible, the estimated signal matrix $\overline{X}$ has rank exactly $\hat{r}$. Second, the SVD truncation is performed on the normalized matrix $\hat{Y}$ rather than on $Y$, i.e., after the noise equalization. By Theorem~\ref{thm: generic singular value gain}, this equalization can amplify the signal singular values relative to the noise, thereby enhancing the recovered signal. This procedure was proposed in~\cite{landa2025dyson}, where it was shown to admit a favorable interpretation as a weighted low-rank approximation, with a direct connection to maximum-likelihood estimation of the low-rank signal under heteroskedastic Gaussian noise.

\section{Additional spectral fits and convergence diagnostics}
\label{app:all_mp_fits}
\begin{figure}[t]
\centering
\includegraphics[width=1.0\linewidth]{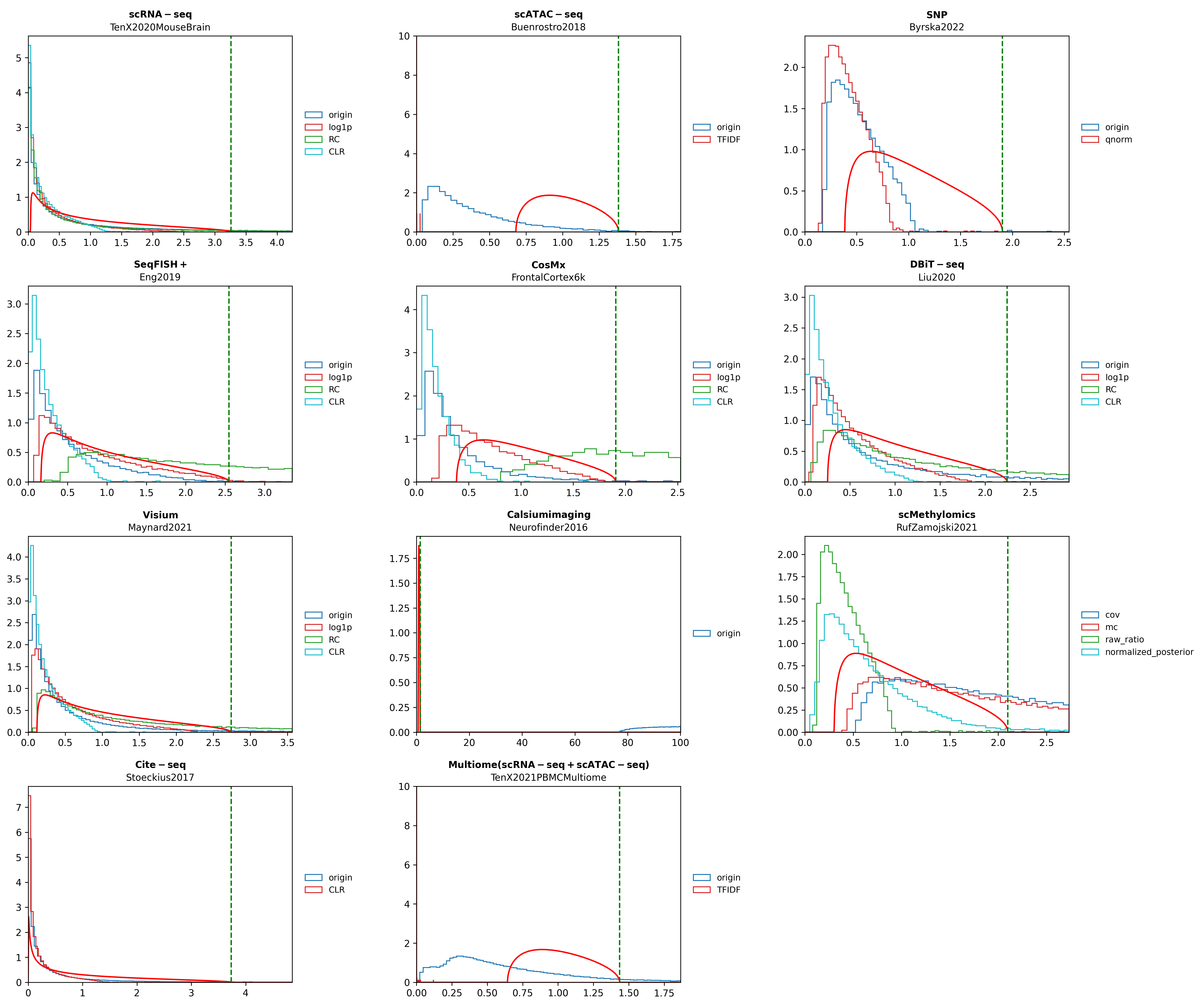}
\caption{Empirical spectral distributions before normalization across all datasets and modalities. Each panel corresponds to one dataset; different histogram colors represent different preprocessing transformations. The red curve indicates the theoretical Marchenko--Pastur (MP) density, and green dashed lines mark the MP upper edge $\beta_+$. Spectra are scaled by a constant for visualization.}
\label{fig:before_dyson}
\end{figure}

\begin{figure}[t]
\centering
\includegraphics[width=1.0\linewidth]{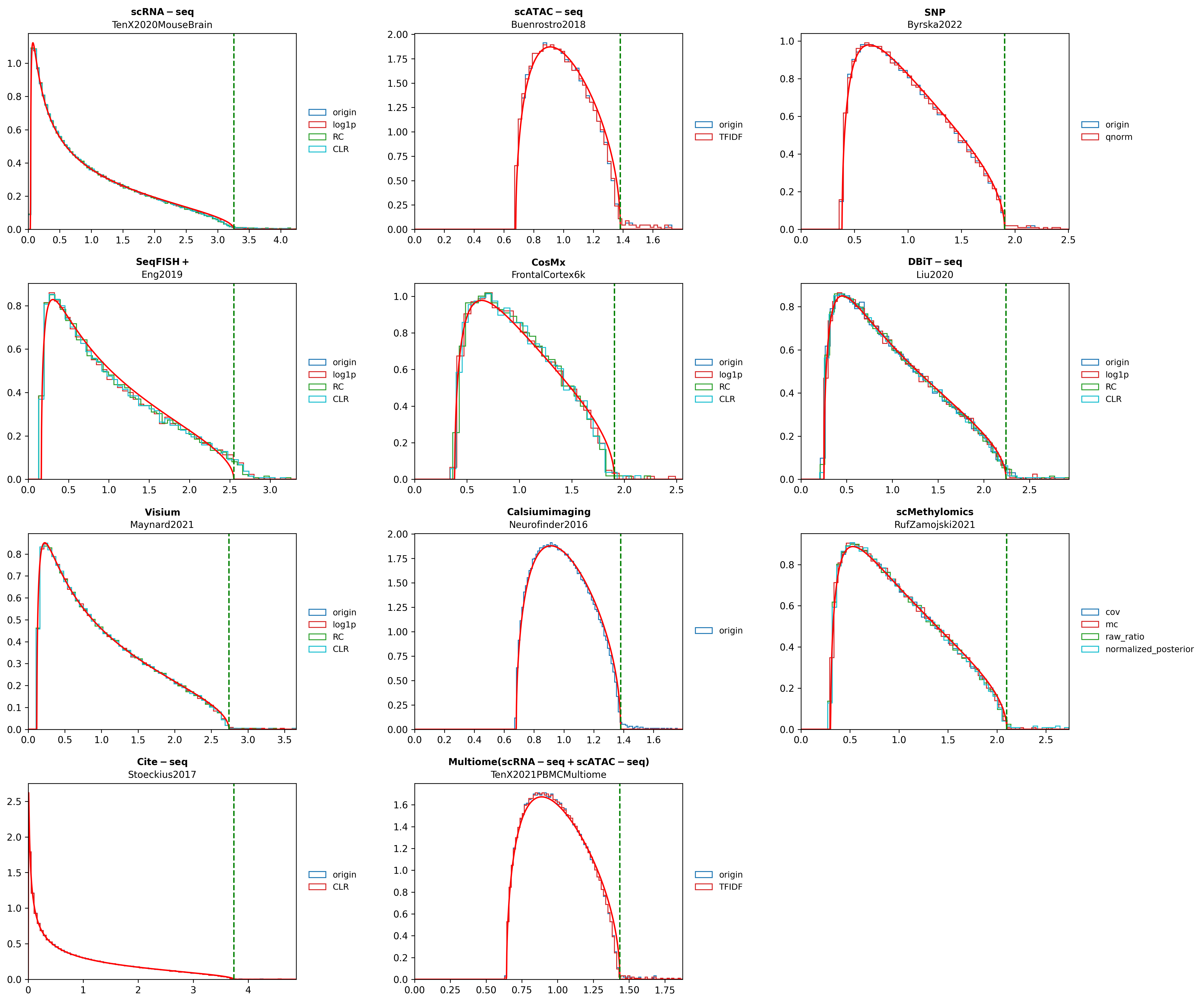}
\caption{Empirical spectral distributions after normalization for the same datasets and transformations as in Figure~\ref{fig:before_dyson}. After normalization, all spectra closely follow the theoretical MP distribution (red curve) regardless of modality or preprocessing. Green dashed lines mark the MP upper edge $\beta_+$.}
\label{fig:after_dyson}
\end{figure}

This appendix provides the complete spectral fits and convergence diagnostics for all 11 datasets and all preprocessing transformations considered in Section~\ref{sec:experiments-mp}. Figure~\ref{fig:before_dyson} shows the empirical spectral distributions before normalization, Figure~\ref{fig:after_dyson} shows the corresponding distributions after applying Algorithm \ref{alg:BRINE}, and Figure~\ref{fig:dyson_convergence} traces the convergence of the optimization across iterations.

We briefly describe the datasets and transformations below. Unless otherwise noted, data acquisition and filtering followed the protocols established in~\cite{stanley2025principled}. For \textbf{single-cell RNA sequencing}, we use 10X Genomics Chromium V3.1 (10X2020MouseBrain) and CITE-seq RNA (Stoeckius2017), with three transformations: Log1p ($\log(1+x)$), Relative Counts (column-stochastic normalization), and Centered Log-Ratio (CLR). For \textbf{spatial transcriptomics}, we use four platforms, SeqFISH+ (Eng2019), CosMx (FrontalCortex6k), DBiT-seq (Liu2020), and Visium (Maynard2021), with the same three transformations. For \textbf{scATAC-seq} (Buenrostro2018), we apply TF-IDF (Term Frequency-Inverse Document Frequency). For \textbf{genotyping} (Bryska2022), we apply quantile normalization (qnorm). For \textbf{single-cell methylomics} (RufZamojski2021), we consider four representations: methylated counts (mc), coverage counts (cov), the raw ratio ($Y_{ij} = \text{mc}_{ij}/\text{cov}_{ij}$) and the normalized posterior ratio ($\hat{Y}_{ij} \propto (\text{mc}_{ij}+\alpha)/(\text{cov}_{ij}+\beta)$, representing the Bayesian posterior mean estimate of the methylation rate). For \textbf{calcium imaging} (Neurofinder2016), no additional transformation is applied. Finally, we include a \textbf{multimodal} dataset (10X2021PBMCMultiome~\cite{10xgenomics_pbmc}), in which single-cell RNA expression and chromatin accessibility (ATAC) are concatenated along the feature axis into a single matrix. Quality control is applied to both modalities simultaneously: cells are required to have at least 100 detected genes and 100 accessible chromatin sites, and cells with mitochondrial UMI counts exceeding 10\% of total library size are excluded. Features detected in fewer than 100 cells are removed. TF-IDF is applied to the combined matrix.

\begin{figure}[t]
\centering
\includegraphics[width=1.0\linewidth]{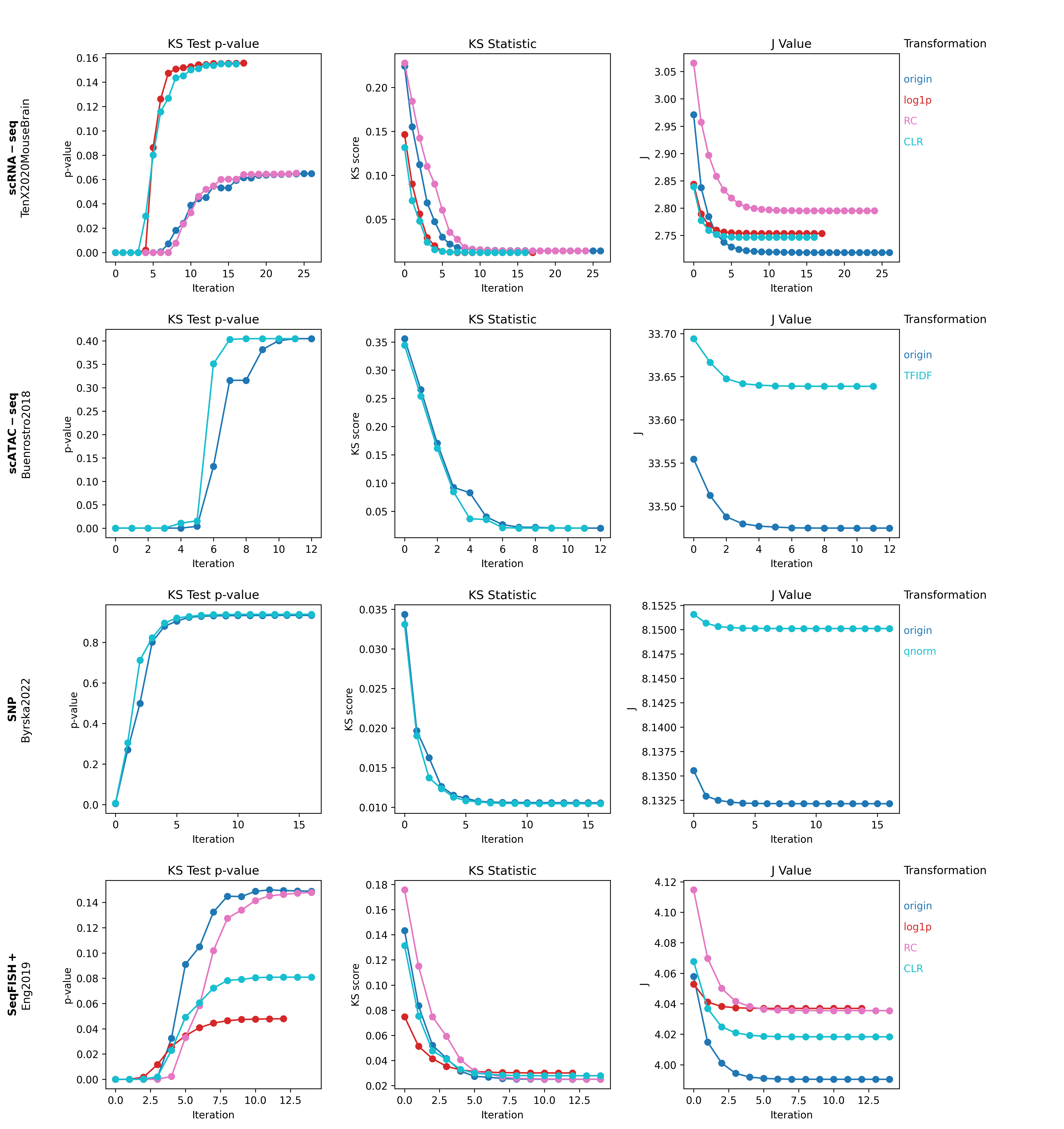}
\caption{Convergence of BRINE normalization across all datasets. Each row corresponds to one dataset; each color represents a distinct preprocessing transformation. The three columns show the Kolmogorov--Smirnov $p$-value, KS statistic, and the estimated potential $\hat{J}_t$ over successive iterations.}
\label{fig:dyson_convergence}
\end{figure}
\begin{figure}[t]
\centering
\ContinuedFloat
\includegraphics[width=1.0\linewidth]{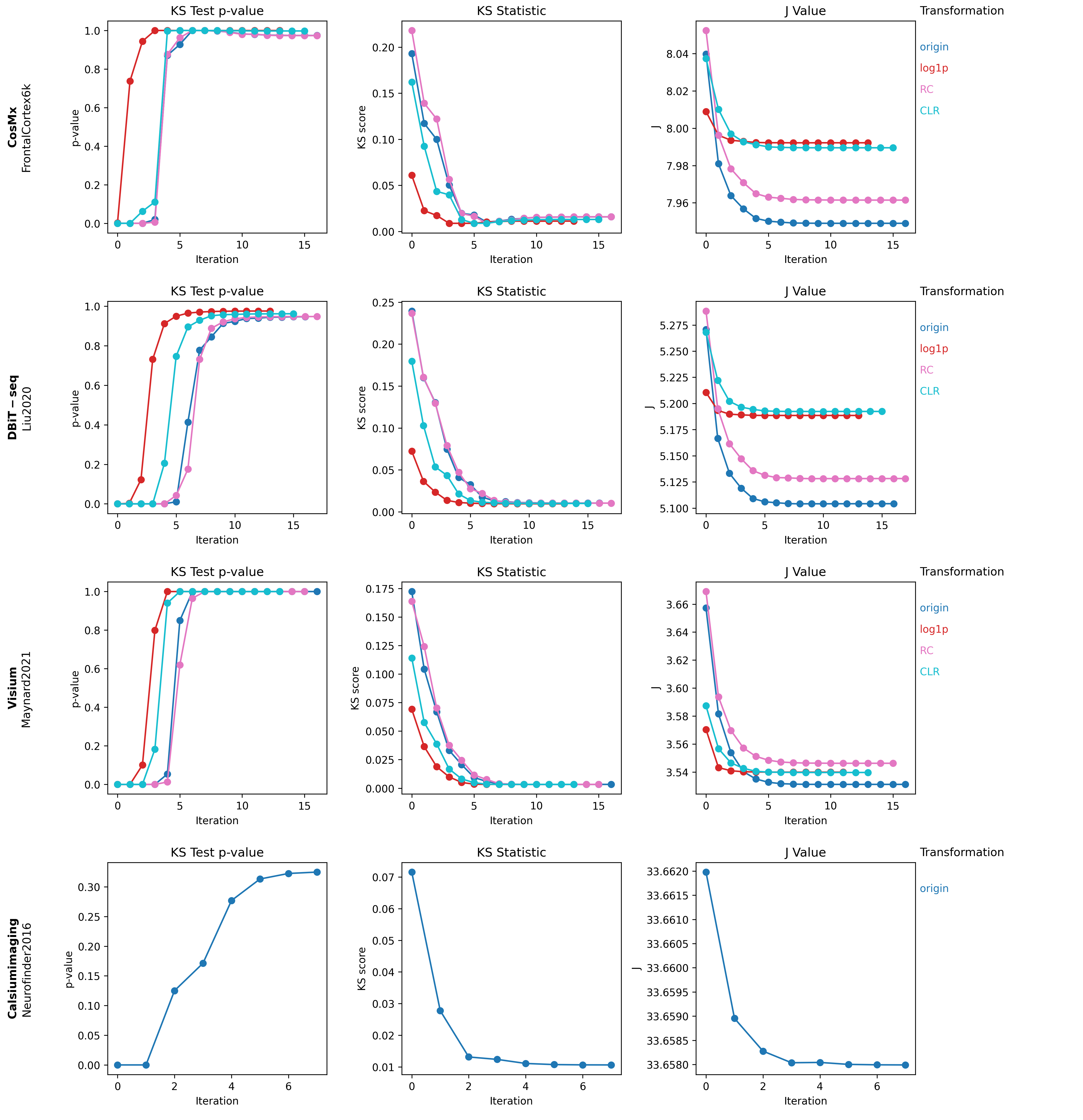}
\caption[]{Convergence of BRINE normalization across all datasets (cont).}

\end{figure}
\begin{figure}[t]
\centering
\ContinuedFloat
\includegraphics[width=1.0\linewidth]{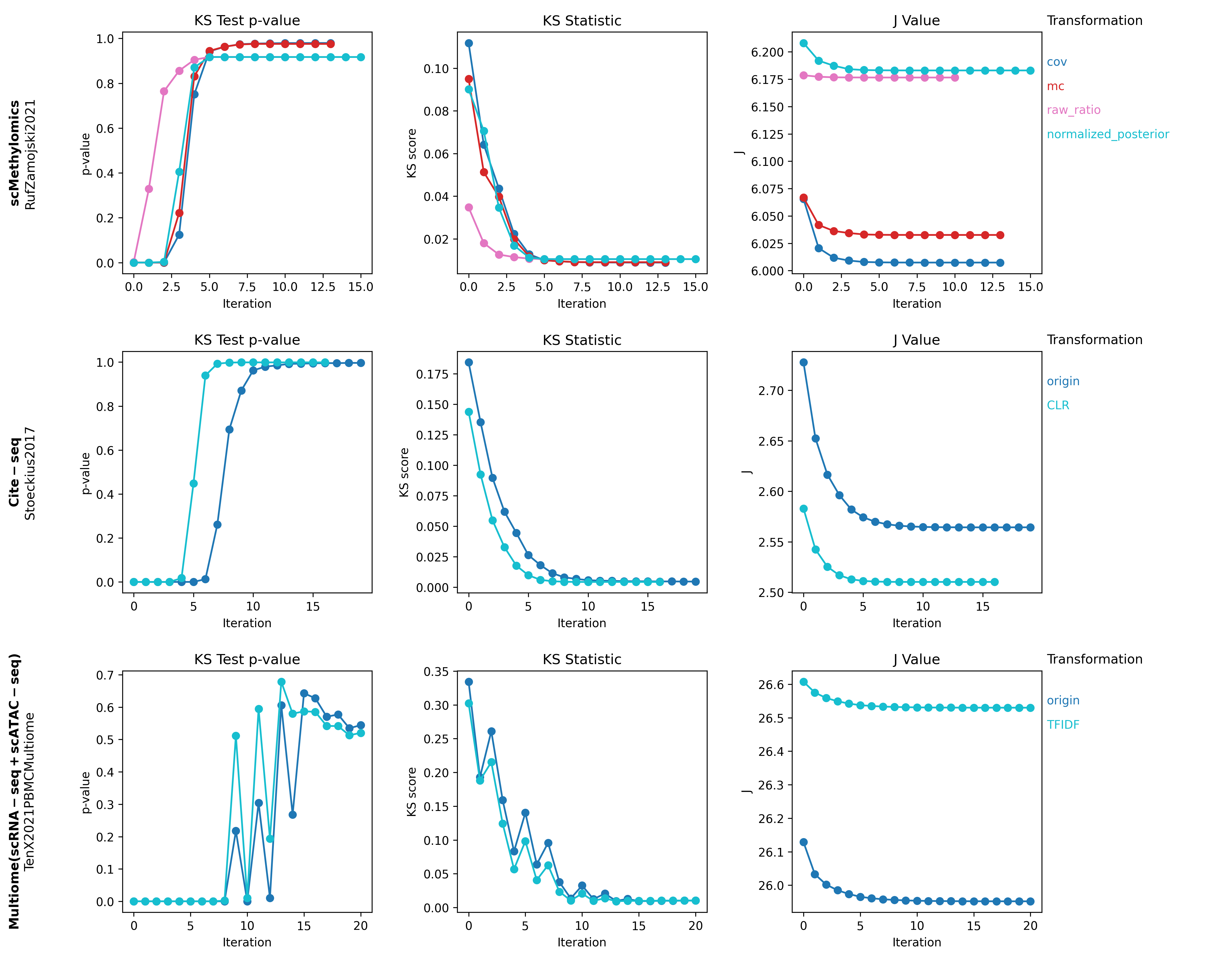}
\caption[]{Convergence of BRINE normalization across all datasets (cont).}

\end{figure}

\end{appendices}

\clearpage
\bibliographystyle{plain}
\bibliography{mybib}

@book{horn1994topics,
  title={Topics in matrix analysis},
  author={Horn, Roger A and Johnson, Charles R},
  year={1994},
  publisher={Cambridge university press}
}

@article{zhang2022heteroskedastic,
  title={Heteroskedastic PCA: Algorithm, optimality, and applications},
  author={Zhang, Anru R and Cai, T Tony and Wu, Yihong},
  journal={The Annals of Statistics},
  volume={50},
  number={1},
  pages={53--80},
  year={2022},
  publisher={Institute of Mathematical Statistics}
}

@article{gavish2014optimal,
  title={The optimal hard threshold for singular values is $4 / \sqrt{3} $},
  author={Gavish, Matan and Donoho, David L},
  journal={IEEE Transactions on Information Theory},
  volume={60},
  number={8},
  pages={5040--5053},
  year={2014},
  publisher={IEEE}
}

@article{alt2017local,
  title={Local law for random Gram matrices},
  author={Alt, Johannes and Erd{\H{o}}s, L{\'a}szl{\'o} and Kr{\"u}ger, Torben and others},
  journal={Electronic Journal of Probability},
  volume={22},
  year={2017},
  publisher={The Institute of Mathematical Statistics and the Bernoulli Society}
}

@article{hong2020selecting,
  title={Selecting the number of components in PCA via random signflips},
  author={Hong, David and Sheng, Yue and Dobriban, Edgar},
  journal={arXiv preprint arXiv:2012.02985},
  year={2020}
}

@article{fan2020estimating,
  title={Estimating number of factors by adjusted eigenvalues thresholding},
  author={Fan, Jianqing and Guo, Jianhua and Zheng, Shurong},
  journal={Journal of the American Statistical Association},
  pages={1--10},
  year={2020},
  publisher={Taylor \& Francis}
}

@article{ke2021estimation,
  title={Estimation of the number of spiked eigenvalues in a covariance matrix by bulk eigenvalue matching analysis},
  author={Ke, Zheng Tracy and Ma, Yucong and Lin, Xihong},
  journal={Journal of the American Statistical Association},
  pages={1--19},
  year={2021},
  publisher={Taylor \& Francis}
}

@article{hong2018optimally,
  title={Optimally weighted PCA for high-dimensional heteroscedastic data},
  author={Hong, David and Fessler, Jeffrey A and Balzano, Laura},
  journal={arXiv preprint arXiv:1810.12862},
  year={2018}
}

@article{leeb2021optimal,
  title={Optimal spectral shrinkage and PCA with heteroscedastic noise},
  author={Leeb, William and Romanov, Elad},
  journal={IEEE Transactions on Information Theory},
  volume={67},
  number={5},
  pages={3009--3037},
  year={2021},
  publisher={IEEE}
}

@article{leeb2021matrix,
  title={Matrix denoising for weighted loss functions and heterogeneous signals},
  author={Leeb, William E},
  journal={SIAM Journal on Mathematics of Data Science},
  volume={3},
  number={3},
  pages={987--1012},
  year={2021},
  publisher={SIAM}
}

@article{liu2018pca,
  title={$ e $ PCA: high dimensional exponential family PCA},
  author={Liu, Lydia T and Dobriban, Edgar and Singer, Amit and others},
  journal={The Annals of Applied Statistics},
  volume={12},
  number={4},
  pages={2121--2150},
  year={2018},
  publisher={Institute of Mathematical Statistics}
}

@article{shen2005analysis,
  title={Analysis of call centre arrival data using singular value decomposition},
  author={Shen, Haipeng and Huang, Jianhua Z},
  journal={Applied Stochastic Models in Business and Industry},
  volume={21},
  number={3},
  pages={251--263},
  year={2005},
  publisher={Wiley Online Library}
}

@article{hafemeister2019normalization,
  title={Normalization and variance stabilization of single-cell RNA-seq data using regularized negative binomial regression},
  author={Hafemeister, Christoph and Satija, Rahul},
  journal={Genome biology},
  volume={20},
  number={1},
  pages={1--15},
  year={2019},
  publisher={Springer}
}

@inproceedings{wallach2006topic,
  title={Topic modeling: beyond bag-of-words},
  author={Wallach, Hanna M},
  booktitle={Proceedings of the 23rd international conference on Machine learning},
  pages={977--984},
  year={2006}
}

@article{salmon2014poisson,
  title={Poisson noise reduction with non-local PCA},
  author={Salmon, Joseph and Harmany, Zachary and Deledalle, Charles-Alban and Willett, Rebecca},
  journal={Journal of mathematical imaging and vision},
  volume={48},
  number={2},
  pages={279--294},
  year={2014},
  publisher={Springer}
}

@article{choi2017selecting,
  title={Selecting the number of principal components: Estimation of the true rank of a noisy matrix},
  author={Choi, Yunjin and Taylor, Jonathan and Tibshirani, Robert},
  journal={The Annals of Statistics},
  pages={2590--2617},
  year={2017},
  publisher={JSTOR}
}

@article{kritchman2008determining,
  title={Determining the number of components in a factor model from limited noisy data},
  author={Kritchman, Shira and Nadler, Boaz},
  journal={Chemometrics and Intelligent Laboratory Systems},
  volume={94},
  number={1},
  pages={19--32},
  year={2008},
  publisher={Elsevier}
}

@article{baik2005phase,
  title={Phase transition of the largest eigenvalue for nonnull complex sample covariance matrices},
  author={Baik, Jinho and Arous, G{\'e}rard Ben and P{\'e}ch{\'e}, Sandrine and others},
  journal={The Annals of Probability},
  volume={33},
  number={5},
  pages={1643--1697},
  year={2005},
  publisher={Institute of Mathematical Statistics}
}

@article{benaych2011eigenvalues,
  title={The eigenvalues and eigenvectors of finite, low rank perturbations of large random matrices},
  author={Benaych-Georges, Florent and Nadakuditi, Raj Rao},
  journal={Advances in Mathematics},
  volume={227},
  number={1},
  pages={494--521},
  year={2011},
  publisher={Elsevier}
}

@article{baik2006eigenvalues,
  title={Eigenvalues of large sample covariance matrices of spiked population models},
  author={Baik, Jinho and Silverstein, Jack W},
  journal={Journal of Multivariate Analysis},
  volume={97},
  number={6},
  pages={1382--1408},
  year={2006},
  publisher={Elsevier}
}

@article{geman1980limit,
  title={A limit theorem for the norm of random matrices},
  author={Geman, Stuart},
  journal={The Annals of Probability},
  pages={252--261},
  year={1980},
  publisher={JSTOR}
}

@article{yin1988limit,
  title={On the limit of the largest eigenvalue of the large dimensional sample covariance matrix},
  author={Yin, Yong-Qua and Bai, Zhi-Dong and Krishnaiah, Pathak R},
  journal={Probability theory and related fields},
  volume={78},
  number={4},
  pages={509--521},
  year={1988},
  publisher={Springer}
}

@book{bai2010spectral,
  title={Spectral analysis of large dimensional random matrices},
  author={Bai, Zhidong and Silverstein, Jack W},
  volume={20},
  year={2010},
  publisher={Springer}
}

@article{nadler2008finite,
  title={Finite sample approximation results for principal component analysis: A matrix perturbation approach},
  author={Nadler, Boaz and others},
  journal={The Annals of Statistics},
  volume={36},
  number={6},
  pages={2791--2817},
  year={2008},
  publisher={Institute of Mathematical Statistics}
}

@article{paul2007asymptotics,
  title={Asymptotics of sample eigenstructure for a large dimensional spiked covariance model},
  author={Paul, Debashis},
  journal={Statistica Sinica},
  pages={1617--1642},
  year={2007},
  publisher={JSTOR}
}

@article{marvcenko1967distribution,
  title={Distribution of eigenvalues for some sets of random matrices},
  author={Mar{\v{c}}enko, Vladimir A and Pastur, Leonid Andreevich},
  journal={Mathematics of the USSR-Sbornik},
  volume={1},
  number={4},
  pages={457},
  year={1967},
  publisher={IOP Publishing}
}

@article{hoff2007model,
  title={Model averaging and dimension selection for the singular value decomposition},
  author={Hoff, Peter D},
  journal={Journal of the American Statistical Association},
  volume={102},
  number={478},
  pages={674--685},
  year={2007},
  publisher={Taylor \& Francis}
}

@article{cattell1966scree,
  title={The scree test for the number of factors},
  author={Cattell, Raymond B},
  journal={Multivariate behavioral research},
  volume={1},
  number={2},
  pages={245--276},
  year={1966},
  publisher={Taylor \& Francis}
}

@article{wold1978cross,
  title={Cross-validatory estimation of the number of components in factor and principal components models},
  author={Wold, Svante},
  journal={Technometrics},
  volume={20},
  number={4},
  pages={397--405},
  year={1978},
  publisher={Taylor \& Francis}
}

@article{owen2009bi,
  title={Bi-cross-validation of the SVD and the nonnegative matrix factorization},
  author={Owen, Art B and Perry, Patrick O and others},
  journal={The Annals of Applied Statistics},
  volume={3},
  number={2},
  pages={564--594},
  year={2009},
  publisher={Institute of Mathematical Statistics}
}

@article{chatterjee2015matrix,
  title={Matrix estimation by universal singular value thresholding},
  author={Chatterjee, Sourav and others},
  journal={The Annals of Statistics},
  volume={43},
  number={1},
  pages={177--214},
  year={2015},
  publisher={Institute of Mathematical Statistics}
}

@article{johnstone2018pca,
  title={Pca in high dimensions: An orientation},
  author={Johnstone, Iain M and Paul, Debashis},
  journal={Proceedings of the IEEE},
  volume={106},
  number={8},
  pages={1277--1292},
  year={2018},
  publisher={IEEE}
}

@article{fan2014principal,
  title={Principal component analysis for big data},
  author={Fan, Jianqing and Sun, Qiang and Zhou, Wen-Xin and Zhu, Ziwei},
  journal={Wiley StatsRef: Statistics Reference Online},
  pages={1--13},
  year={2014},
  publisher={Wiley Online Library}
}

@article{johnstone2017roy,
  title={Roy’s largest root test under rank-one alternatives},
  author={Johnstone, IM and Nadler, Boaz},
  journal={Biometrika},
  volume={104},
  number={1},
  pages={181--193},
  year={2017},
  publisher={Oxford University Press}
}

@article{shekhar2016comprehensive,
  title={Comprehensive classification of retinal bipolar neurons by single-cell transcriptomics},
  author={Shekhar, Karthik and Lapan, Sylvain W and Whitney, Irene E and Tran, Nicholas M and Macosko, Evan Z and Kowalczyk, Monika and Adiconis, Xian and Levin, Joshua Z and Nemesh, James and Goldman, Melissa and others},
  journal={Cell},
  volume={166},
  number={5},
  pages={1308--1323},
  year={2016},
  publisher={Elsevier}
}

@article{idel2016review,
  title={A review of matrix scaling and sinkhorn's normal form for matrices and positive maps},
  author={Idel, Martin},
  journal={arXiv preprint arXiv:1609.06349},
  year={2016}
}

@article{sinkhorn1967concerning,
  title={Concerning nonnegative matrices and doubly stochastic matrices},
  author={Sinkhorn, Richard and Knopp, Paul},
  journal={Pacific Journal of Mathematics},
  volume={21},
  number={2},
  pages={343--348},
  year={1967},
  publisher={Mathematical Sciences Publishers}
}

@article{sinkhorn1967diagonal,
  title={Diagonal equivalence to matrices with prescribed row and column sums},
  author={Sinkhorn, Richard},
  journal={The American Mathematical Monthly},
  volume={74},
  number={4},
  pages={402--405},
  year={1967},
  publisher={JSTOR}
}

@article{grun2014validation,
  title={Validation of noise models for single-cell transcriptomics},
  author={Gr{\"u}n, Dominic and Kester, Lennart and Van Oudenaarden, Alexander},
  journal={Nature Methods},
  volume={11},
  number={6},
  pages={637--640},
  year={2014},
  publisher={Nature Publishing Group}
}

@article{morris1982natural,
  title={Natural exponential families with quadratic variance functions},
  author={Morris, Carl N},
  journal={The Annals of Statistics},
  pages={65--80},
  year={1982},
  publisher={JSTOR}
}

@article{hong2018asymptotic,
  title={Asymptotic performance of PCA for high-dimensional heteroscedastic data},
  author={Hong, David and Balzano, Laura and Fessler, Jeffrey A},
  journal={Journal of Multivariate Analysis},
  volume={167},
  pages={435--452},
  year={2018},
  publisher={Elsevier}
}

@article{johanson2018genome,
  title={Genome-wide analysis reveals no evidence of trans chromosomal regulation of mammalian immune development},
  author={Johanson, Timothy M and Coughlan, Hannah D and Lun, Aaron TL and Bediaga, Naiara G and Naselli, Gaetano and Garnham, Alexandra L and Harrison, Leonard C and Smyth, Gordon K and Allan, Rhys S},
  journal={PLoS Genetics},
  volume={14},
  number={6},
  pages={e1007431},
  year={2018},
  publisher={Public Library of Science San Francisco, CA USA}
}

@book{ajanki2019quadratic,
  title={Quadratic vector equations on complex upper half-plane},
  author={Ajanki, Oskari and Erd{\H{o}}s, L{\'a}szl{\'o} and Kr{\"u}ger, Torben},
  volume={261},
  number={1261},
  year={2019},
  publisher={American Mathematical Society}
}

@inproceedings{erdHos2019random,
  title={Random matrices with slow correlation decay},
  author={Erd{\H{o}}s, L{\'a}szl{\'o} and Kr{\"u}ger, Torben and Schr{\"o}der, Dominik},
  booktitle={Forum of Mathematics, Sigma},
  volume={7},
  year={2019},
  organization={Cambridge University Press}
}

@article{landa2022biwhitening,
  title={Biwhitening reveals the rank of a count matrix},
  author={Landa, Boris and Zhang, Thomas TCK and Kluger, Yuval},
  journal={SIAM Journal on Mathematics of Data Science},
  volume={4},
  number={4},
  pages={1420--1446},
  year={2022},
  publisher={SIAM}
}

@article{ding2022tracy,
  title={Tracy-widom distribution for heterogeneous gram matrices with applications in signal detection},
  author={Ding, Xiucai and Yang, Fan},
  journal={IEEE Transactions on Information Theory},
  volume={68},
  number={10},
  pages={6682--6715},
  year={2022},
  publisher={IEEE}
}

@article{gavish2022matrix,
  title={Matrix Denoising with Partial Noise Statistics: Optimal Singular Value Shrinkage of Spiked F-Matrices},
  author={Gavish, Matan and Leeb, William and Romanov, Elad},
  journal={arXiv preprint arXiv:2211.00986},
  year={2022}
}

@article{cochran1977statistically,
  title={Statistically weighted principal component analysis of rapid scanning wavelength kinetics experiments},
  author={Cochran, Robert N and Horne, Frederick H},
  journal={Analytical Chemistry},
  volume={49},
  number={6},
  pages={846--853},
  year={1977},
  publisher={ACS Publications}
}

@article{tamuz2005correcting,
  title={Correcting systematic effects in a large set of photometric light curves},
  author={Tamuz, Omer and Mazeh, Tsevi and Zucker, Shay},
  journal={Monthly Notices of the Royal Astronomical Society},
  volume={356},
  number={4},
  pages={1466--1470},
  year={2005},
  publisher={Blackwell Science Ltd Oxford, UK}
}

@article{foi2009clipped,
  title={Clipped noisy images: Heteroskedastic modeling and practical denoising},
  author={Foi, Alessandro},
  journal={Signal Processing},
  volume={89},
  number={12},
  pages={2609--2629},
  year={2009},
  publisher={Elsevier}
}

@article{burgess2019spatial,
  title={Spatial transcriptomics coming of age},
  author={Burgess, Darren J},
  journal={Nature Reviews Genetics},
  volume={20},
  number={6},
  pages={317--317},
  year={2019},
  publisher={Nature Publishing Group}
}

@article{gavish2017optimal,
  title={Optimal shrinkage of singular values},
  author={Gavish, Matan and Donoho, David L},
  journal={IEEE Transactions on Information Theory},
  volume={63},
  number={4},
  pages={2137--2152},
  year={2017},
  publisher={IEEE}
}

@article{leeb2022optimal,
  title={Optimal singular value shrinkage for operator norm loss: Extending to non-square matrices},
  author={Leeb, William},
  journal={Statistics \& Probability Letters},
  volume={186},
  pages={109472},
  year={2022},
  publisher={Elsevier}
}

@article{nadakuditi2014optshrink,
  title={Optshrink: An algorithm for improved low-rank signal matrix denoising by optimal, data-driven singular value shrinkage},
  author={Nadakuditi, Raj Rao},
  journal={IEEE Transactions on Information Theory},
  volume={60},
  number={5},
  pages={3002--3018},
  year={2014},
  publisher={IEEE}
}

@article{donoho2023screenot,
  title={ScreeNOT: Exact MSE-optimal singular value thresholding in correlated noise},
  author={Donoho, David and Gavish, Matan and Romanov, Elad},
  journal={The Annals of Statistics},
  volume={51},
  number={1},
  pages={122--148},
  year={2023},
  publisher={Institute of Mathematical Statistics}
}

@article{su2022adaptive,
  title={Adaptive Optimal Shrinkage of Singular Values under Colored and Dependent Noise},
  author={Su, Pei-Chun and Wu, Hau-Tieng},
  journal={arXiv preprint arXiv:2207.03466},
  year={2022}
}

@article{cole2019performance,
  title={Performance assessment and selection of normalization procedures for single-cell RNA-seq},
  author={Cole, Michael B and Risso, Davide and Wagner, Allon and DeTomaso, David and Ngai, John and Purdom, Elizabeth and Dudoit, Sandrine and Yosef, Nir},
  journal={Cell systems},
  volume={8},
  number={4},
  pages={315--328},
  year={2019},
  publisher={Elsevier}
}

@article{ding2020high,
author = {Xiucai Ding},
title = {{High dimensional deformed rectangular matrices with applications in matrix denoising}},
volume = {26},
journal = {Bernoulli},
number = {1},
publisher = {Bernoulli Society for Mathematical Statistics and Probability},
pages = {387 -- 417},
year = {2020},
doi = {10.3150/19-BEJ1129},
URL = {https://doi.org/10.3150/19-BEJ1129}
}

@article{landa2025dyson,
  title={The dyson equalizer: Adaptive noise stabilization for low-rank signal detection and recovery},
  author={Landa, Boris and Kluger, Yuval},
  journal={Information and Inference: A Journal of the IMA},
  volume={14},
  number={1},
  pages={iaae036},
  year={2025},
  publisher={Oxford University Press}
}

@article{marshall1968scaling,
  title={Scaling of matrices to achieve specified row and column sums},
  author={Marshall, Albert W and Olkin, Ingram},
  journal={Numerische Mathematik},
  volume={12},
  number={1},
  pages={83--90},
  year={1968},
  publisher={Springer}
}

@article{zheng2017massively,
  title={Massively parallel digital transcriptional profiling of single cells},
  author={Zheng, Grace XY and Terry, Jessica M and Belgrader, Phillip and Ryvkin, Paul and Bent, Zachary W and Wilson, Ryan and Ziraldo, Solongo B and Wheeler, Tobias D and McDermott, Geoff P and Zhu, Junjie and others},
  journal={Nature communications},
  volume={8},
  number={1},
  pages={14049},
  year={2017},
  publisher={Nature Publishing Group UK London}
}

@article{abdelaal2019comparison,
  title={A comparison of automatic cell identification methods for single-cell RNA sequencing data},
  author={Abdelaal, Tamim and Michielsen, Lieke and Cats, Davy and Hoogduin, Dylan and Mei, Hailiang and Reinders, Marcel JT and Mahfouz, Ahmed},
  journal={Genome biology},
  volume={20},
  number={1},
  pages={194},
  year={2019},
  publisher={Springer}
}

@misc{10xgenomics_pbmc,
  author = {{10x Genomics}},
  title = {PBMC from a Healthy Donor, Granulocytes Removed through Cell Sorting (10k)},
  year = {2021},
  url = {https://www.10xgenomics.com/resources/datasets/pbmc-from-a-healthy-donor-granulocytes-removed-through-cell-sorting-10-k-1-standard-2-0-0}
}

@article{stanley2025principled,
  title={Principled PCA separates signal from noise in omics count data},
  author={Stanley III, Jay S and Yang, Junchen and Li, Ruiqi and Lindenbaum, Ofir and Kobak, Dmitry and Landa, Boris and Kluger, Yuval},
  journal={bioRxiv},
  year={2025}
}

@article{mardia1977uniform,
  author  = {Mardia, K. V. and Khatri, C. G.},
  title   = {Uniform Distribution on a Stiefel Manifold},
  journal = {Journal of Multivariate Analysis},
  volume  = {7},
  number  = {3},
  pages   = {468--473},
  year    = {1977}
}

@book{chikuse2003statistics,
  author    = {Chikuse, Yasuko},
  title     = {Statistics on Special Manifolds},
  series    = {Lecture Notes in Statistics},
  volume    = {174},
  publisher = {Springer},
  address   = {New York},
  year      = {2003}
}

@article{zhou2025deflated,
  title={Deflated-HeteroPCA: Overcoming the Curse of Ill-Conditioning in Heteroskedastic PCA},
  author={Zhou, Yuchen and Chen, Yuxin},
  journal={The Annals of Statistics},
  volume={53},
  number={1},
  pages={91--116},
  year={2025},
  publisher={Institute of Mathematical Statistics},
  doi={10.1214/24-AOS2456}
}

@inproceedings{zhang2024matrix,
  title={Matrix Denoising with Doubly Heteroscedastic Noise: Fundamental Limits and Optimal Spectral Methods},
  author={Zhang, Yihan and Mondelli, Marco},
  booktitle={Advances in Neural Information Processing Systems},
  volume={37},
  pages={93060--93117},
  year={2024},
  url={https://proceedings.neurips.cc/paper_files/paper/2024/hash/a96368eb38bce0956a1132154d70d72d-Abstract-Conference.html}
}

@article{traag2019louvain,
  title={From Louvain to Leiden: guaranteeing well-connected communities},
  author={Traag, Vincent A and Waltman, Ludo and Van Eck, Nees Jan},
  journal={Scientific reports},
  volume={9},
  number={1},
  pages={5233},
  year={2019},
  publisher={Nature Publishing Group UK London}
}

@article{stuart2019comprehensive,
  title   = {Comprehensive Integration of Single-Cell Data},
  author  = {Stuart, Tim and Butler, Andrew and Hoffman, Paul
             and Hafemeister, Christoph and Papalexi, Efthymia
             and Mauck, III, William M. and Hao, Yuhan
             and Stoeckius, Marlon and Smibert, Peter
             and Satija, Rahul},
  journal = {Cell},
  volume  = {177},
  pages   = {1888--1902},
  year    = {2019},
  doi     = {10.1016/j.cell.2019.05.031}
}

@article{lin2011selinv,
  author  = {Lin, Lin and Yang, Chao and Meza, Juan C. and Lu, Jianfeng and Ying, Lexing and E, Weinan},
  title   = {{SelInv}---An Algorithm for Selected Inversion of a Sparse Symmetric Matrix},
  journal = {ACM Transactions on Mathematical Software},
  volume  = {37},
  number  = {4},
  pages   = {40:1--40:19},
  year    = {2011},
  doi     = {10.1145/1916461.1916464}
}

@article{bekas2007estimator,
  author  = {Bekas, C. and Kokiopoulou, E. and Saad, Yousef},
  title   = {An estimator for the diagonal of a matrix},
  journal = {Applied Numerical Mathematics},
  volume  = {57},
  number  = {11--12},
  pages   = {1214--1229},
  year    = {2007},
  doi     = {10.1016/j.apnum.2007.01.003}
}

@misc{baston2022stochastic,
  author        = {Baston, Robert A. and Nakatsukasa, Yuji},
  title         = {Stochastic diagonal estimation: probabilistic bounds and an improved algorithm},
  year          = {2022},
  eprint        = {2201.10684},
  archivePrefix = {arXiv},
  primaryClass  = {cs.DS},
  note          = {arXiv:2201.10684},
  doi           = {10.48550/arXiv.2201.10684}
}

\end{document}